\documentclass{article}

\usepackage[preprint]{corl_2026}

\usepackage{times}
\usepackage{amsmath,amsfonts,amssymb}
\usepackage[dvipsnames]{xcolor}

\usepackage{todonotes}
\usepackage{comment}
\usepackage{soul}

\usepackage{graphicx}
\usepackage{booktabs}
\usepackage{subcaption}
\usepackage{wrapfig}
\usepackage{booktabs}
\usepackage{multirow}
\usepackage{tabularx}
\usepackage{adjustbox}
\usepackage{makecell}

\usepackage{algorithm}
\usepackage{algpseudocode}

\usepackage{pifont}
\usepackage{bbm}

\newcommand{\yc}[1]{\textcolor{blue}{#1}}

\title{EgoEngine: From Egocentric Human Videos \\ to High-Fidelity Dexterous Robot Demonstrations}

\author{
Yangcen Liu$^{1}$,
Shuo Cheng$^{1}$,
Xinchen Yin$^{1\dagger}$,
Woo Chul Shin$^{1\dagger}$,
Alfred Cueva$^{1\dagger}$,\\
\textbf{Yiran Yang}$^{2}$,
\textbf{Zhenyang Chen}$^{1}$,
\textbf{Chuye Zhang}$^{1}$,
\textbf{Danfei Xu}$^{1}$ \\
$^1$Georgia Institute of Technology \quad $^2$Tsinghua University
}

\begin{document}

\maketitle
\thispagestyle{empty}
\pagestyle{empty}

\begin{abstract}
Dexterous manipulation is limited by the cost of collecting large-scale robot demonstrations.
Egocentric human videos offer a scalable source of diverse manipulation behaviors, but directly using them for robot learning requires bridging two gaps: the visual gap between human and robot observations, and the action gap between human motion and robot-executable action. 
We propose \textbf{EgoEngine}, a scalable framework for transforming egocentric human manipulation videos into high-fidelity robot data. 
Given an egocentric RGB video, EgoEngine produces: (i) a high-fidelity robot observation video replacing human with robot while preserving scene context and temporal alignment, and (ii) a task-aligned, executable robot action trajectory under feasibility constraints.
Experiments in simulation and on real robots show that EgoEngine enables scalable conversion of human videos into robot data and, to our knowledge, demonstrates the first zero-shot visuomotor dexterous policy learning from egocentric human videos without real-robot demonstrations.
Project website: \url{https://egoengine.github.io}.
\end{abstract}


\begin{figure}[H]
  \centering
       \centering
        \includegraphics[width=0.99\linewidth]{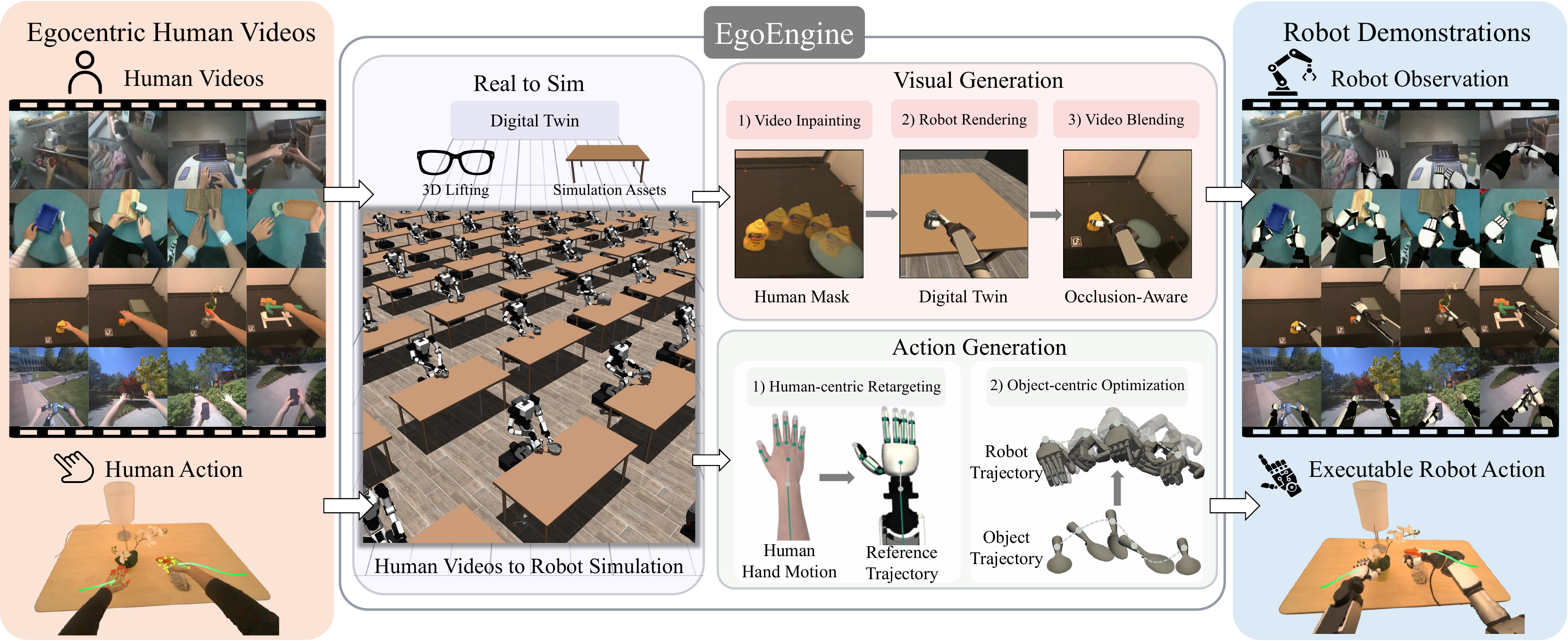}
        \caption{\textbf{EgoEngine} is a scalable data engine that converts egocentric human videos into robot demonstrations.
        Given an egocentric human video, EgoEngine creates a digital twin and jointly produces (1) a high-fidelity, temporally consistent \textbf{robot observation video} and (2) executable \textbf{action trajectories} aligned with the video.
        The generated demonstrations serve as training data that can be distilled into downstream visuomotor policies for \textbf{zero-shot} execution. }
       \label{fig:teaserfig}
        \vspace{-1em}
\end{figure}
 
\section{Introduction}

Scaling robot learning hinges on access to large volumes of high-quality data that capture diverse and physically grounded interactions between robots and environments~\cite{o2024open,khazatsky2024droid}.
Collecting such data on real robots remains costly and time-consuming, especially for dexterous manipulation.
Existing data collection pipelines commonly rely on teleoperation, which is expensive to scale due to limited hardware availability, complex interfaces, and high-DoF contact-rich control~\cite{yin2025dexteritygen,xu2025dexumi,egoverse}. 
As a consequence, many dexterous policies are trained on narrow robot demonstration distributions, motivating scalable non-robot data sources that can reduce or eliminate dependence on robot demonstration collection.
In contrast, egocentric human videos are abundant and naturally capture contact-rich dexterous manipulation across diverse scenes, objects, and interaction styles~\cite{grauman2022ego4d, egoverse, egodex, egoscale}.
Wearable interfaces such as Aria glasses~\cite{projectaria} further make scalable high-fidelity collection efficient for robot learning~\cite{kareer2024egomimic,shi2025zeromimic}.

However, human videos are not robot demonstrations~\cite{immimic,egobridge,thegreatlawrence}. 
The challenge is twofold: visually, human arms and hands occlude the scene and differ substantially from the robot embodiment; on the action side, differences in morphology, kinematics, actuation, and contact dynamics make directly retargeted robot trajectories physically infeasible.
Although several works~\cite{kareer2024egomimic, wang2023mimicplay} use human videos for pretraining or co-training, they still rely primarily on robot teleoperation demonstrations for policy learning.
The central challenge is therefore to turn human videos into robot demonstrations that preserve the task while providing robot observations and executable robot actions~\cite{liu2025egozero,phantom}.

In this work, we introduce \textbf{EgoEngine}, a scalable framework to convert egocentric human videos to high-fidelity robot demonstrations that simultaneously addresses the human-to-robot visual gap and action gap.
As shown in Fig.~\ref{fig:teaserfig}, given a human RGB video, EgoEngine generates a paired robot demonstration consisting of:
(1) a \textbf{robot observation video} that depicts the robot performing the interaction in the original source scene, and
(2) an \textbf{executable robot trajectory} aligned with the task object motion.
By producing observation-action pairs that remain consistent with the original human video and physically executable for the robot, EgoEngine enables high-fidelity large-scale dataset generation and supports \textbf{zero-shot} visuomotor policy learning without any robot teleoperation data.

EgoEngine addresses the core challenges of human-to-robot transfer through an object-centric visual--action generation pipeline.
EgoEngine first preprocesses the video to build a digital twin that aligns the scene, object, and robot embodiment. 
For visual generation, EgoEngine replaces the human with the robot appearance, producing robot observation videos that preserve the original scene context and object state in each frame.
For action generation, EgoEngine first retargets human motion into a robot reference trajectory, then refines it in simulation to produce an executable robot trajectory supervised by reference object motion.
To improve efficiency, EgoEngine uses a Monte Carlo Tree Search-style (MCTS-style) adaptive mode-switching strategy, escalating from Replay to Model Predictive Control (MPC) and Reinforcement Learning (RL) when stronger refinement is needed.
After the two branches, the generated demonstrations support downstream policy training.

Overall, our contributions are threefold:
\begin{itemize}
    \item We treat egocentric human videos as a scalable data source of robot supervision, and present EgoEngine, a two-branch data engine for generating paired robot observation--action demonstrations from egocentric human videos.
    
    \item EgoEngine formulates human-to-robot demonstration generation as jointly bridging the visual and action gaps, and identifies executable action generation as the primary factor of downstream policy performance.

    \item We demonstrate the quality of the generated data through visual fidelity analysis, action evaluation in simulation, and downstream policy distillation, enabling zero-shot dexterous manipulation on a real humanoid robot without using any real robot demonstrations.
\end{itemize}

\section{Related Work}

\textbf{Human-to-Robot Action Generation.}
Human-to-robot action generation is commonly built on teleoperation and retargeting, where human motion is mapped to a robot embodiment~\cite{qin2023anyteleop,sewmimic,opentv,geometricretargeting,humanpolicy}.
Recent glove-based~\cite{osma} and hand-as-interface systems~\cite{xu2025dexumi} improve dexterous data action quality, but remain expensive to scale and require skilled teleoperators. 
However, offline retargeted trajectories can be infeasible under embodiment and dynamics mismatch~\cite{immimic}. 
Some methods address this limitation with structured motion planning~\cite{dexmimicgen,deximit,lodestar}, but require task-specific design. 
Other Real2Sim2Real methods use RL~\cite{zhao2024dexh2r, sim2realvision, mandi2025dexmachinafunctionalretargetingbimanual, h2s2r, refinedp} or MPC~\cite{spider, omniretarget} in simulation to refine trajectories by incorporating object-centric objectives. 
EgoEngine follows this line and focuses on executable trajectory generation by adaptively balancing solver capability and efficiency.

\textbf{Human-to-Robot Video Generation.}
To bridge the visual gap, early methods like EgoMimic~\cite{kareer2024egomimic} mainly reduce the visual gap with lightweight arm masks, while others generate pseudo latent visual features~\cite{immimic}. 
Most methods follow an inpainting--rendering--blending pipeline to replace the human with the robot appearance in videos~\cite{phantom,masquerade,motiontrans,li2025h2r}.
Recent motion- and physics-conditioned video generation and editing methods show promise for controllable robot video synthesis~\cite{xu2024motion,geng2024motion,zhang2024physdreamer,zhang2025think,xhumanoid,vace,rigvid}, but still struggle with physically consistent high-DoF dexterous interactions.
EgoEngine follows the controllable inpainting--rendering--blending pipeline and further grounds it in a digital twin for better physical consistency.

\textbf{Learning from Human Videos.}
A series of works have explored learning robot policies from human demonstrations~\cite{wang2023mimicplay,kareer2024egomimic,dexinthewild}, aiming to bridge the visual and action gaps between humans and robots.
Some methods use human videos for pretraining or high-level policy planning~\cite{egovla,wang2023mimicplay}, while later works formulate the problem as domain adaptation or co-training across human and robot embodiments~\cite{kareer2024egomimic,immimic,egobridge,thegreatlawrence}.
However, these methods still rely on robot demonstrations as the target domain, limiting their ability to learn zero-shot from human videos alone.
Recent methods~\cite{shi2025zeromimic,liu2025egozero} explore zero-shot policy learning from human videos, but without explicit action refinement, they are mainly demonstrated on simpler manipulation settings or lower-DoF grippers.
Instead, EgoEngine turns human videos into demonstrations for zero-shot dexterous manipulation.
\section{EgoEngine Framework}
\label{sec:egoengine_framework}

As illustrated in Fig.~\ref{fig:teaserfig}, EgoEngine converts egocentric human videos into robot-executable demonstrations for policy learning.
Given a human video, EgoEngine first reconstructs an object-centric digital twin comprising camera geometry, depth, object trajectory, and human/object masks.
It then runs two parallel branches: the action branch converts human motion into executable robot actions, and the visual branch converts human egocentric frames into robot-view observations.
Together, they produce paired robot demonstrations $(\tilde{o}_t, \tilde{a}_t)$ for training a downstream visuomotor policy.

\subsection{Human Video to Simulation}
\label{sec:human_preprocess}

For each human video, we reconstruct an object-centric digital twin as follows.
Human demonstrations $\tau^{(h)}$ are captured using Aria Gen2 glasses~\cite{projectaria}, which provide synchronized RGB frames and per-frame 3D hand poses with 21 hand keypoints.
We estimate absolute depth maps from the RGB frames with FoundationStereo~\cite{foundationstereo}.
We then use SAM2~\cite{ravi2024sam2} for two types of masks: hand-keypoint prompts provide human arm--hand masks for removing the demonstrator, while a first-frame point prompt tracks the task-object mask throughout the clip.
Given the RGBD frames, tracked object masks, and object mesh, FoundationPose~\cite{wen2024foundationpose} estimates a temporally consistent 6D object trajectory $\{T_o^t\}_{t=1}^{T}$.
The resulting camera geometry, depth, masks, hand poses, object mesh, and object trajectory together define the digital twin consumed by the action and visual branches.

\subsection{Action Generation}
\label{action_branch}

\subsubsection{Human-Centric Retargeting}
We first retarget the human hand motion to a reference robot trajectory.
Given the fingertip positions and orientations
$\{(p_{\mathrm{tip},k}^t,R_{\mathrm{tip},k}^t)\}_{k=1}^5$
along with the wrist orientation $R_{\mathrm{wrist}}^t$ from the human video at time $t$, we solve an inverse-kinematics problem with MINK~\cite{mink}:
\begin{equation}
q_t^\ast = \arg\min_{q \in \mathcal{Q}}
\mathcal{L}_{\mathrm{tip}}(q; t)
+ \lambda_w \mathcal{L}_{\mathrm{wrist}}(q; t),
\end{equation}
where $\mathcal{L}_{\mathrm{tip}}$ and $\mathcal{L}_{\mathrm{wrist}}$ are L2 losses aligning the robot fingertips with the human fingertip poses and the robot wrist with the human wrist orientation, and $\mathcal{Q}$ is the feasible configuration space subject to joint limits and self-collision constraints.
The resulting trajectory $\tau^{\mathrm{ref}}=\{q_t^\ast\}_{t=1}^{T}$ imitates the human motion on the robot and serves as the initial motion prior for action refinement.

\begin{wrapfigure}{r}{0.52\columnwidth}
    \centering
    \vspace{-2em}
    \includegraphics[width=0.51\columnwidth]{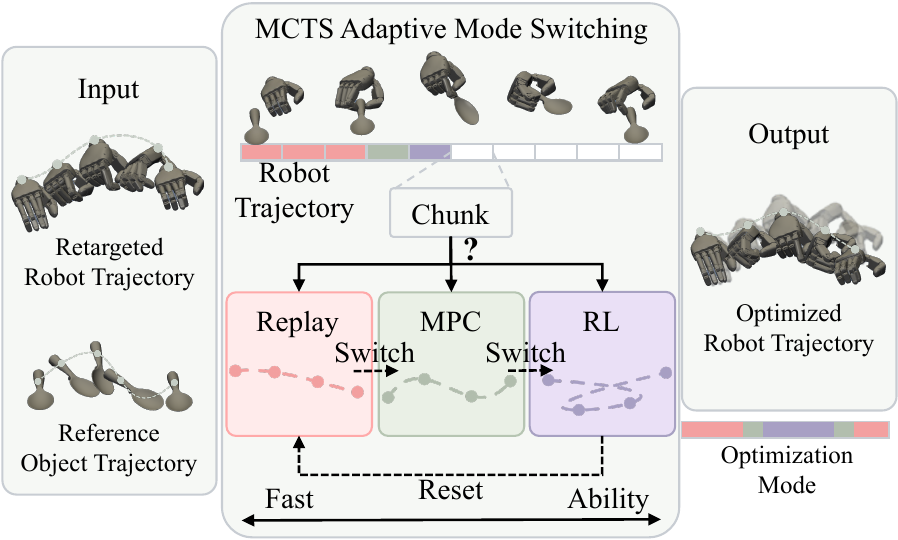}
    \caption{\textbf{Adaptive MCTS-style switching optimization.}
    EgoEngine progressively selects among Replay, MPC, and RL to solve each trajectory chunk with the cheapest feasible solver.}
    \label{fig:algorithm}
    \vspace{-2em}
\end{wrapfigure}

\subsubsection{Object-Centric Trajectory Optimization}
After retargeting, EgoEngine faces a long-horizon, high-DoF trajectory optimization problem: the reference trajectory preserves the human motion prior, but must be converted into executable robot actions that reproduce the demonstrated object motion.
Retargeting alone is insufficient for two reasons:
(1) replaying it often fails under the human-to-robot embodiment gap, including kinematic mismatch and contact-dynamics discrepancies;
and (2) human videos provide an observable proprioceptive trajectory rather than action commands, leaving a proprio-to-action gap for applying forces and sustaining contacts.
We refine the reference trajectory in simulation with an object-centric objective, using object motion extracted from human videos as the task-level target for executable robot actions.

Let $T_{o}^{t}$ denote the 6D object pose tracked from the human video (Sec.~\ref{sec:human_preprocess}), and let $\hat{T}_{o}^{t}$ denote the corresponding object pose in simulation under the robot's executed control.
We define the object pose tracking error as:
\begin{equation}
e^{t}
=
\sqrt{
\lambda_p\, d_p\!\left(\mathrm{trans}(\hat{T}_{o}^{t}),\, \mathrm{trans}(T_{o}^{t})\right)^2
+
\lambda_R\, d_R\!\left(\mathrm{rot}(\hat{T}_{o}^{t}),\, \mathrm{rot}(T_{o}^{t})\right)^2
},
\end{equation}
where $d_p(\cdot,\cdot)$ is the Euclidean distance in $\mathbb{R}^3$ and $d_R(\cdot,\cdot)$ is the geodesic distance on $\mathrm{SO}(3)$.
We use a threshold object-tracking objective with early termination: the episode terminates if $e^t$ exceeds a threshold $C$.
Within the valid regime ($e^t \le C$), the reward is $r_{\mathrm{obj}}^t=C-e^t$, so lower error gives higher reward while rewards remain non-negative until termination.
This object-centric reward is combined with auxiliary terms for contact, action smoothness, and human-mimetic regularization (see Appendix for details).

\textbf{Solver modes.}
EgoEngine decomposes the long-horizon trajectory into temporal chunks and optimizes each chunk progressively.
For each chunk, EgoEngine uses three solver modes with increasing optimization capability.
\emph{Replay} directly executes the reference robot trajectory.
\emph{MPC} searches over short-horizon action samples around the reference trajectory, providing local correction at a moderate cost.
\emph{RL} trains a hand residual policy for difficult chunks.
The policy observes the hand state, object pose, and reference retargeted command, and predicts a residual $\delta a_t \sim \pi_\phi(\cdot \mid s_t)$ added to the reference command, $a_t = a^{\mathrm{base}}_t + \delta a_t$.
We optimize $\pi_\phi$ with PPO~\cite{ppo,rsl-rl}, consistent with residual RL and hybrid control formulations~\cite{residualrl}.

\textbf{MCTS-style Adaptive Mode Switching.}
Unlike prior dexterous RL methods that learn universal or chained policies for long-horizon manipulation~\cite{seqdexterity}, often requiring extensive reward engineering~\cite{viral,eureka} or large-scale simulation~\cite{omnireset}, EgoEngine uses a Monte Carlo Tree Search (MCTS)-style mode-switching strategy~\cite{league}.
Here, MCTS-style refers to a lightweight, heuristic-based progressive search over solver modes, rather than a full MCTS algorithm.
Instead of applying the strongest solver to the full trajectory, in each chunk, EgoEngine starts from Replay and escalates to MPC or RL only when the current mode cannot produce a feasible or sufficiently improved rollout, as shown in Fig.~\ref{fig:algorithm}.
Replay is retained when the reference trajectory satisfies the object-centric criterion; otherwise, EgoEngine invokes MPC for local correction and falls back to RL for stronger refinement when MPC remains insufficient.
To avoid local minima from optimizing each chunk in isolation, EgoEngine uses a two-chunk optimization window, jointly solving the current and next chunks but executing only the current chunk once both are feasible.
This design improves efficiency by avoiding unnecessary refinement and eliminating the need to maintain an RL policy over the full trajectory.

\subsection{Visual Generation}
\label{visual_branch}

\textbf{Human removal.}
The raw egocentric frame $I_t^{(h)}$ shows the human arms and hands performing the task; these regions must be removed before compositing in a robot rendering.
We mask the arm-hand regions using the SAM2 masks from Sec.~\ref{sec:human_preprocess} and fill them with Inpaint-Anything v2~\cite{inpaintanything}, recovering the scene and object content occluded by the demonstrator.
This yields a demonstrator-free frame $\bar{I}_t$ retaining the scene and target object, onto which the robot is composited.

\textbf{Occlusion-aware blending.}
Given the robot trajectory from the action branch, we render the robot at the egocentric viewpoint to obtain $R_t$, and recover an occlusion-aware mask via two-pass differential rendering.
Keeping object geometry opaque in both passes, we render the scene with the robot fully transparent ($I_{\text{bg}}^t$) and with the robot opaque ($I_{\text{rob}}^t$);
The mask is computed per pixel $p$ from the thresholded RGB difference:
\begin{equation}
\tilde{M}_r^t(p) \;=\; \mathbf{1}\!\bigl[\,\|I_{\text{rob}}^t(p)-I_{\text{bg}}^t(p)\| > 0\,\bigr].
\end{equation}
Since objects are present in both passes, this implicitly removes the occluded robot pixels, yielding $\tilde{M}_r^t$ as the visible
robot mask.
The final generated observation is blended with:
\begin{equation}
\tilde{o}_t^{(r)} = \tilde{M}_r^t \odot R_t + (1-\tilde{M}_r^t)\odot \bar{I}_t .
\end{equation}

\subsection{Policy Distillation}
\label{policy_distillation}

After the action and visual branches, each human video is converted into a robot demonstration consisting of synchronized robot observation and action.
Aggregating these across videos yields a synthetic robot dataset
$\tilde{\mathcal{D}}_{\mathrm{robot}}=\{(\tilde{o},\tilde{a})\}$,
where $\tilde{o}$ denotes the generated robot observation from the visual branch with proprioception, and $\tilde{a}$ denotes the corresponding action from the action branch.
We train a visuomotor policy $\pi_\theta$ with HPT~\cite{hpt} on this dataset with an $\ell_2$ action regression loss:
\begin{equation}
\min_{\theta}\;
\mathbb{E}_{(\tilde{o},\tilde{a})\sim \tilde{\mathcal{D}}_{\mathrm{robot}}}
\left[
\|\pi_\theta(\tilde{o})-\tilde{a}\|_2^2
\right].
\end{equation}
The policy distills both branches into a closed-loop controller mapping robot observation to action.
\section{Experiments}
\label{sec:experiments}

\subsection{Experiment Setup}
\label{sec:exp_setup}

We structure our evaluation around three questions.
First, \textbf{are the generated robot observations visually aligned with real robot observations?}
We compare synthesized and real robot observations in appearance and representation space (Sec.~\ref{sec:visual}).
Second, \textbf{are the generated robot actions executable and task-aligned?}
We test whether they reproduce demonstrated object interactions in simulation on TACO and Aria (Sec.~\ref{sec:action}).
Third, \textbf{do the generated observation--action pairs support zero-shot policy learning?}
We train policies from EgoEngine-generated demonstrations and evaluate them on the real robot (Sec.~\ref{sec:downstream}).

\textbf{Datasets and settings.}
We use two data sources for evaluation.
TACO~\cite{liu2024taco} provides 2,500 video sequences for visual and action evaluation.
We collect the Aria dataset with Aria Gen2 Glasses~\cite{projectaria}, containing 200 real-world egocentric human videos across four tasks.
For comparison, we also collect 200 real-robot teleoperation demonstrations, which are not used by EgoEngine.
In simulation, we use a bimanual RB-Y1 with two 7-DoF arms and two 12-DoF XHands.
On the real robot, we use a single-arm RB-Y1 with one XHand.

\textbf{Tasks.}
In simulation, we evaluate on 16 selected demonstration pairs from TACO that are compatible with our robot embodiment and digital-twin reconstruction pipeline.
In the real world, we evaluate on four Aria tasks: \emph{Mustard} (placing a mustard bottle onto a target plate),
\emph{Drawer} (opening a drawer and placing a cube inside),
\emph{Hammer} (picking up a hammer and striking a nail),
and \emph{Flower} (holding a water bottle to a flower).
Together, these tasks cover diverse interaction patterns, including pick-and-place, tool use, and fine contact.

\textbf{Metrics.}
We use metrics matched to each evaluation question.
For visual fidelity, we report qualitative comparisons and feature-space alignment with real robot observations.
For action fidelity, we report success rate, object-tracking reward, and generation cost.
For downstream policy performance, we report real-robot success rate under fixed rollout trials.

\begin{figure*}[t]
\centering

\begin{minipage}[t]{0.56\textwidth}
    \centering
    \vspace{-1pt}
    \includegraphics[width=\linewidth]{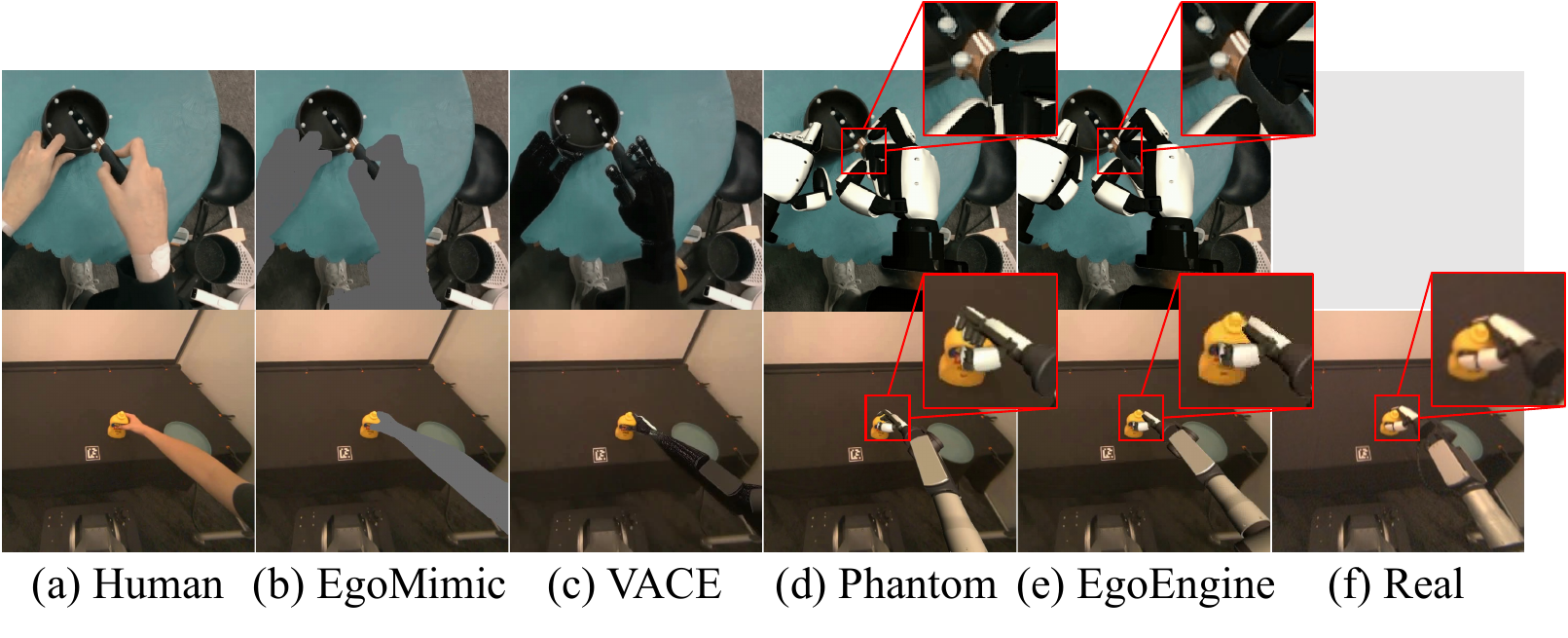}
    \captionof{figure}{Two samples from TACO and Aria Dataset. Qualitative visual comparison between (a) original human video, (b) EgoMimic inpainting, (c) VACE (WAN2.1), (d) Phantom, (e) EgoEngine (Ours), (f) Real teleoperation videos. Compared to (d), EgoEngine produces more physically consistent robot observations with the digital twin.}
    \label{fig:visual_comparison}
\end{minipage}
\hfill
\begin{minipage}[t]{0.43\textwidth}
    \centering
    \vspace{0pt}

    \includegraphics[width=1.0\linewidth]{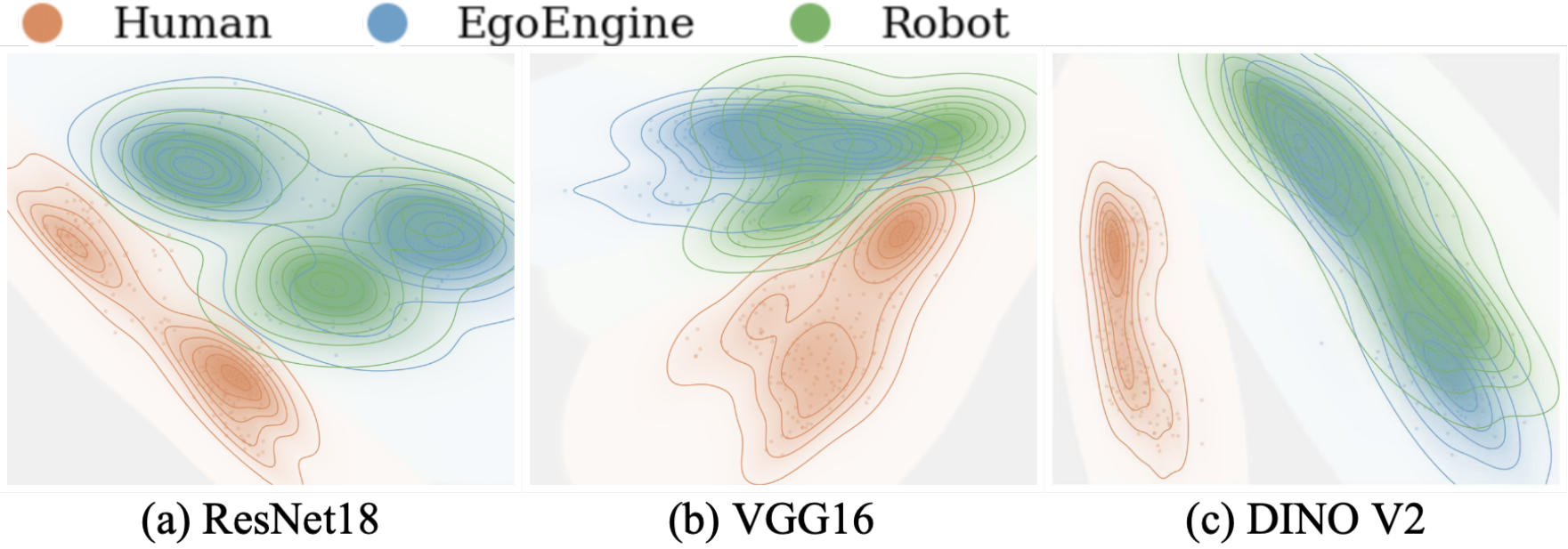}
    \vspace{-1.5em}
    
    {\captionsetup{type=figure}
    \captionof{figure}{KDE~\cite{kde} of features extracted from ResNet18, VGG16, and DINOv2.}
    \label{fig:visual_kde}
    }

    \vspace{0em}

    {\small
    \setlength{\tabcolsep}{4pt}
    \renewcommand{\arraystretch}{0.84}
    \begin{tabular}{lccc}
        \toprule
        Method (FD$\downarrow$) & ResNet18 & VGG16 & DINOv2 \\
        \midrule
        Human Video & 764.5 & 670.2 & 602.9 \\
        EgoMimic & 830.5 & 812.1 & 579.6 \\
        VACE & 713.6 & 745.3 & 488.0 \\
        Phantom & 620.0 & 650.8 & \textbf{470.6} \\
        \midrule
        EgoEngine & \textbf{614.7} & \textbf{644.2} & 473.1 \\
        \bottomrule
    \end{tabular}
    }

    \vspace{-0.6em}

    {\captionsetup{type=table}
    \captionof{table}{Fr\'echet Distance (FD) of feature similarity comparison across methods.}
    \label{tab:visual_fid}
    }
\end{minipage}

\vspace{-2em}

\end{figure*}

\subsection{Visual Fidelity of Generated Robot Observations}
\label{sec:visual}

We evaluate visual fidelity by measuring how well EgoEngine synthesizes robot egocentric observations that match real robot observations.
We compare EgoEngine with several representative human-to-robot video editing baselines:
(a) the original \textbf{human videos},
(b) \textbf{EgoMimic}~\cite{kareer2024egomimic}, which masks the human region toward a robot-like color,
(c) \textbf{VACE}~\cite{vace}, built on WAN2.1~\cite{wan}, a video generation model conditioned on video, text, and image template prompts,
and (d) \textbf{Phantom}~\cite{phantom}, which directly blends a rendered robot into the scene without modeling occlusion or physical contact.

\textbf{EgoEngine generates robot observations that are physically consistent and closely aligned with real robot observations.} 
We validate the conclusion through both qualitative comparisons and feature-space analysis.
Qualitative comparisons in Fig.~\ref{fig:visual_comparison} show that EgoEngine produces more physically consistent robot observations than the baselines, especially around robot--object contact and visibility ordering.
These examples evaluate appearance realism, physical consistency, and contact/occlusion synthesis under representative tasks.
We further analyze feature-space alignment with real robot observations using three fixed pretrained encoders: ResNet18~\cite{he2016resnet} (used as the policy encoder), VGG16~\cite{simonyan2015vgg}, and DINOv2~\cite{oquab2024dinov2}.
We extract per-frame features and compare the distributions of synthesized observations and real robot observations.
As shown in Fig.~\ref{fig:visual_kde}, EgoEngine features are closer to the real-robot feature distribution than the original human videos in KDE~\cite{kde} visualization.
Tab.~\ref{tab:visual_fid} quantifies this: EgoEngine reduces Fr\'echet Distance (FD) from the last layer across encoders versus baselines, except on DINOv2, where it matches Phantom (473.1 vs.\ 470.6).

\subsection{Action Fidelity of Generated Robot Trajectories}
\label{sec:action}
We evaluate whether EgoEngine can convert retargeted human motion into executable robot trajectories.
We compare EgoEngine with three action-generation modes:
(1) \textbf{Mink (Replay)}~\cite{mink}, which directly executes the retargeted reference trajectory,
(2) \textbf{Spider (MPC)}~\cite{spider}, which performs local trajectory optimization;
and (3) \textbf{H2S2R (RL)}~\cite{h2s2r}, which learns strong residual refinement.
We report success rate (\emph{SR}), normalized rollout progress before early termination (\emph{Step}), and normalized object-tracking reward relative to the perfect reference trajectory (\emph{Reward}).
We also measure generation cost with a device-agnostic metric, the average simulation steps required to optimize a single successful trajectory timestep (\emph{Cost}).

\textbf{EgoEngine generates executable, task-aligned robot trajectories.}
As shown in Tab.~\ref{tab:actionsim}, Replay is fast but often fails under embodiment mismatch and contact-rich dynamics.
MPC improves over Replay on some TACO trajectories, especially for pushing or reorientation motions, but remains unreliable for grasp-centric stages in the Aria tasks.
For Aria, these two only succeed for some Drawer demos.
RL achieves stronger raw performance, indicating that difficult contact-rich interactions require stronger trajectory refinement.
In contrast, EgoEngine preserves strong trajectory quality across both TACO and Aria, showing that it can convert retargeted human motion into executable and task-aligned robot trajectories.

\textbf{EgoEngine improves the quality--efficiency tradeoff through MCTS-style adaptive mode switching, especially for long-horizon tasks.}
As shown in Tab.~\ref{tab:actionsim}, EgoEngine preserves strong trajectory quality while reducing generation cost compared with full RL refinement.
The efficiency gain comes from assigning solver capacity only where it is needed: Fig.~\ref{fig:mode_switching_vis} shows that easier chunks are mainly handled by Replay and MPC, while difficult contact-rich chunks fall back to RL.
MPC is selected less frequently than Replay and RL, suggesting that local correction is useful for a subset of chunks but insufficient for many contact-rich stages.
In Fig.~\ref{fig:throughput}, EgoEngine improves Aria generation efficiency by 22.0\%, from 2.36 demos/hour with RL to 2.88 demos/hour on a single RTX 4090 without parallelization.
On TACO, a long-horizon bimanual dataset with harder and more complex tasks, the average trajectory length is 327.5 timesteps, which is 2.39$\times$ that of Aria; EgoEngine therefore achieves larger efficiency gains by avoiding full-trajectory refinement when cheaper solvers are sufficient.
Together, these results show that EgoEngine preserves strong trajectory quality while substantially improving generation efficiency, especially as task horizon and complexity increase.

\begin{table*}[t]
\centering
\small

\begin{minipage}[t]{0.65\textwidth}
\centering
\vspace{0pt}
\setlength{\tabcolsep}{4.5pt}
\resizebox{\linewidth}{!}{%
\begin{tabular}{lcccc@{\hspace{1.4em}}cccc}
\toprule
& \multicolumn{4}{c}{TACO} & \multicolumn{4}{c}{Aria} \\
\cmidrule(lr){2-5}\cmidrule(lr){6-9}
Method 
& SR$\uparrow$ & Step$\uparrow$ & Reward$\uparrow$ & Cost$\downarrow$
& SR$\uparrow$ & Step$\uparrow$ & Reward$\uparrow$ & Cost$\downarrow$\\
\midrule
Mink~\cite{mink} & 0.17 & 0.29 & 0.29 & \textbf{1.00} & 0.10 & 0.66 & 0.62 & \textbf{1.00} \\
Spider~\cite{spider}  & 0.25 & 0.42 & 0.39 & 7,923 & 0.20 & 0.69 & 0.65 & 4,382\\
H2S2R~\cite{h2s2r}  & \textbf{0.83} & \textbf{0.86}  & \textbf{0.70} & 73,675 & \textbf{0.90} & \textbf{0.94} & \textbf{0.85} & 20,237 \\
\midrule
EgoEngine  & \textbf{0.83} & 0.84   & 0.67   & 34,842  & \textbf{0.90} & 0.91 & 0.83 & 16,560\\
\bottomrule
\end{tabular}
}
\captionof{table}{Simulation results of action generation on TACO (16 demonstration pairs) and Aria (4 tasks). Metrics: success rate (SR), rollout completion step ratio (Step), object-tracking reward ratio vs.\ perfect trajectories (Reward), and average simulation steps per trajectory timestep (Cost).}
\label{tab:actionsim}
\end{minipage}
\hfill
\begin{minipage}[t]{0.34\textwidth}
\centering
\vspace{0pt}
\includegraphics[width=\linewidth]{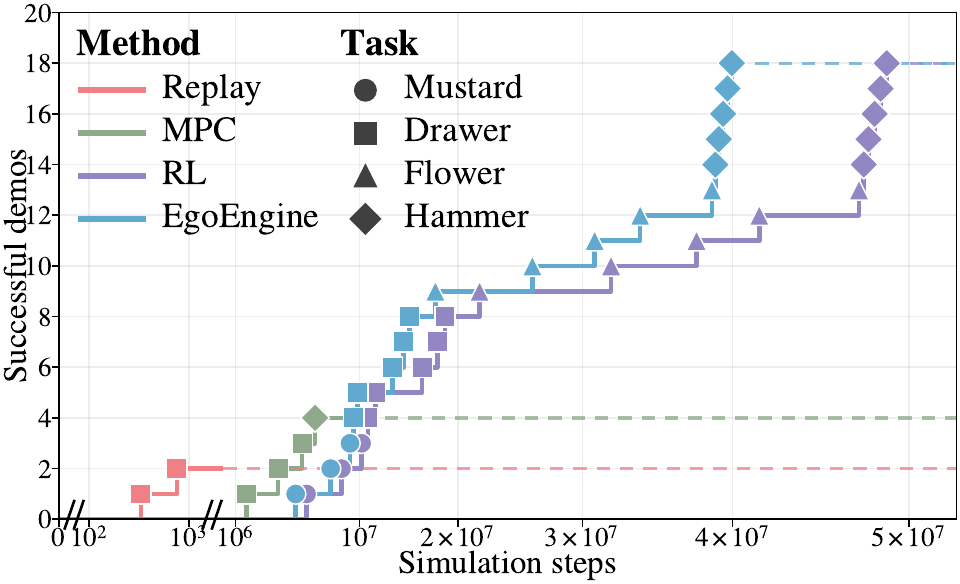}
\vspace{-1.6em}
\captionof{figure}{Throughput comparison: Successful demo number generated over simulation steps on 20 Aria demos across 4 tasks.}
\label{fig:throughput}
\end{minipage}

\vspace{0.5em}

\begin{minipage}[t]{0.53\textwidth}
    \centering
    \vspace{0pt}
    \includegraphics[width=\linewidth]{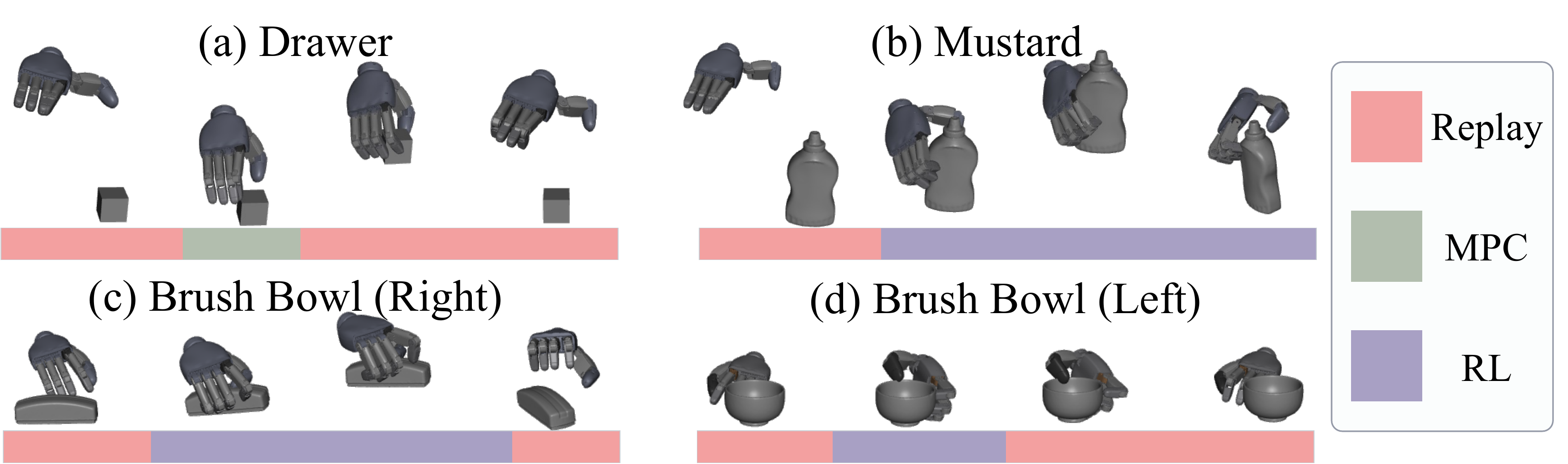}
    \captionof{figure}{Visualization of Adaptive MCTS-style mode switching across trajectory chunks on four demonstrations from the Aria and TACO datasets.}
    \label{fig:mode_switching_vis}
\end{minipage}
\hfill
\begin{minipage}[t]{0.46\textwidth}
    \centering
    \vspace{0pt}
    \resizebox{\linewidth}{!}{%
    \begin{tabular}{lcccc}
        \toprule
        Method & Mustard & Drawer & Flower & Hammer \\
        \midrule
        Human Video & 0.00 & 0.10 & 0.00 & 0.00 \\
        Phantom~\cite{phantom} & 0.00 & 0.05 & 0.00 & 0.00 \\
        Real Robot & 
        \textbf{0.80} & \textbf{0.80} & \textbf{0.70} & 0.25 \\
        \midrule
        EgoEngine & 0.40 & 0.35 & \textbf{0.70} & \textbf{0.60} \\
        \bottomrule
    \end{tabular} 
    }
    \captionof{table}{Policy success rate (SR) on four Aria tasks, comparing EgoEngine with Human Videos (direct retargeting), Phantom, and real robot teleoperation.}
    \label{tab:realrobot_succ}
\end{minipage}

\vspace{-1em}
\end{table*}

\subsection{Downstream Policy Distillation}
\label{sec:downstream}

For each Aria task, we train the visuomotor policy from EgoEngine-generated data and evaluate it on the real robot.
Tab.~\ref{tab:realrobot_succ} compares EgoEngine with (1) \textbf{Human videos}, (2) \textbf{Phantom}~\cite{phantom}, which performs video editing during both training and inference, and (3) \textbf{Real robot} teleoperation demonstrations.
All policies share the same architecture, demonstration count, and evaluation protocol.

\begin{wrapfigure}{r}{0.6\columnwidth}
    \vspace{-2em}
    \centering
    \includegraphics[width=\linewidth]{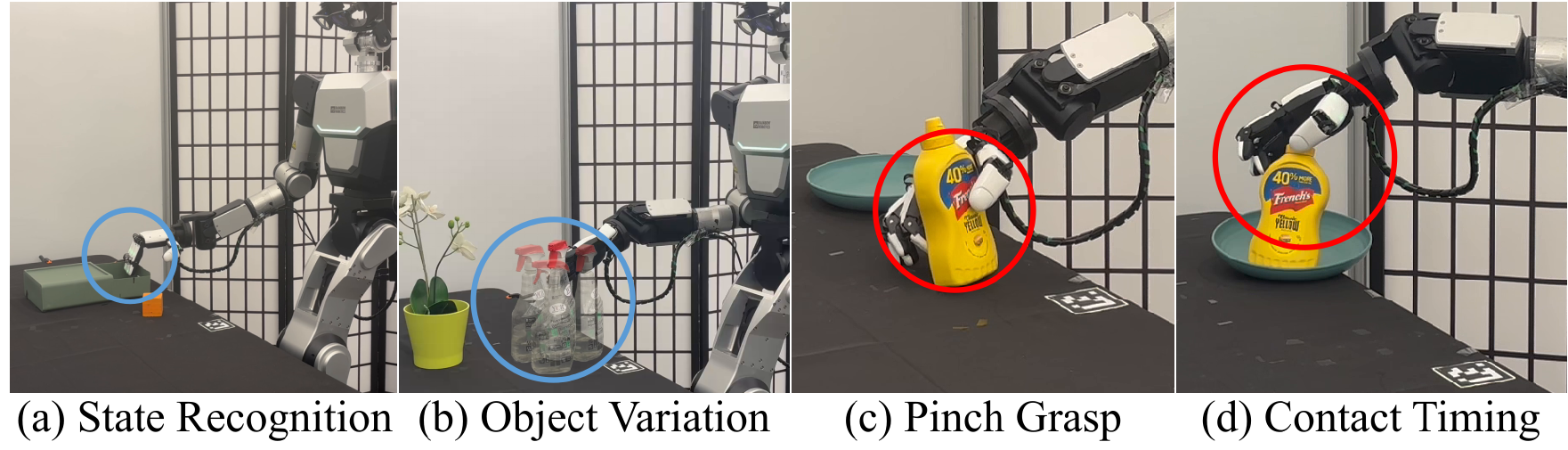}
    \vspace{-1.5em}
    \caption{
    Behavioral strengths and failure modes of EgoEngine. The policy handles (a) state recognition/recovery and (b) object variation; failures are mainly (c) unstable pinch grasps and (d) contact-timing mismatch.}
    \label{fig:behavior_analysis}
    \vspace{-1.5em}
\end{wrapfigure}

\textbf{EgoEngine-generated demonstrations support zero-shot real-robot policy learning from egocentric human videos.}
Human videos and Phantom videos achieve near-zero success and mostly fail due to grasping pose, indicating that visual conversion alone is insufficient for dexterous policy learning.
In contrast, EgoEngine achieves non-trivial zero-shot performance across tasks and matches or exceeds real robot demonstrations on 2 of 4 tasks.
Fig.~\ref{fig:behavior_analysis} (a, b) provide two example behaviors supporting this result.
For \emph{Hammer}, Real Robot reaches a low SR, as unintended early contacts often disturb the hammer before grasping. EgoEngine instead refines the grasp with a slight wrist rotation, leading to a more stable hammer grasp.
The learned policy can recognize task state, recover from intermediate failures, and adapt to object pose variations, suggesting that EgoEngine-generated demonstrations provide usable supervision for real-robot visuomotor policy learning.

\begin{wraptable}{r}{0.30\linewidth}
\centering
\vspace{-1em}
\small
\begin{tabular}{lc}
\toprule
Method & SR $\uparrow$ \\
\midrule
Human Videos & 0.03 \\
+Visual branch   & 0.05 \\
+Action branch   & 0.43 \\
\midrule
EgoEngine    & \textbf{0.51} \\
\bottomrule
\end{tabular}
\caption{Ablation of the visual and action branches on downstream success rate (averaged over four Aria tasks).}
\label{tab:downstream_ablation}
\vspace{-2em}
\end{wraptable}
\textbf{Executable action generation provides the primary improvement, while visual generation provides an additional gain.}
We further perform an ablation study over four settings on all four Aria tasks: Human videos, visual branch only (+Visual branch), action branch only (+Action branch), and full EgoEngine, to isolate how visual and action generation contribute to policy performance.
Tab.~\ref{tab:downstream_ablation} shows that removing the action branch causes the largest performance drop, leaving only a few successful trials on Drawer.
This highlights the importance of action generation: without action refinement, the retargeted proprioceptive trajectory either fails to apply sufficient contact forces for grasping or produces incorrect object orientations after grasping, resulting in poor object interactions for policy learning.
Removing the visual branch still preserves a substantial SR, with only a small drop from full EgoEngine.
This trend is consistent with other works~\cite{motiontrans}, where policies can tolerate moderate embodiment appearance mismatch and the visual encoder could be object-centric.
Fig.~\ref{fig:behavior_analysis}(c,d) further shows that the remaining failures are mainly action-side errors, such as unstable grasps for Mustard and mistimed Drawer contacts, where the contact reward encourages premature contact or late release.
\section{Conclusion}
We present EgoEngine, a scalable data engine that converts egocentric human videos into paired robot demonstrations, consisting of executable action trajectories and aligned egocentric robot observations. 
By jointly grounding action generation and visual synthesis through object-centric reconstruction and physically feasible optimization, EgoEngine turns egocentric human videos into training data suitable for dexterous robot learning. Across simulation, visual comparison, and real-robot policy distillation, our results show that EgoEngine produces high-quality demonstrations that enable zero-shot policy learning on the real robot.

\section{Limitations}

EgoEngine still has limitations along three axes.
(1) \textbf{Quality.} EgoEngine already generates demonstrations that support zero-shot downstream policy learning, but the visual branch currently uses blending-based synthesis rather than fully learned photorealism, while action generation can still suffer from contact modeling errors and the sim-to-real gap.
(2) \textbf{Scalability.} EgoEngine scales with human video collection, but building the digital twin remains a bottleneck.
Obtaining high-quality object assets, estimating object states under severe occlusion, and handling deformable objects remain challenging.
(3) \textbf{Efficiency.} Simulation-based trajectory optimization is still slow at very large scale, though trajectories run in parallel; future work could use pretrained models to speed it up.

\clearpage
\bibliography{references}

\newpage

\appendix
\counterwithin{figure}{section}
\counterwithin{table}{section}
\numberwithin{equation}{section}
\section*{Appendix}

\section{Hardware and Data Preprocess}
\label{sec:appendix_hardware}

\begin{figure*}[h]
  \centering
  \includegraphics[width=0.9\linewidth]{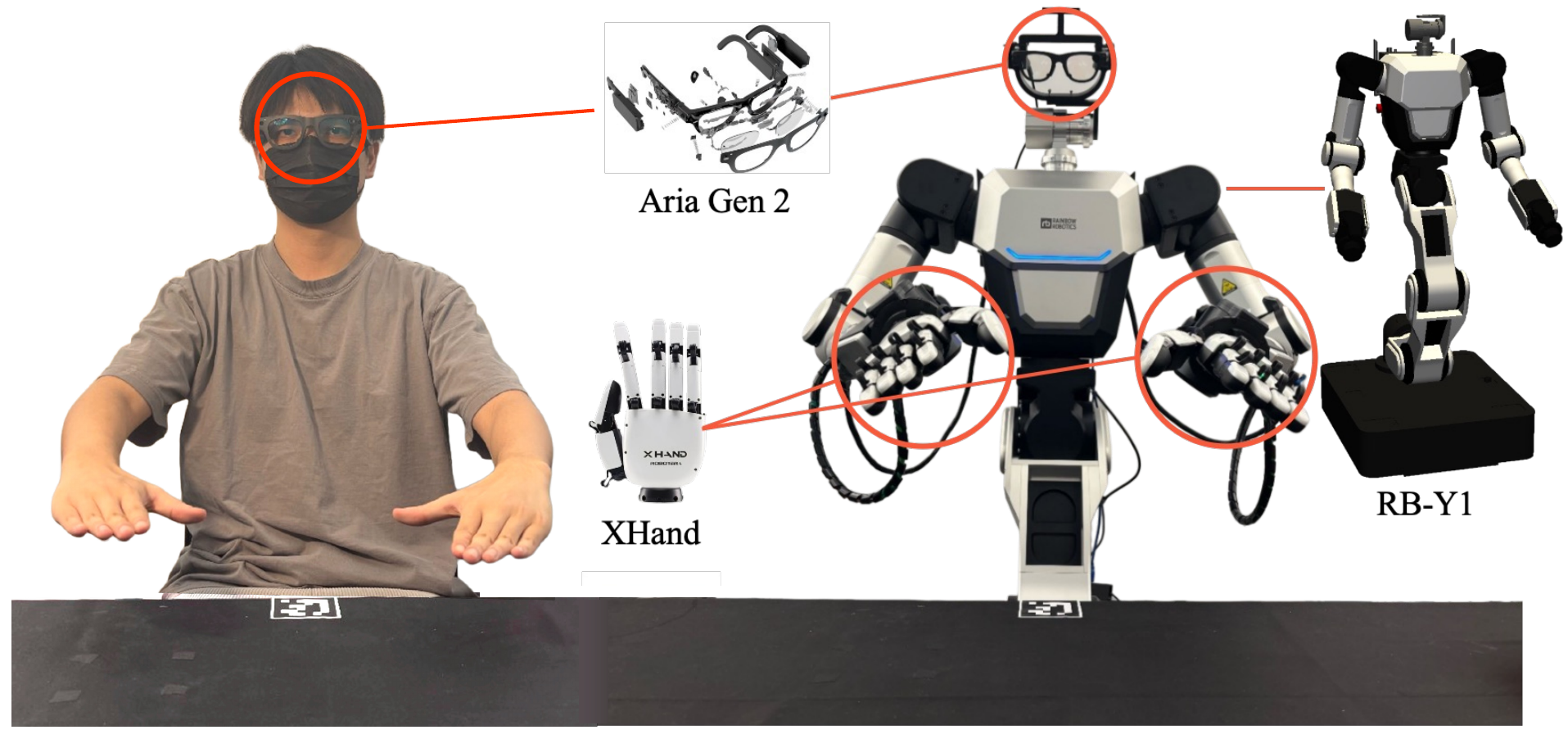}
  \caption{\textbf{System overview.} We use Aria Gen2 for egocentric perception for both human and robot. 
  Then we retarget demonstrations to the RB-Y1 humanoid equipped with XHands (bimanual in simulation, single-arm with one XHand on the real robot) for dexterous manipulation.}
  \label{fig:system_overview}
\end{figure*}

The overall data collection and robot system is illustrated in Fig.~\ref{fig:system_overview}.
A humanoid robot RB-Y1 equipped with XHand dexterous hands is used in our experiments to minimize the action gap. 
We use a bimanual setup with two arms and two XHands in simulation, and a single-arm setup with one XHand for the real-robot experiments.
We collect human demonstration videos with Aria Gen2 to establish a comprehensive dataset for our study.
To minimize the visual gap between human and robot demonstrations, we mount the same glasses on the robot for egocentric view perception. 

\noindent\textbf{Aria Gen2 Glasses.}
Aria Gen2 is a next-generation research device in the Project Aria program~\cite{projectaria}. 
It serves as an egocentric perception platform to record human demonstrations from the wearer’s viewpoint, and is then used for robot egocentric perception, providing state-of-the-art hand tracking, SLAM, and depth estimation results.

\noindent\textbf{RB-Y1 Humanoid Robot.}
RB-Y1 is a full-body humanoid robot that we use as the target embodiment for executing the retargeted demonstrations.
It provides a whole-body manipulation platform with a torso and two arms, enabling bimanual interaction with everyday objects while maintaining stable body motion.
In our system, we command the robot through a standard low-level controller interface (e.g., joint-space position/velocity control) and execute trajectories produced by our retargeting module.

\noindent\textbf{XHand.}
XHand is a 12 DoF dexterous robotic hand, with 3 DoF on each of the thumb and index fingers, and 2 DoF on each of the remaining fingers.
We mount it on the RB-Y1 end-effector to enable better fine-grained object interaction.

\subsection{Camera Calibration}
\label{calibration}

For the Aria dataset, we use AprilTag-based calibration to accurately estimate camera-to-robot extrinsics during data collection and deployment.
For the TACO dataset, we directly estimate a pseudo robot base from the object position.
In limited-data real-world dexterous manipulation, such calibration is a practical component for building a high-fidelity and executable human-to-robot transfer pipeline. 
By explicitly aligning the observation and robot frames, it improves sample efficiency and reduces optimization effort spent compensating for viewpoint mismatch. 
We therefore treat calibration as a standard system design choice, rather than a substantive assumption of the proposed method.

\textbf{Aria: Glasses to Tag to Robot.} 
In our real robot experiments with Aria dataset, we use an AprilTag placed on the workspace table in Fig.~\ref{fig:system_overview} as a common anchor to relate the Aria Gen2 coordinate system used by human recordings to the robot base frame used for action alignment.
Because the Aria worn by the human and the Aria mounted on the robot are physically different frames, we denote them as $\text{aria}_h$ and $\text{aria}_r$, respectively, and write the rigid transformation from the robot-mounted Aria frame to the robot base frame as
${}^{\text{base}}\mathbf{T}_{\text{aria}_r}$.

For human recordings, Aria Gen2 provides hand pose and object pose trajectories in the Aria coordinate system, denoted as ${}^{\text{aria}_h}\mathbf{T}_{\mathrm{hand}}^{t}$ and ${}^{\text{aria}_h}\mathbf{T}_{o}^{t}$.
Since the AprilTag is visible in the Aria stream, we detect the tag and estimate its pose in the Aria frame, ${}^{\text{aria}_h}\mathbf{T}_{\text{tag}}$.
We then convert all trajectories into the tag coordinate system:
\begin{equation}
    {}^{\text{tag}}\mathbf{T}_{x}^{t}
    =
    \bigl({}^{\text{aria}_h}\mathbf{T}_{\text{tag}}\bigr)^{-1}
    {}^{\text{aria}_h}\mathbf{T}_{x}^{t},
    \qquad
    x \in \{o, \mathrm{hand}\}.
\end{equation}

For robot data generation, Aria glasses are rigidly mounted on the robot's head with a fixed extrinsic relative to the head link. Because the head moves with the robot body, the transform from the Aria frame to the robot base depends on the robot configuration and is obtained through forward kinematics.
The robot-mounted Aria observes the same AprilTag, yielding ${}^{\text{aria}_r}\mathbf{T}_{\text{tag}}$, and forward kinematics provides the pose of the Aria device in the robot base frame, ${}^{\text{base}}\mathbf{T}_{\text{aria}_r}(q_t)$, where $q_t$ is the robot configuration.
Combining these yields the tag pose in the robot base frame:
\begin{equation}
    {}^{\text{base}}\mathbf{T}_{\text{tag}}
    =
    {}^{\text{base}}\mathbf{T}_{\text{aria}_r}(q_t)
    {}^{\text{aria}_r}\mathbf{T}_{\text{tag}}.
\end{equation}
With ${}^{\text{base}}\mathbf{T}_{\text{tag}}$ and the human trajectories expressed in the tag frame, we obtain their robot-base representation:
\begin{equation}
    {}^{\text{base}}\mathbf{T}_{x}^{t}
    =
    {}^{\text{base}}\mathbf{T}_{\text{tag}}
    {}^{\text{tag}}\mathbf{T}_{x}^{t},
    \qquad
    x \in \{o, \mathrm{hand}\}.
\end{equation}

As the dataset scales up, we can avoid per-sequence AprilTag calibration by relying on a fixed workcell setup and the robot's kinematic calibration, making explicit tag-based alignment unnecessary in large-scale collection.

\textbf{TACO: Objects to Robot.}
For TACO, we define the robot base from the scene geometry by taking the center of the tool and target objects and applying a fixed $0.6\,\mathrm{m}$ offset.
Since TACO is specified in an object-centric manner, this rule provides a consistent way to instantiate the robot pose across episodes while preserving executability. 
We set the table height to $0.72\,\mathrm{m}$, matching the real-world setup, so that the resulting configuration remains aligned with the physical system.

\subsection{Robot Teleoperation}
\label{teleoperation}

\begin{wrapfigure}{r}{0.42\textwidth}
    \centering
    \vspace{-4\baselineskip}
    \includegraphics[width=0.40\textwidth]{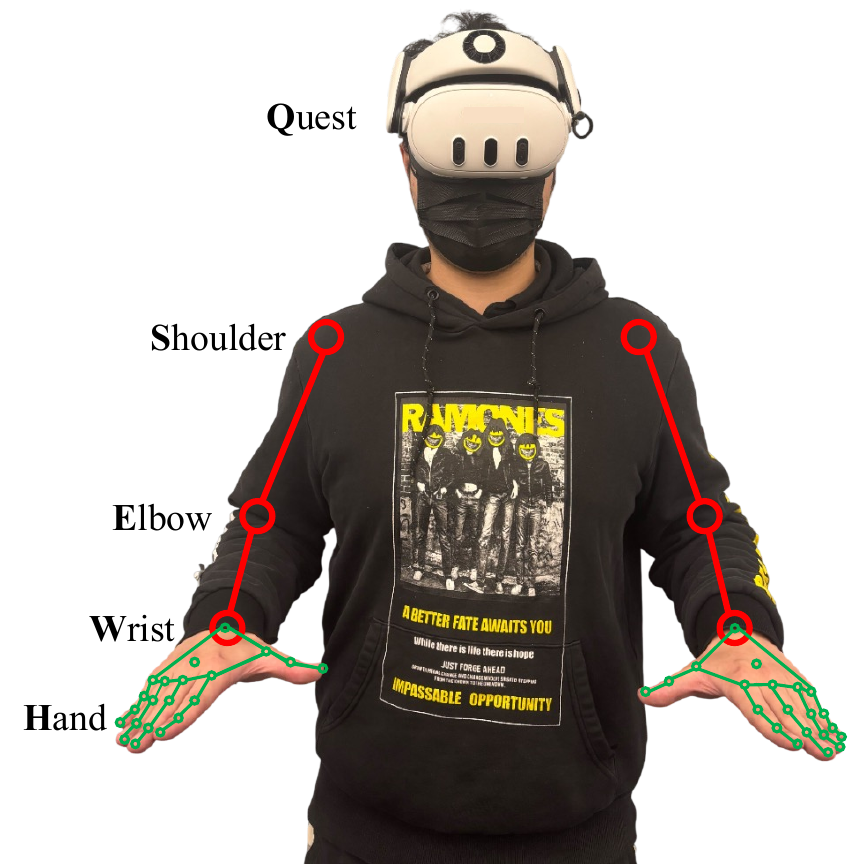}
    \caption{Robot teleoperation setup. We use Meta Quest for visual teleoperation.}
    \label{fig:teleoperation}
    \vspace{-2\baselineskip}
\end{wrapfigure}

We collect real robot teleoperation demonstration data for reference and sample-efficiency comparison in the main paper.
In Sec.~\ref{data_quality}, we compare the sample efficiency of EgoEngine-generated data with real-robot teleoperation data.

We use the state-of-the-art teleoperation system for the RB-Y1 robot, SEW~\cite{sewmimic}, for whole-body arm retargeting, and Open-TeleVision (OpenTV)~\cite{opentv} for dexterous hand teleoperation, as shown in Fig.~\ref{fig:teleoperation}. 
For the upper body, SEW provides an analytical retargeting method, which we use to map the operator's \textbf{S}houlder, \textbf{E}lbow, and \textbf{W}rist motion to the robot arm in real time. 
The human motion observations are streamed from the Aria glasses to the Meta Quest, and the resulting retargeted commands are used for robot whole-body control. 
For the hand, following OpenTV, we retarget the operator's \textbf{H}and motion to the XHand by using the vectors from the palm to the five fingertips as the hand action representation.
We represent the RB-Y1 arm as a Cartesian wrist pose and the XHand as joint positions.
This design is sufficient for our data collection pipeline and provides a practical interface for recording teleoperated demonstrations.

\subsection{Digital Twin Process}
With an RGB video collected from Aria Gen2 glasses, we first apply FoundationStereo~\cite{foundationstereo} to estimate per-frame depth maps. 
Given the resulting RGB-D observations, we then apply FoundationPose~\cite{wen2024foundationpose} together with the ground-truth object mesh to recover the object 6D pose trajectory over time. 

\begin{figure}[t]
    \centering
    \includegraphics[width=\linewidth]{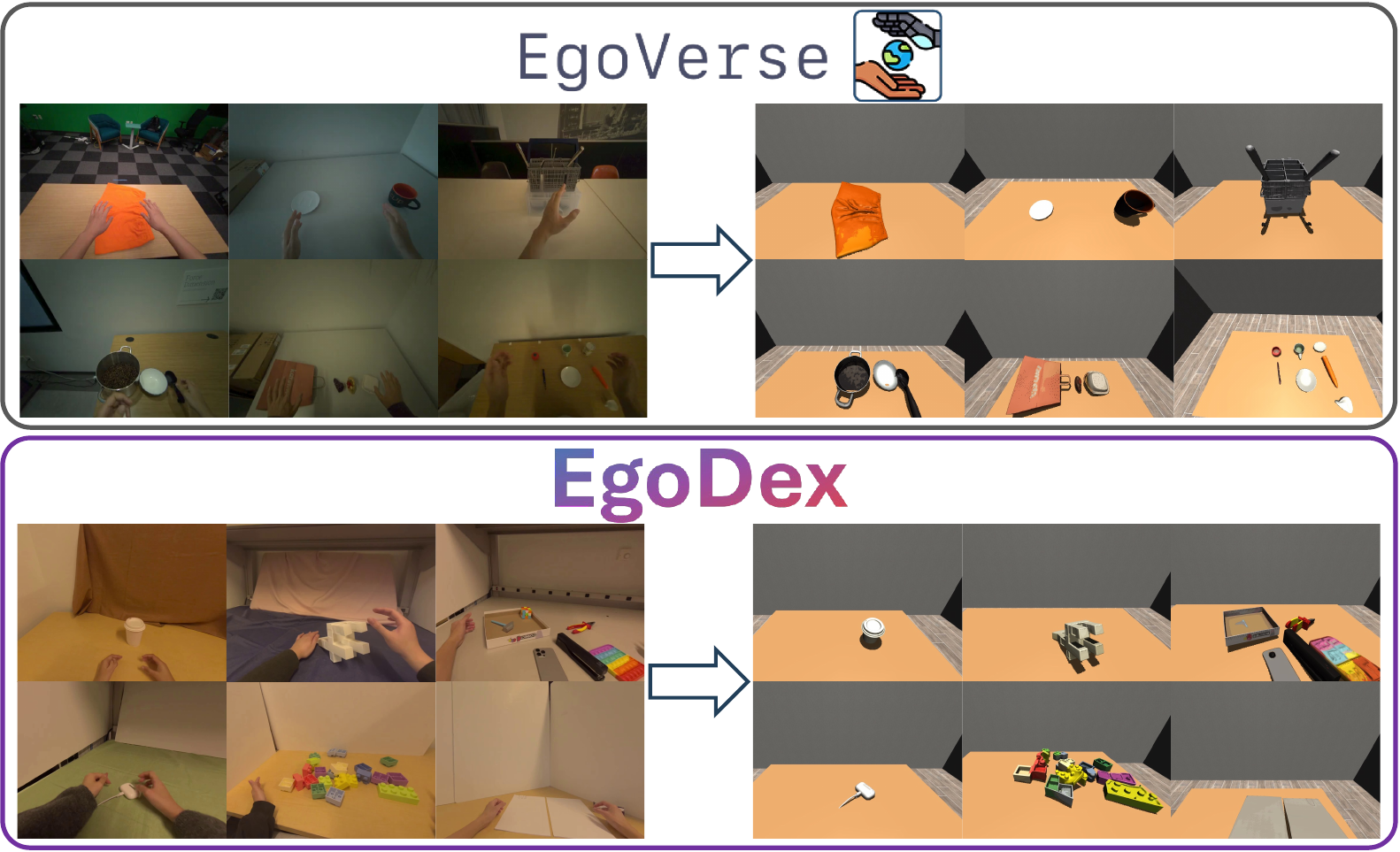}
    \caption{Digital twin reconstruction examples for EgoDex and EgoVerse datasets. The system converts egocentric manipulation videos into simulation scenes covering diverse objects, layouts, and viewpoints.}
    \label{fig:egodexverse}
\end{figure}

For Aria-collected data, we align the reconstructed scene to the simulator using an AprilTag attached in the real scene, which establishes the transformation between the observation frame and the simulator frame for digital twin reconstruction. 
For TACO data, since no AprilTag is available, we instead estimate the robot base frame using a fixed offset heuristic and reconstruct the scene under this approximated alignment. 

\begin{figure*}[p]
    \centering

    \begin{subfigure}[t]{0.98\textwidth}
        \centering
        \includegraphics[width=\linewidth]{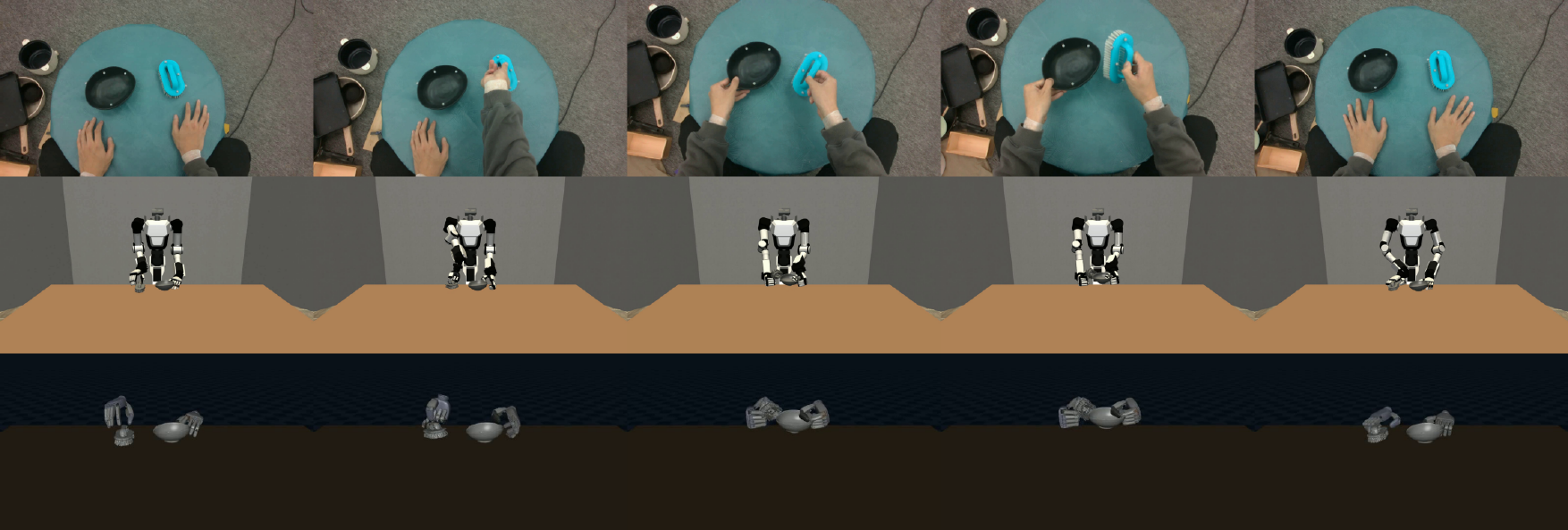}
        \caption{\scriptsize Brush Bowl with Brush.}
        \label{fig:dt_a}
    \end{subfigure}
    \hfill
    \begin{subfigure}[t]{0.98\textwidth}
        \centering
        \includegraphics[width=\linewidth]{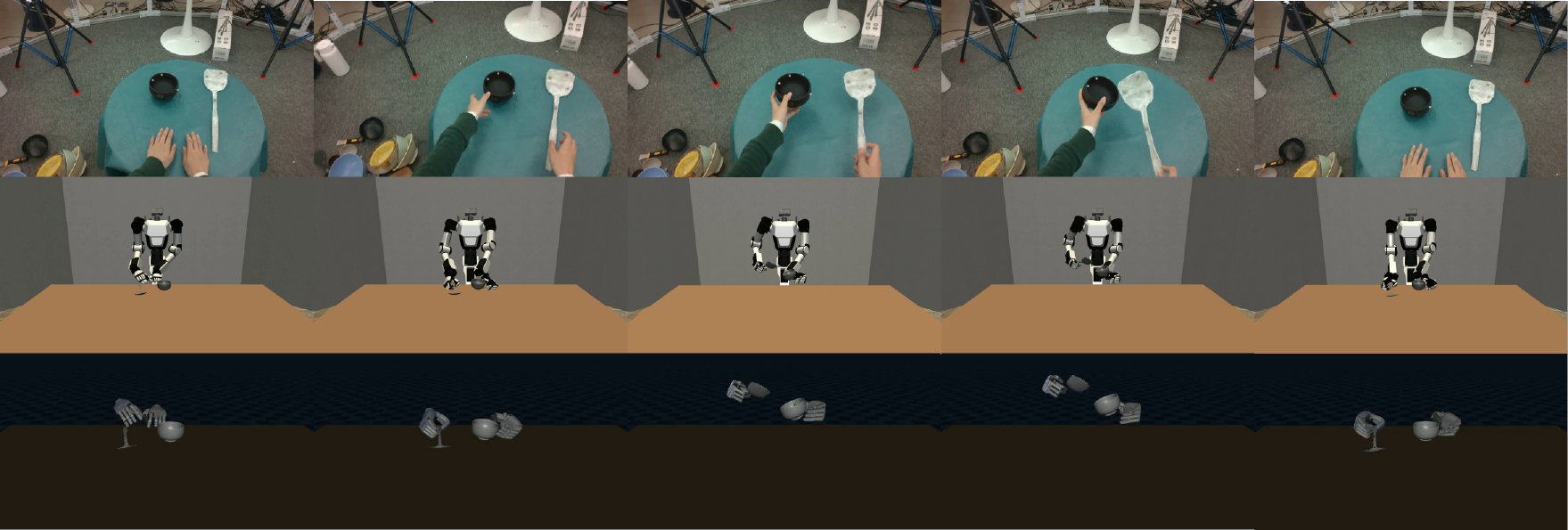}
        \caption{\scriptsize Cut Bowl with Spatula.}
        \label{fig:dt_b}
    \end{subfigure}
    \hfill
    \begin{subfigure}[t]{0.98\textwidth}
        \centering
        \includegraphics[width=\linewidth]{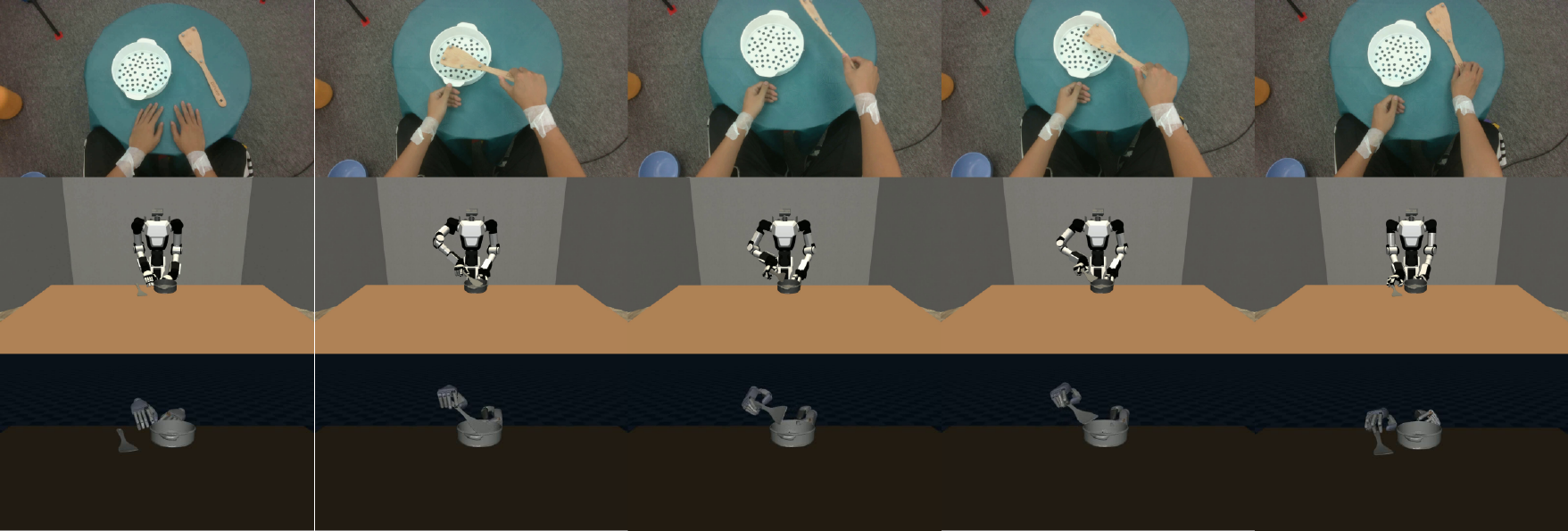}
        \caption{\scriptsize Skim off with Spatula on Plate.}
        \label{fig:dt_c}
    \end{subfigure}
    \hfill
    \begin{subfigure}[t]{0.98\textwidth}
        \centering
        \includegraphics[width=\linewidth]{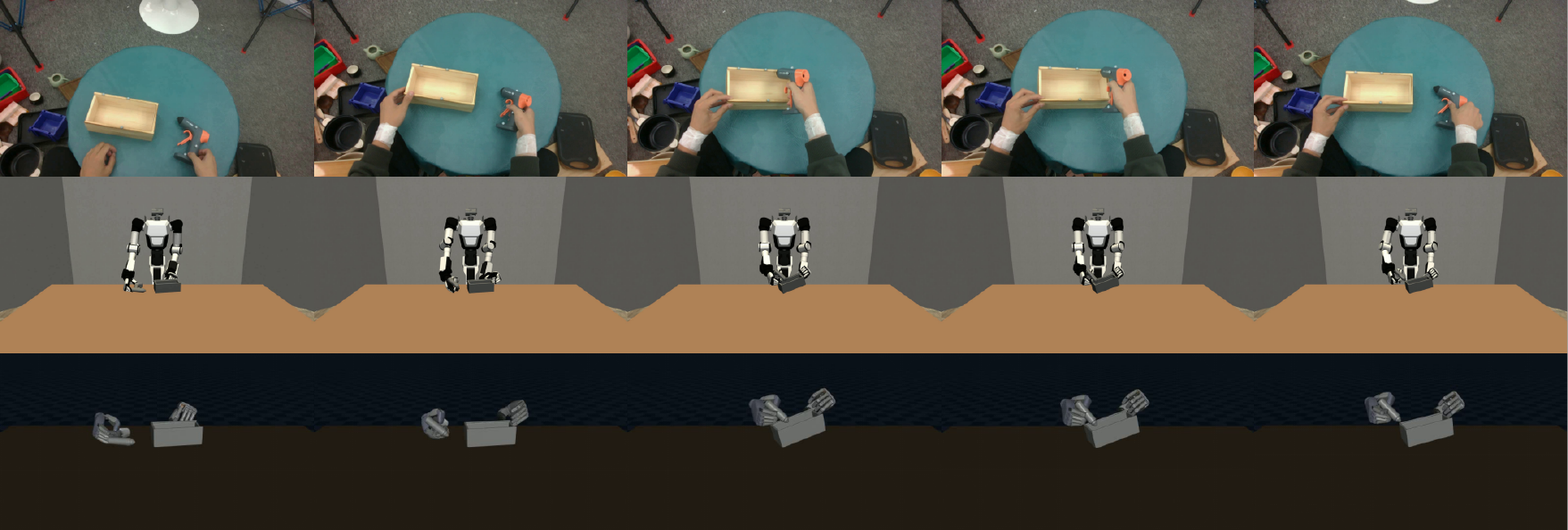}
        \caption{\scriptsize Smear on Box with Glue Gun.}
        \label{fig:dt_d}
    \end{subfigure}

    \caption{Digital twin reconstruction from sample TACO datasets. 
    Starting from TACO human demonstrations, we first recover the human hand pose together with the task-relevant scene entities, including the action, active object, and other passive objects in the scene (first row). 
    We then perform MINK-based retargeting to convert the recovered human hand motion into robot-executable hand trajectories (second row). 
    Finally, we reconstruct a digital twin that preserves the task scene, where the action branch uses a floating Cartesian wrist/base control abstraction with the same XHand for interaction and downstream RL training (third row). 
    (a) Brush bowl. (b) Cut with spatula. (c) Skim off to plate. (d) Smear on box.}
    \label{fig:digital_twin_tasks}
\end{figure*}

After initialization, the object pose trajectory estimated by FoundationPose is propagated into the simulator frame to drive digital twin reconstruction across the full sequence. 
Some qualitative examples of reconstructed digital twins on TACO tasks are shown in Fig.~\ref{fig:digital_twin_tasks}, including brush bowl, cut with spatula, skim off to plate, and smear on box. 
Notably, because TACO does not provide AprilTag-based calibration, these examples rely on the approximated base-frame alignment together with purely vision-based object pose estimation.

\subsection{Scalability of Digital Twin Reconstruction}
EgoEngine relies on a digital twin to bridge egocentric human videos and robot-executable demonstrations.
Given an input egocentric video, the digital twin reconstruction module builds a simulation scene that preserves the task-relevant geometry, object layout, and camera viewpoint of the original interaction.
Importantly, while digital twin reconstruction currently remains a practical requirement of EgoEngine, this stage can be increasingly automated: recent 3D foundation models such as SAM3D~\cite{sam3d} can be used to segment and reconstruct task-relevant objects from egocentric observations, reducing the need for manual scene modeling or object annotation.
This makes digital twin construction a scalable system component rather than a fixed manual bottleneck.
The reconstructed digital twin then provides the common grounding space for both branches of EgoEngine: the visual branch renders the robot into the aligned scene, while the action branch optimizes robot trajectories against the object-centric task motion.

To demonstrate the scalability of this reconstruction stage, we apply the digital twin reconstruction pipeline to egocentric manipulation videos from EgoVerse~\cite{egoverse} and EgoDex~\cite{egodex}.
These datasets contain diverse objects, viewpoints, and contact-rich hand-object interactions, making them suitable testbeds for assessing whether the digital twin construction process can generalize beyond manually curated scenes.
Our results show that EgoEngine can recover task-relevant scene structure and object assets with sufficient fidelity to support downstream robot visual generation and action optimization.

\textbf{EgoDex} is a large-scale dataset and benchmark for egocentric dexterous manipulation collected with ARKit on Apple Vision Pro.
The dataset has 829 hours of 30 Hz 1080p egocentric video with paired 3D pose annotations for the head, upper body, and hands as well as natural language annotation.
It consists entirely of active tabletop manipulation across 194 diverse tasks.

\textbf{EgoVerse} is a collaborative platform for
human data--driven robot learning that unifies data collection,
processing, and access under a shared framework, enabling
contributions from individual researchers, academic labs, and
industry partners.
The current release includes 1,362 hours (80k
episodes) of human demonstrations spanning 1,965 tasks, 240
scenes, and 2,087 unique demonstrators, with standardized formats, manipulation-relevant annotations, and tooling for downstream learning.

In Fig. A.3, we show 12 examples from EgoDex and EgoVerse to illustrate the potential of scaling digital twin reconstruction to diverse egocentric manipulation videos.
These examples cover different objects, layouts, viewpoints, and hand-object interactions, showing that the reconstruction stage is not limited to manually prepared scenes or a fixed set of task objects.
By preserving task-relevant geometry and egocentric scene structure, these digital twins provide the shared representation needed by EgoEngine's visual and action branches.
This demonstrates the potential of scaling EgoEngine to large egocentric video datasets, where diverse human manipulation videos can be automatically lifted into simulation for robot demonstration generation.

\section{Qualitative Results for Visual Branch}
\label{sec:supp_visual_qualitative}

\begin{figure}[p]
\centering
\setlength{\tabcolsep}{0.9pt}
\renewcommand{\arraystretch}{0.98}
\setlength{\fboxsep}{0pt}
\setlength{\fboxrule}{0.4pt}

\newcolumntype{C}[1]{>{\centering\arraybackslash}m{#1}}

\newlength{\datasetcolw}
\newlength{\taskcolw}
\newlength{\imgcolw}

\setlength{\datasetcolw}{0.95cm}      
\setlength{\taskcolw}{1.20cm}         
\setlength{\imgcolw}{0.178\textwidth} 

\newcommand{\datasetlabeltaco}{%
  \parbox[c][7.15cm][c]{\datasetcolw}{\centering
    \vspace*{10mm}%
    \rotatebox[origin=c]{90}{\fontsize{12}{14}\selectfont\bfseries TACO}%
  }%
}

\newcommand{\datasetlabelaria}{%
  \parbox[c][4.95cm][c]{\datasetcolw}{\centering
    \vspace*{10mm}%
    \rotatebox[origin=c]{90}{\fontsize{12}{14}\selectfont\bfseries Aria}%
  }%
}

\newcommand{\tasklabel}[1]{%
  \rotatebox[origin=c]{90}{%
    \parbox[c]{2.15cm}{\centering\fontsize{9.0}{10.0}\selectfont #1}%
  }%
}

\newcommand{\img}[1]{%
  \includegraphics[width=\linewidth]{#1}%
}

\newcommand{\cellimg}[1]{%
  \img{#1}%
}

\newcommand{\cellimgtop}[1]{%
  \raisebox{-0.8mm}[\height][\depth]{\img{#1}}%
}

\makebox[\textwidth][c]{%
\begin{tabular}{@{}C{\datasetcolw}|C{\taskcolw}|C{\imgcolw}|C{\imgcolw}|C{\imgcolw}|C{\imgcolw}|C{\imgcolw}@{}}
\multicolumn{2}{c|}{}
& {\fontsize{9}{11}\selectfont\bfseries Human}
& {\fontsize{9}{11}\selectfont\bfseries EgoMimic}
& {\fontsize{9}{11}\selectfont\bfseries VACE}
& {\fontsize{9}{11}\selectfont\bfseries Phantom}
& {\fontsize{9}{11}\selectfont\bfseries EgoEngine} \\
\hline

\multirow[c]{6}{*}{\datasetlabeltaco}
& \multirow[c]{2}{*}{\tasklabel{(Brush, Brush, Bowl)}}
& \cellimgtop{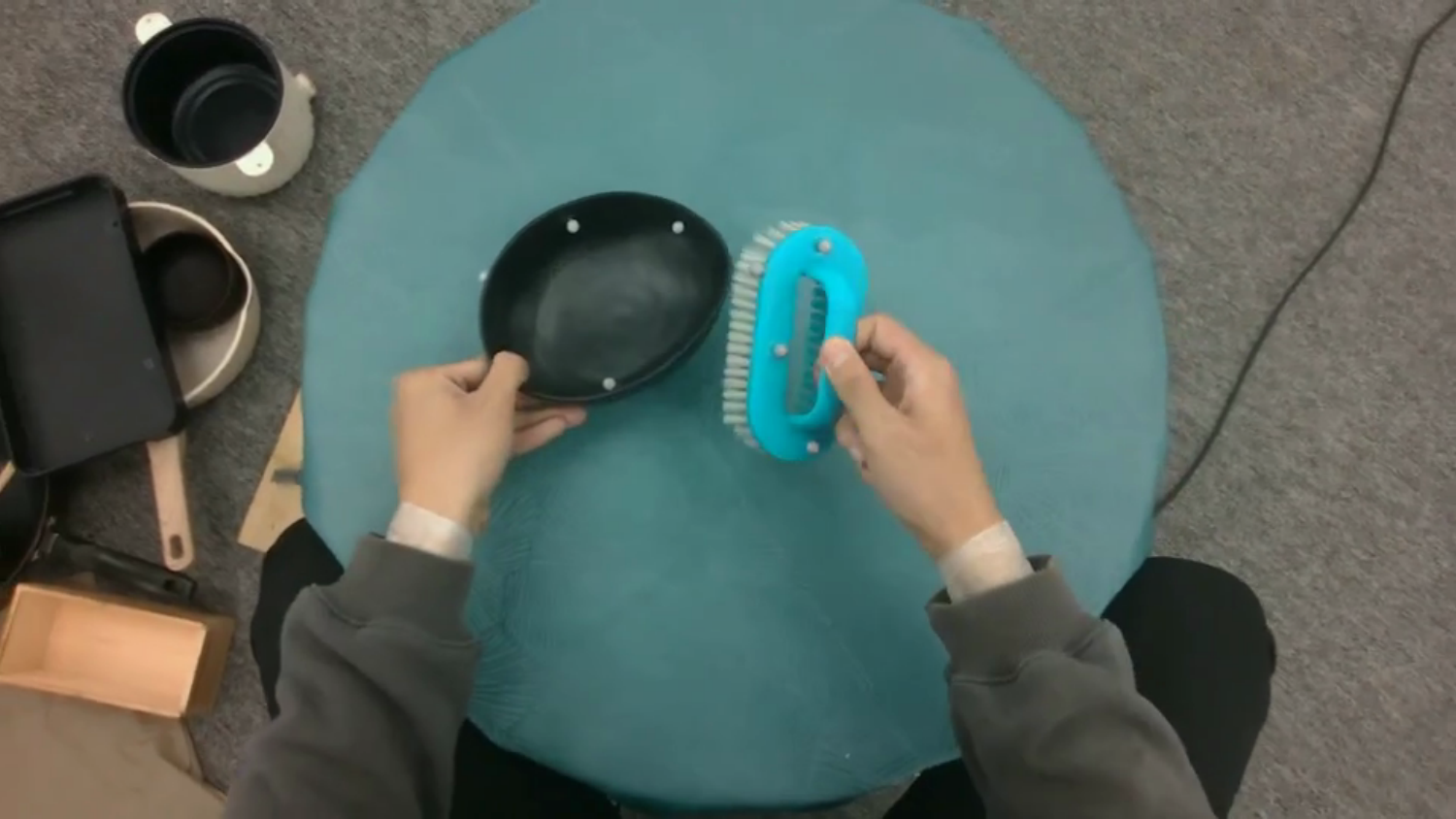}
& \cellimgtop{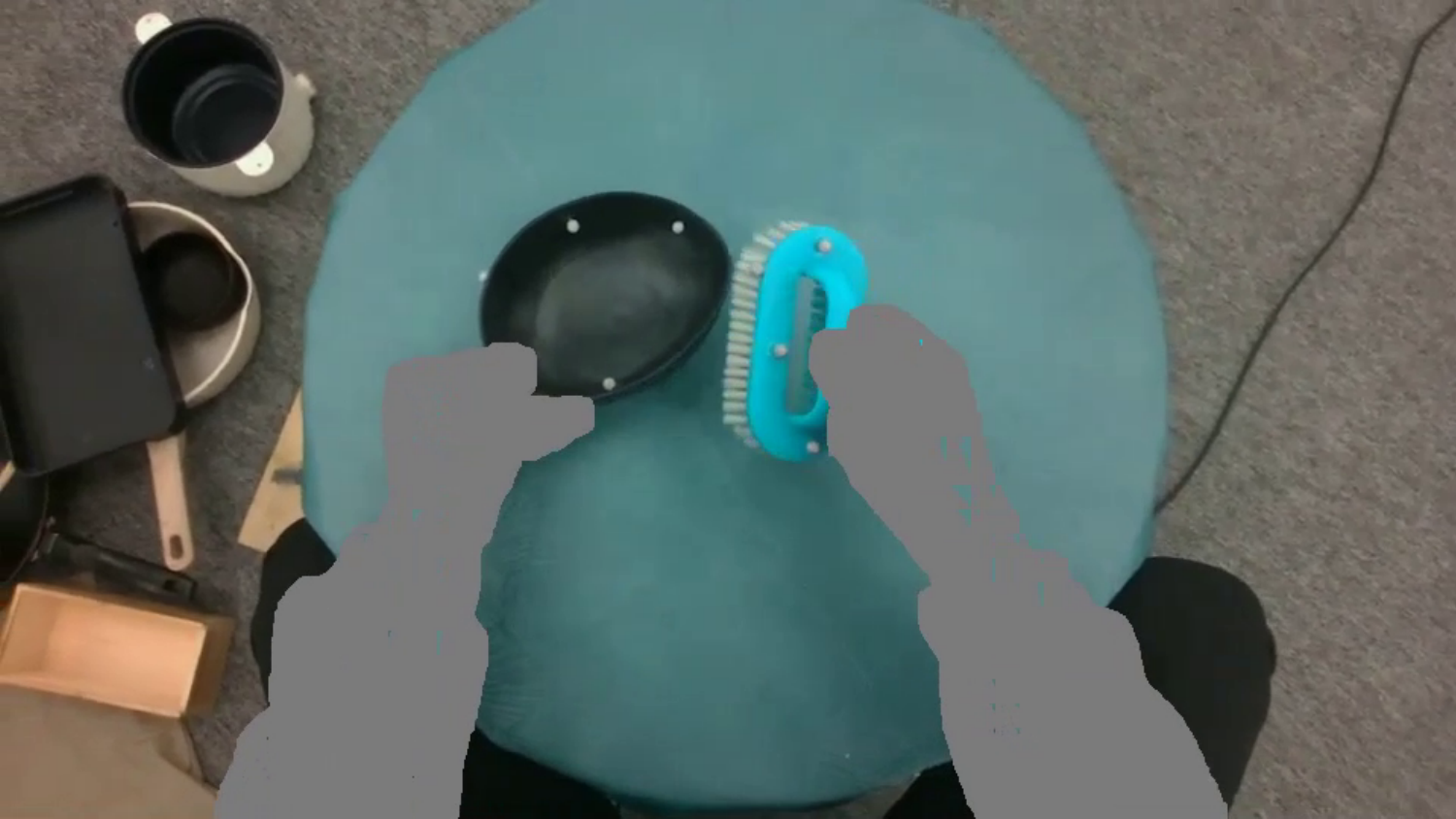}
& \cellimgtop{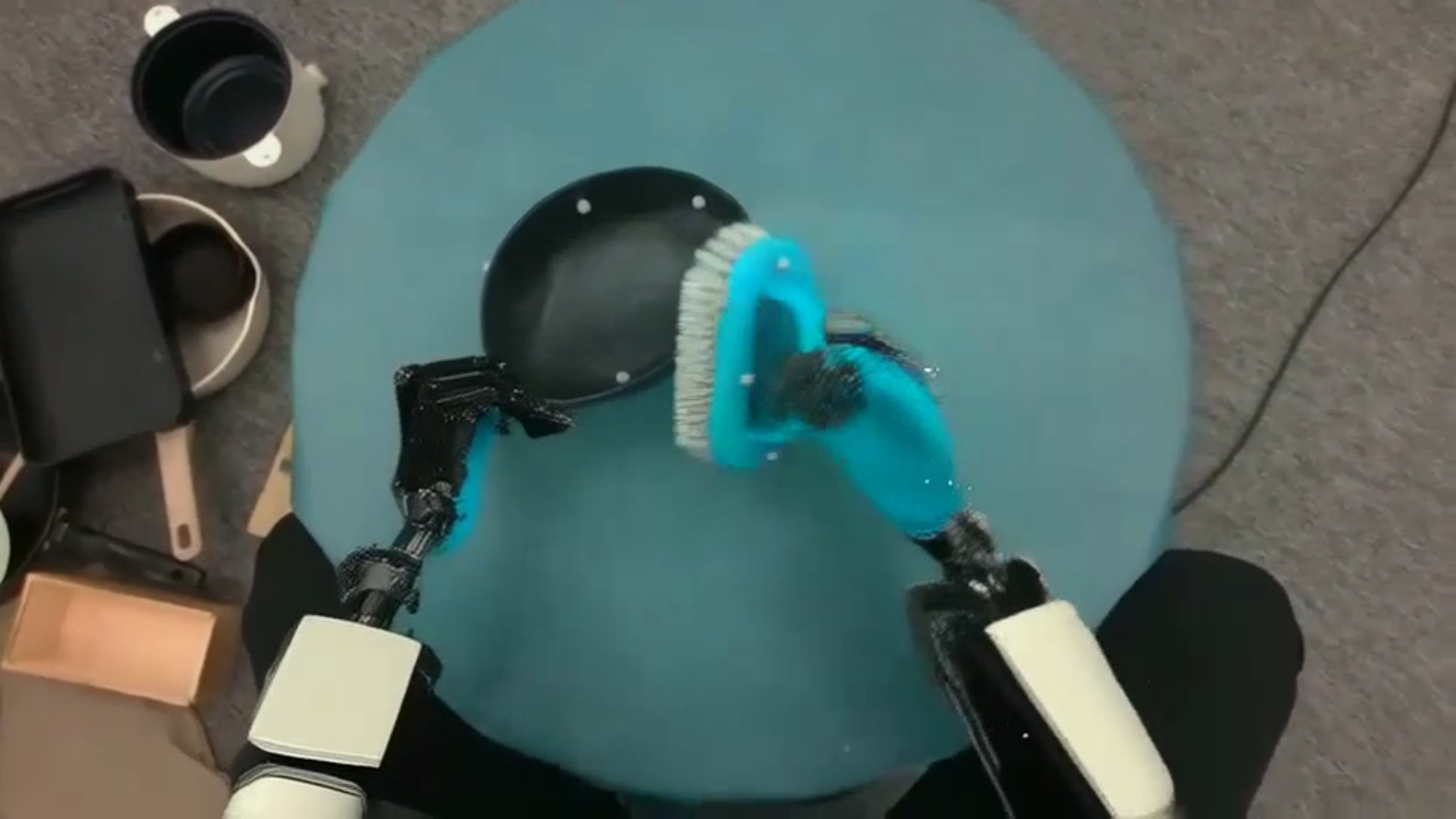}
& \cellimgtop{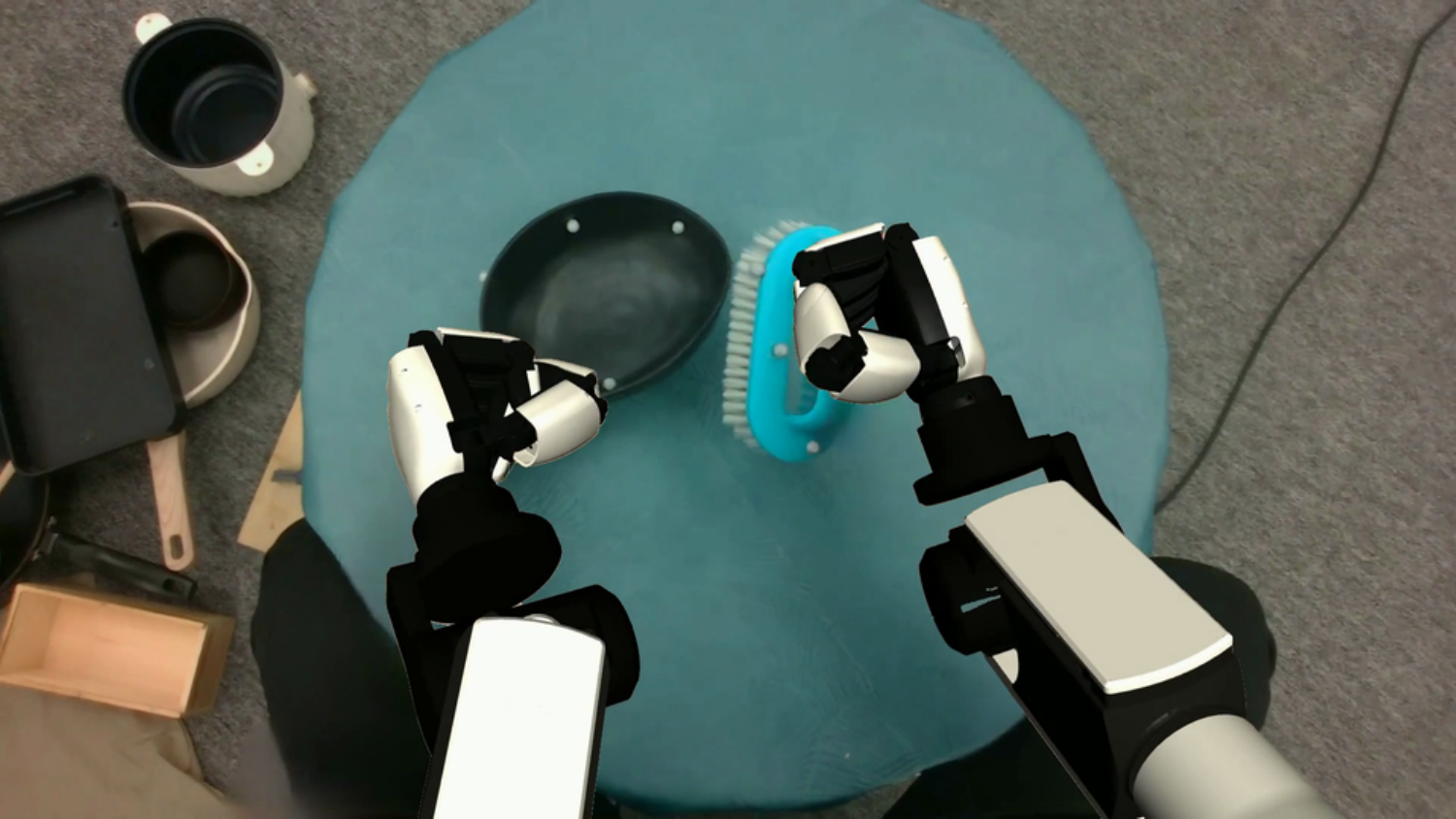}
& \cellimgtop{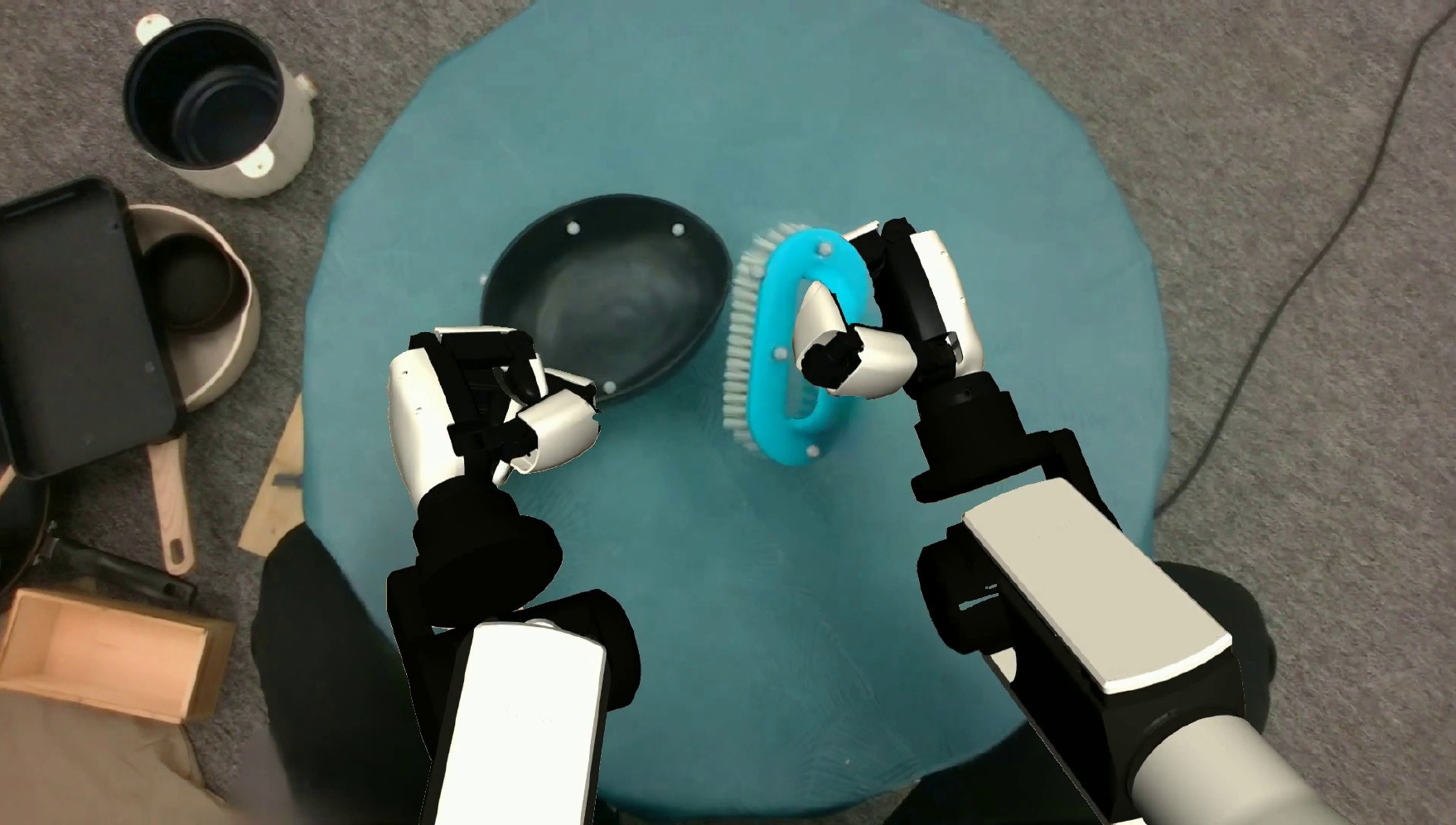} \\[0.4ex]
&
& \cellimg{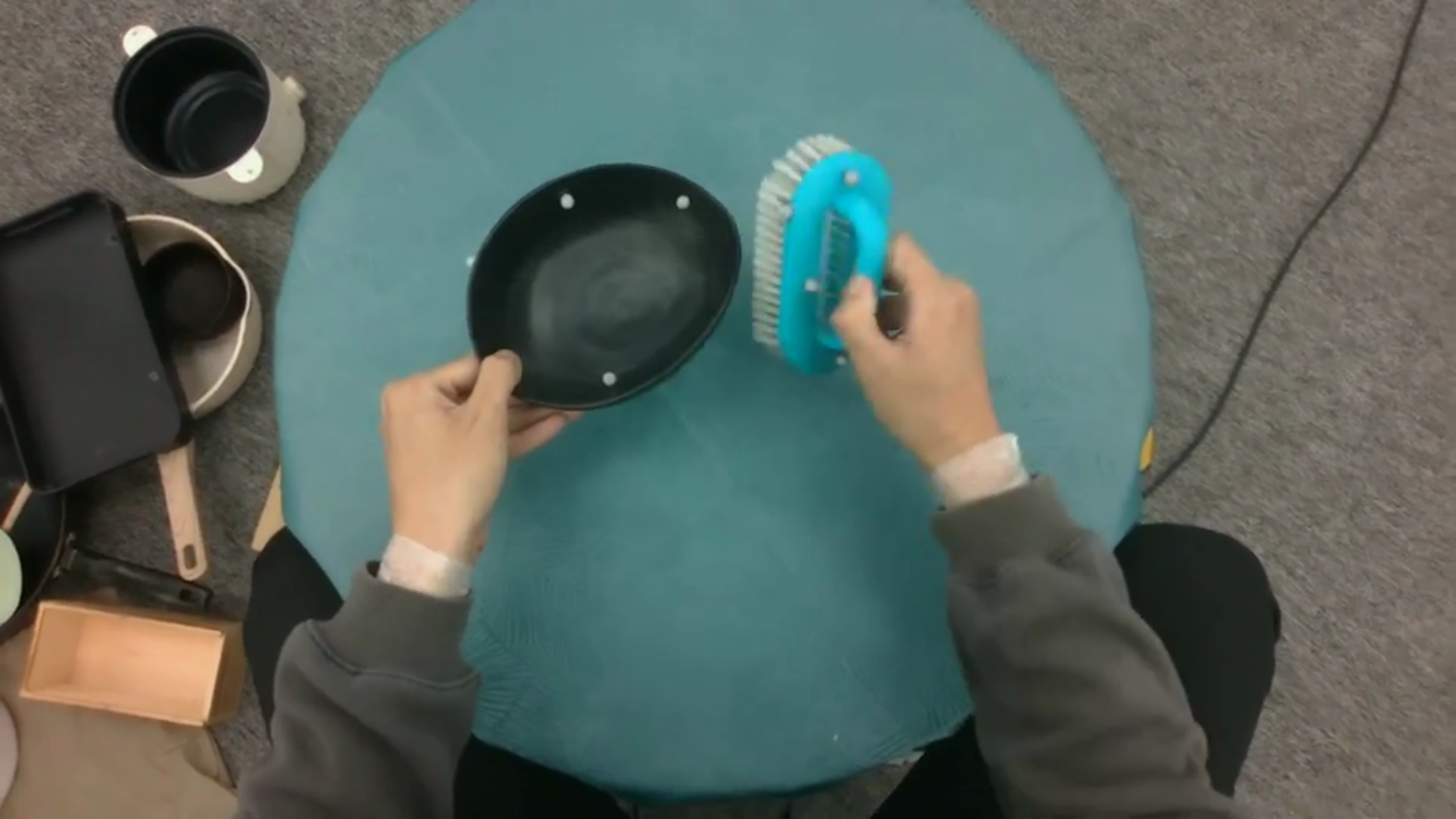}
& \cellimg{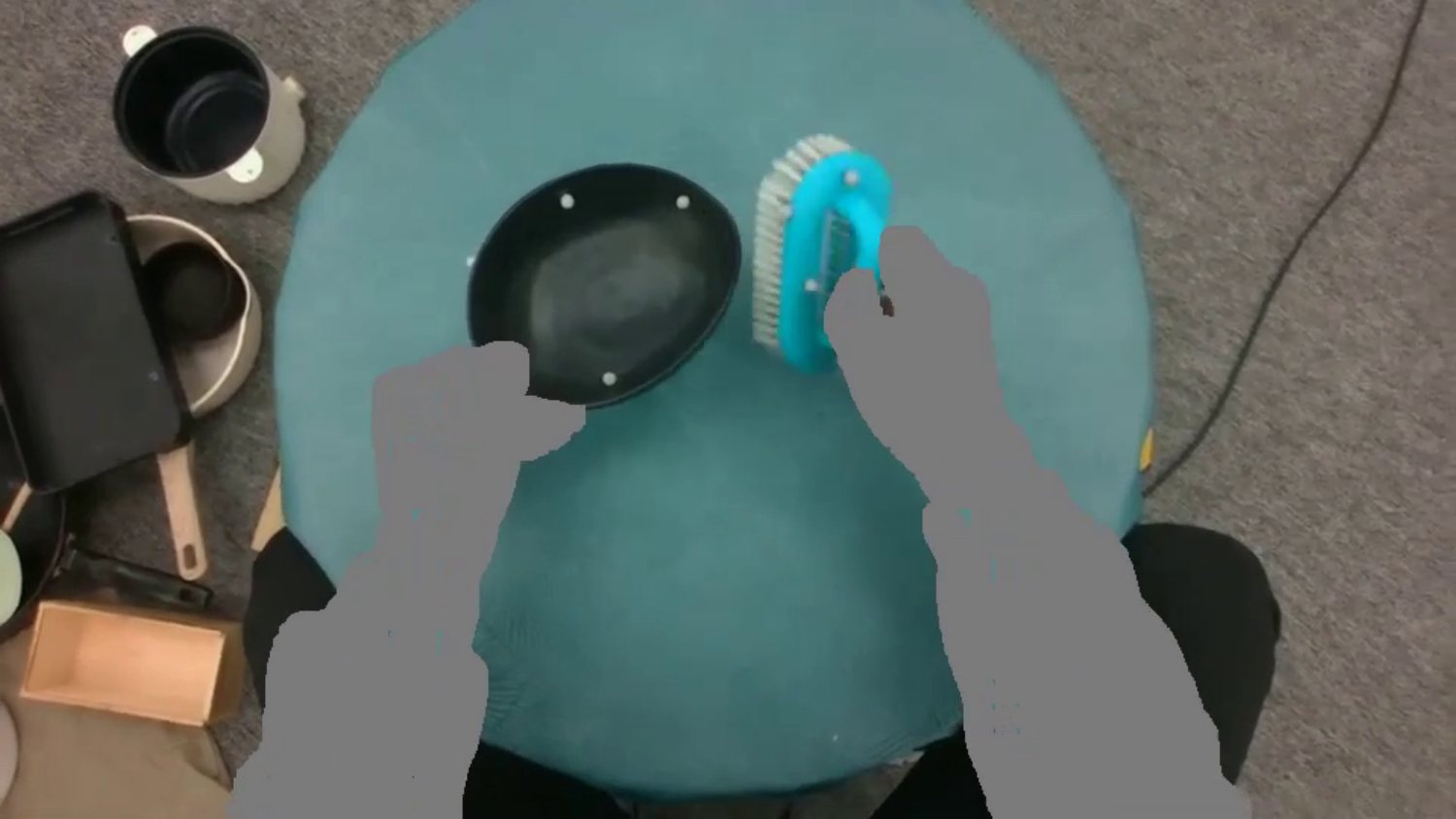}
& \cellimg{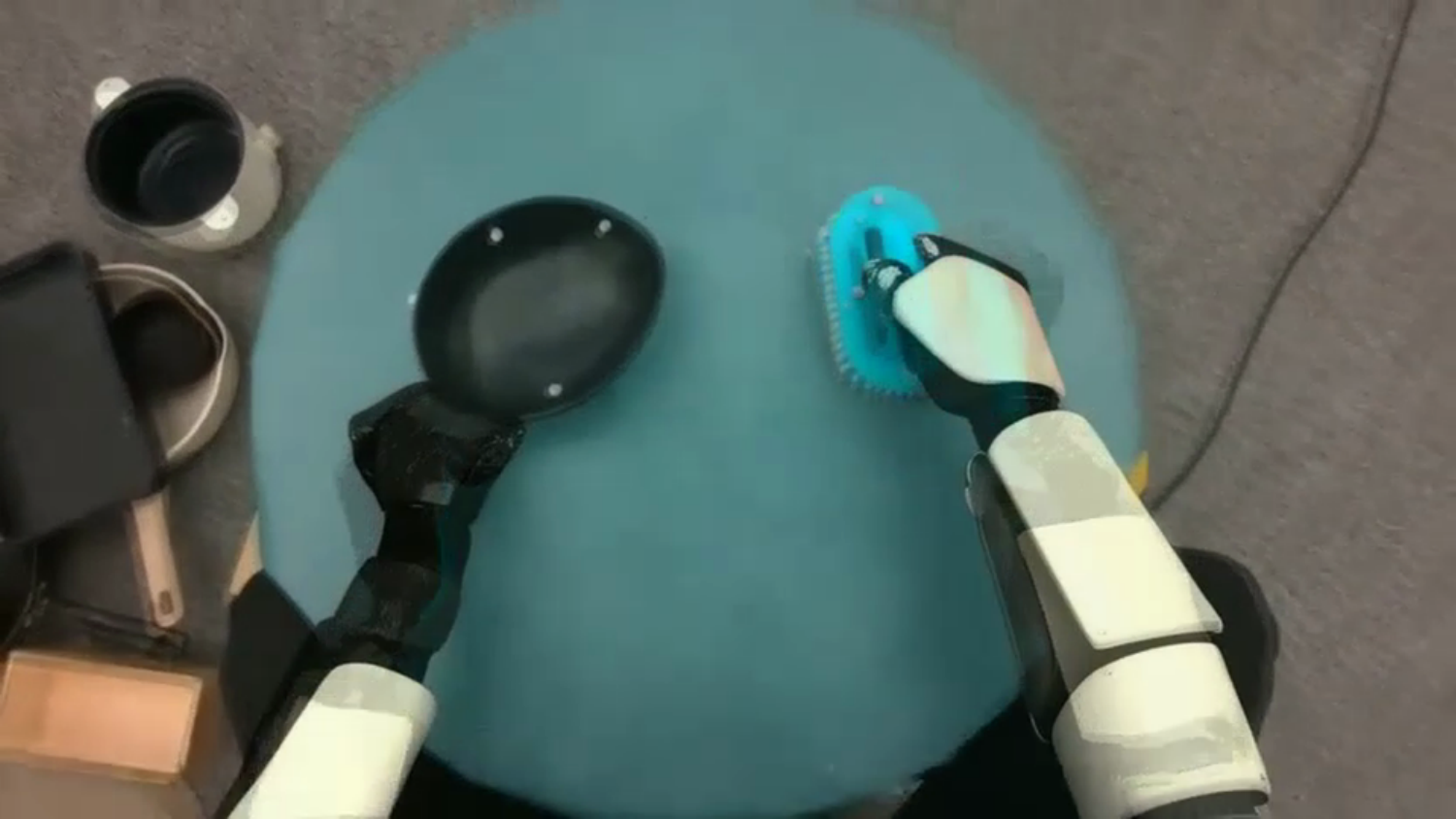}
& \cellimg{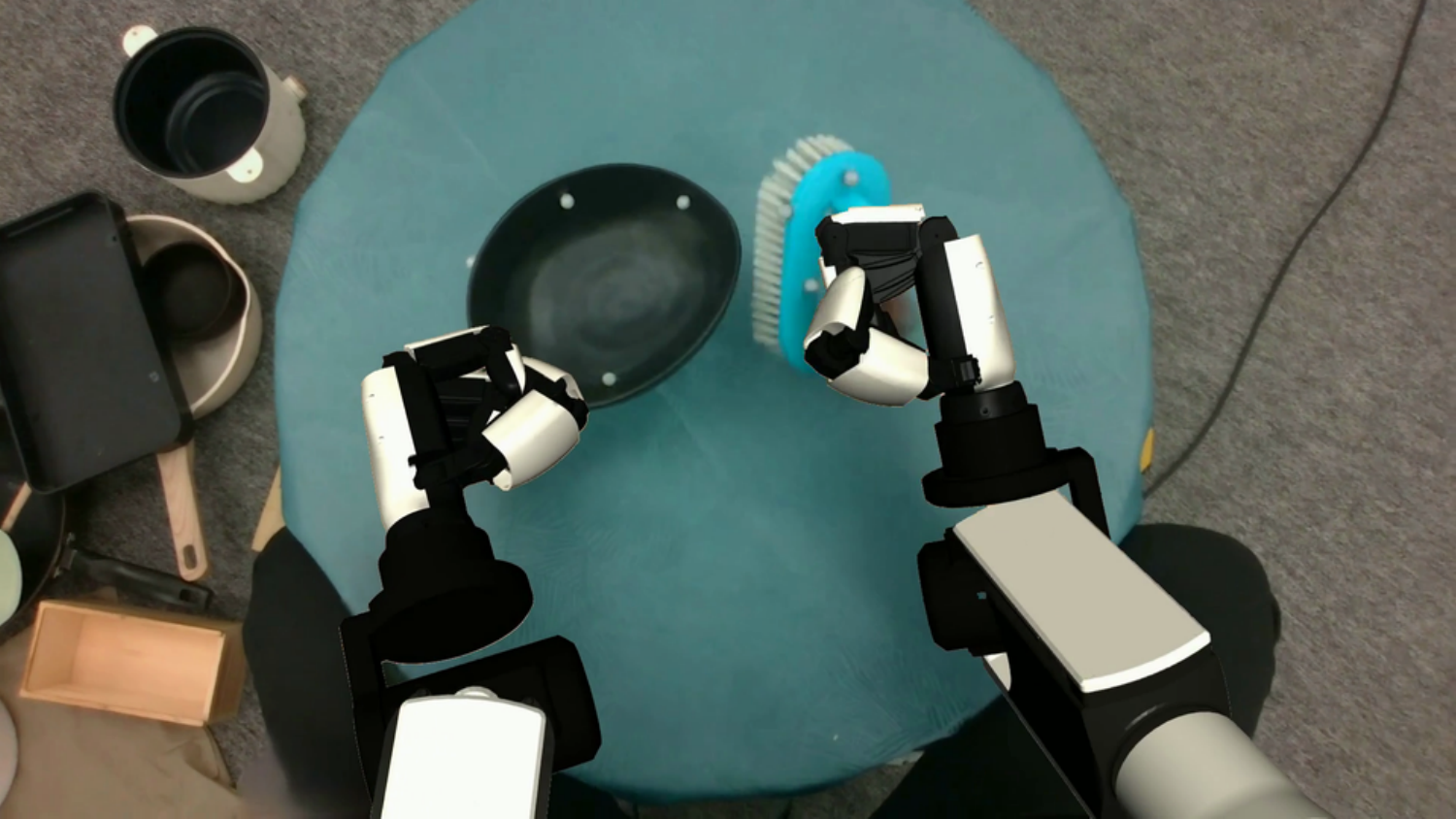}
& \cellimg{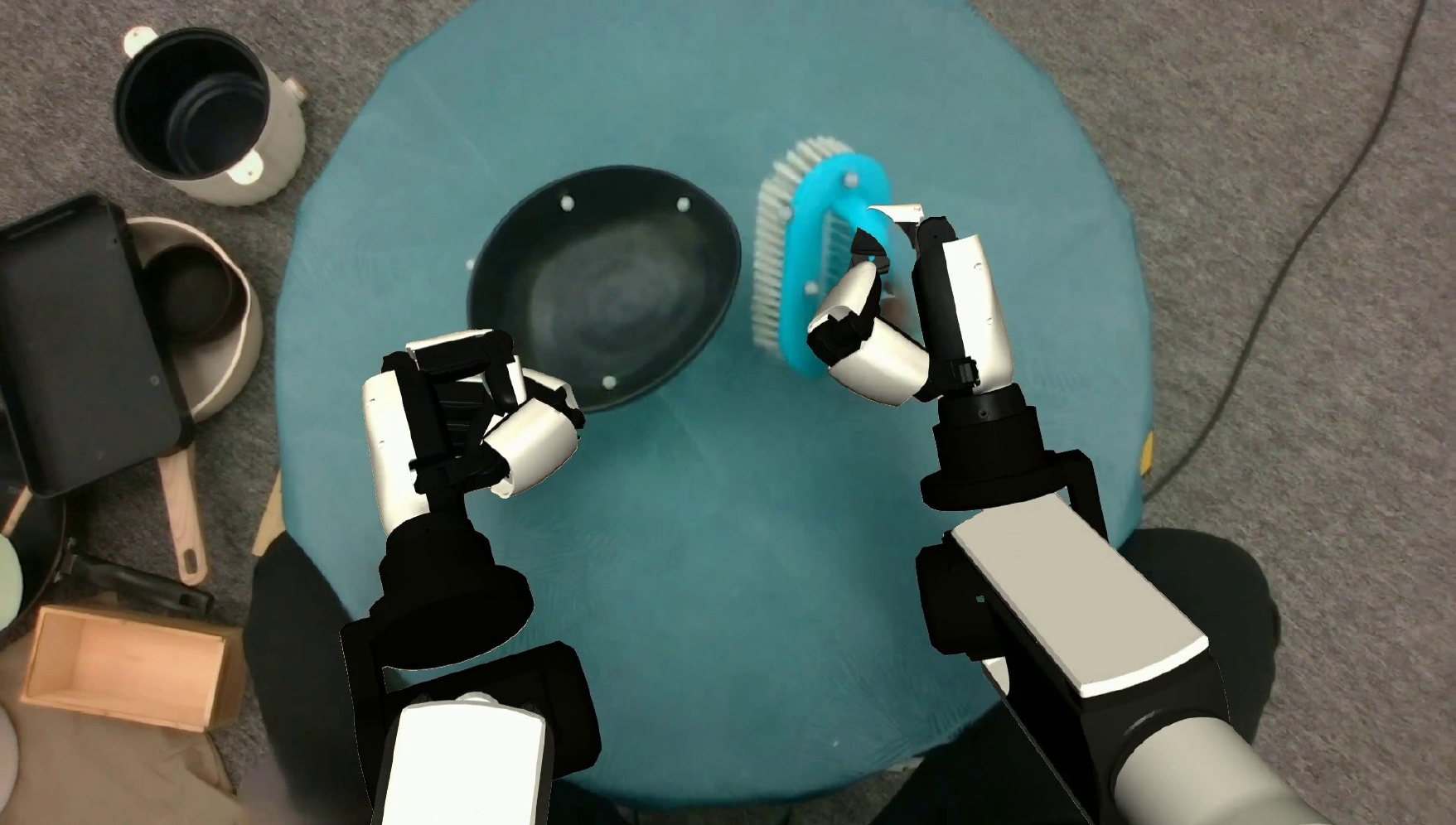} \\[-0.3ex]
\cline{2-7}

& \multirow[c]{2}{*}{\tasklabel{(Cut, Knife, Bowl)}}
& \cellimgtop{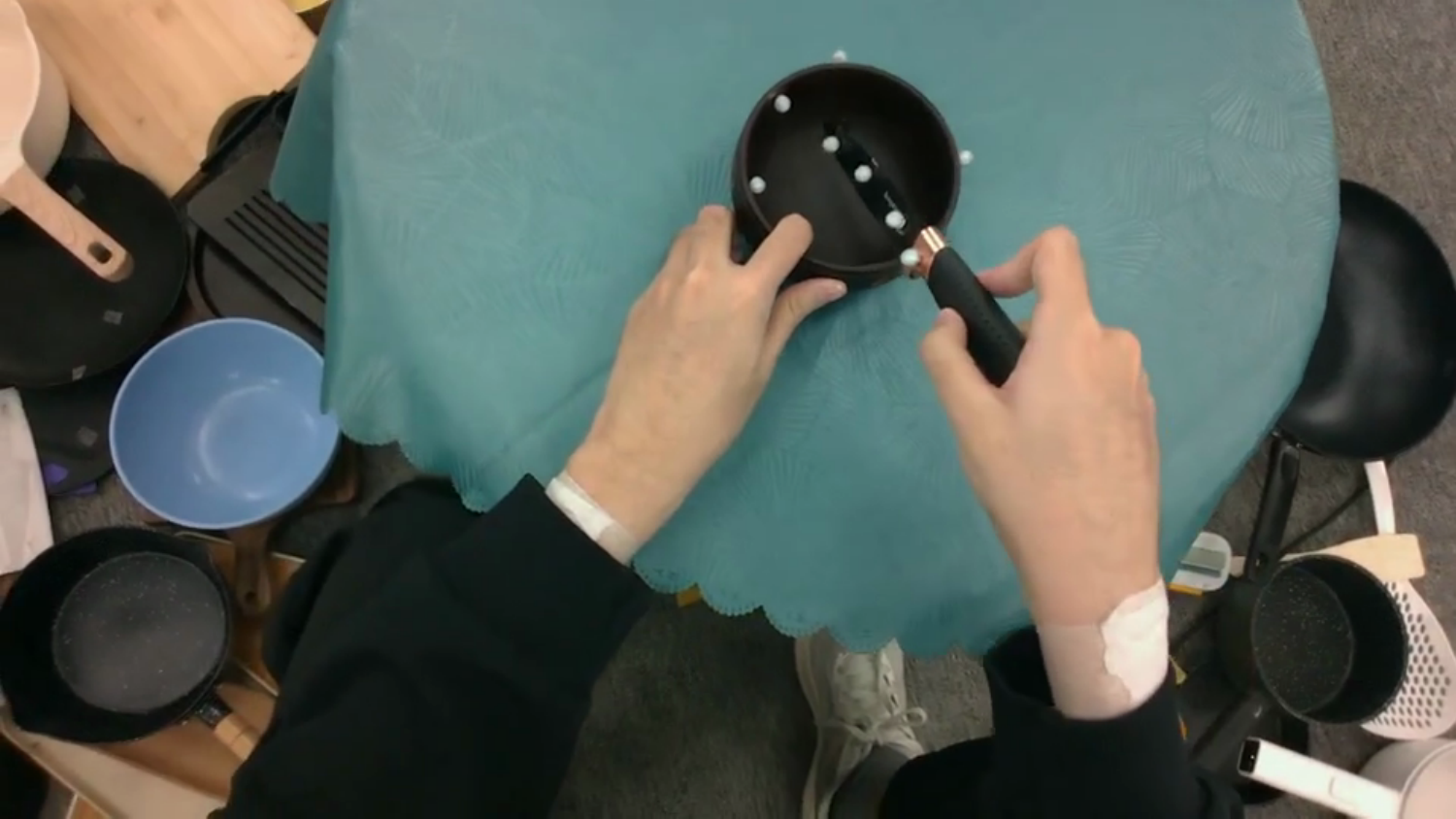}
& \cellimgtop{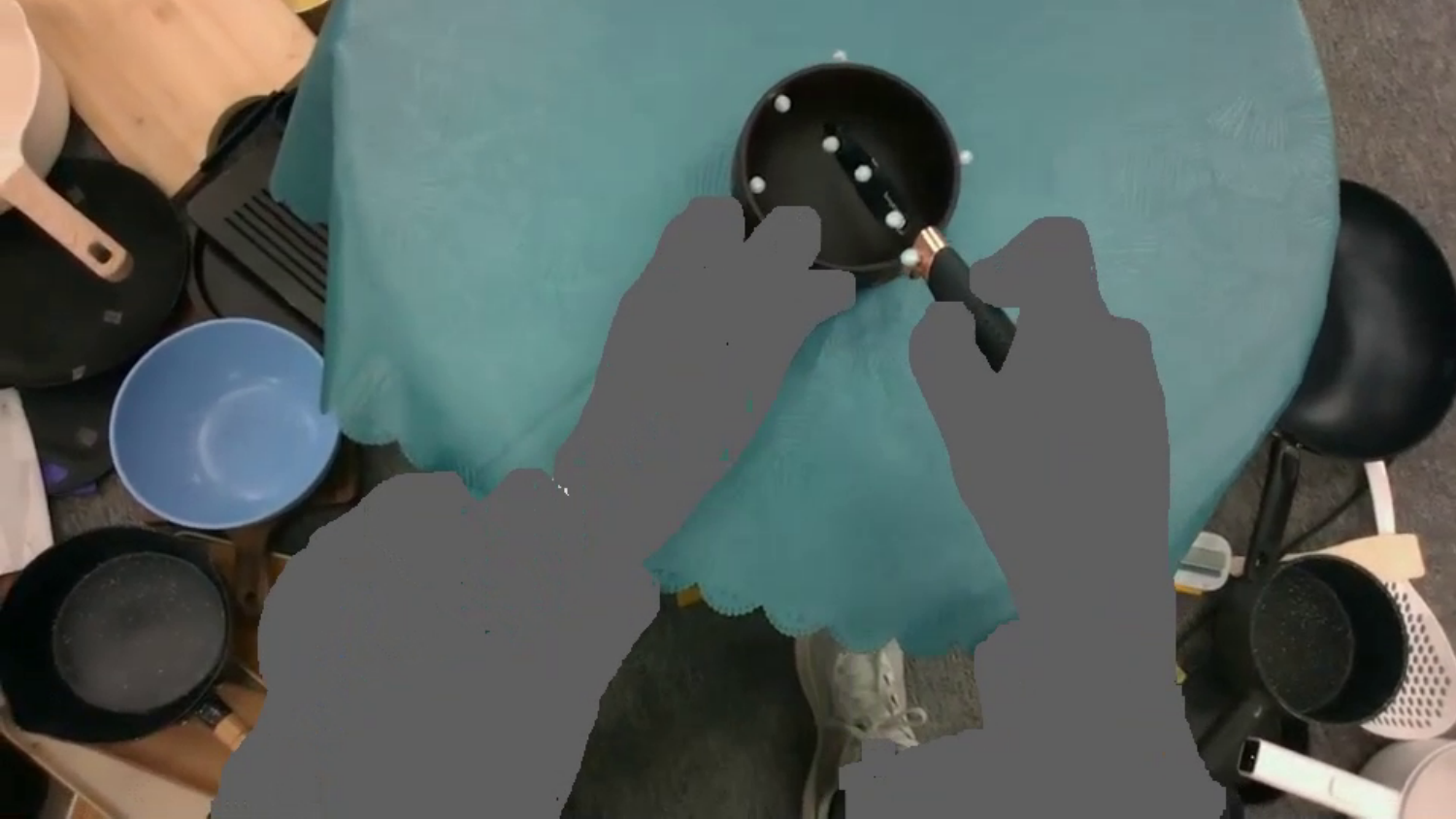}
& \cellimgtop{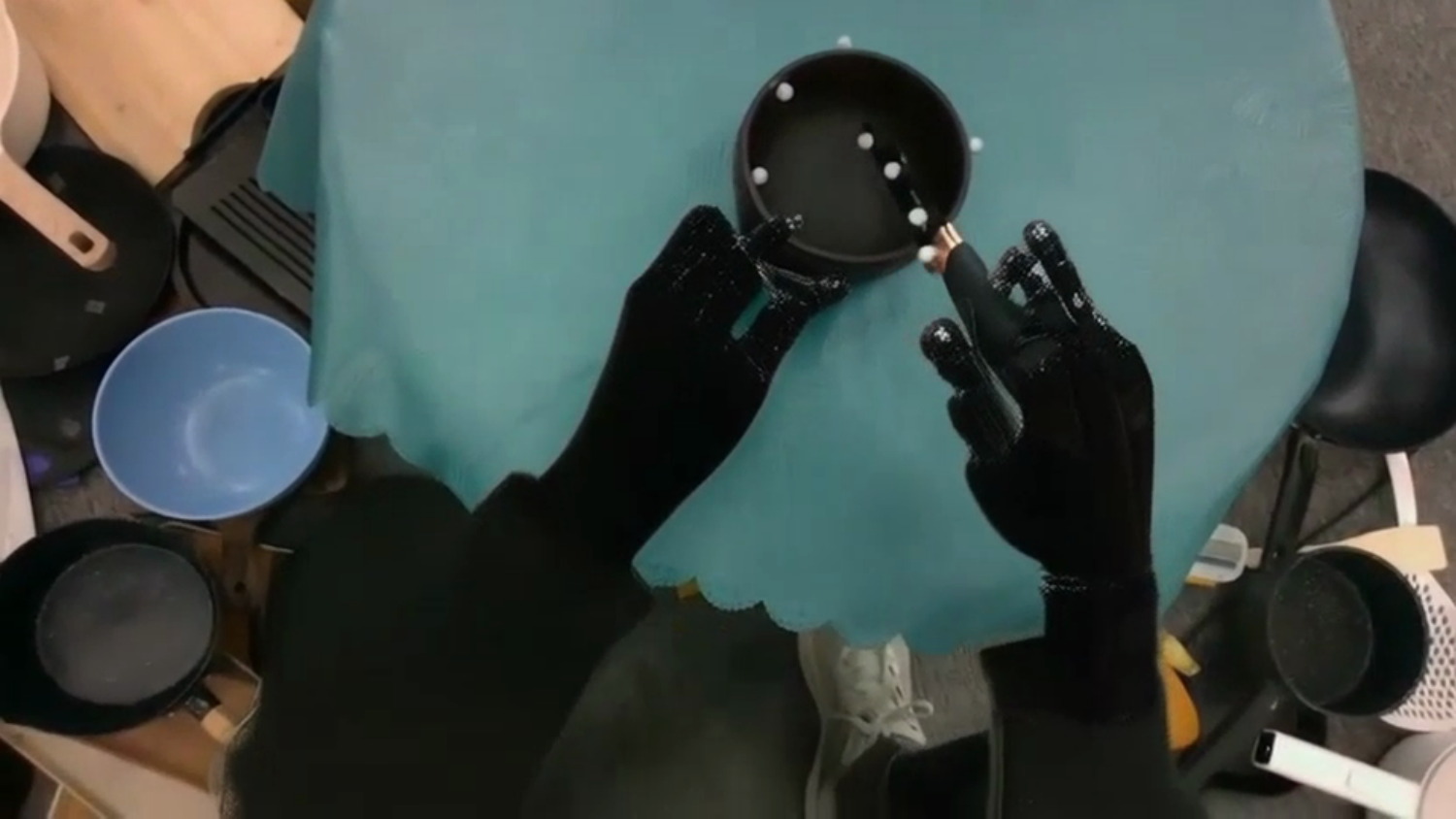}
& \cellimgtop{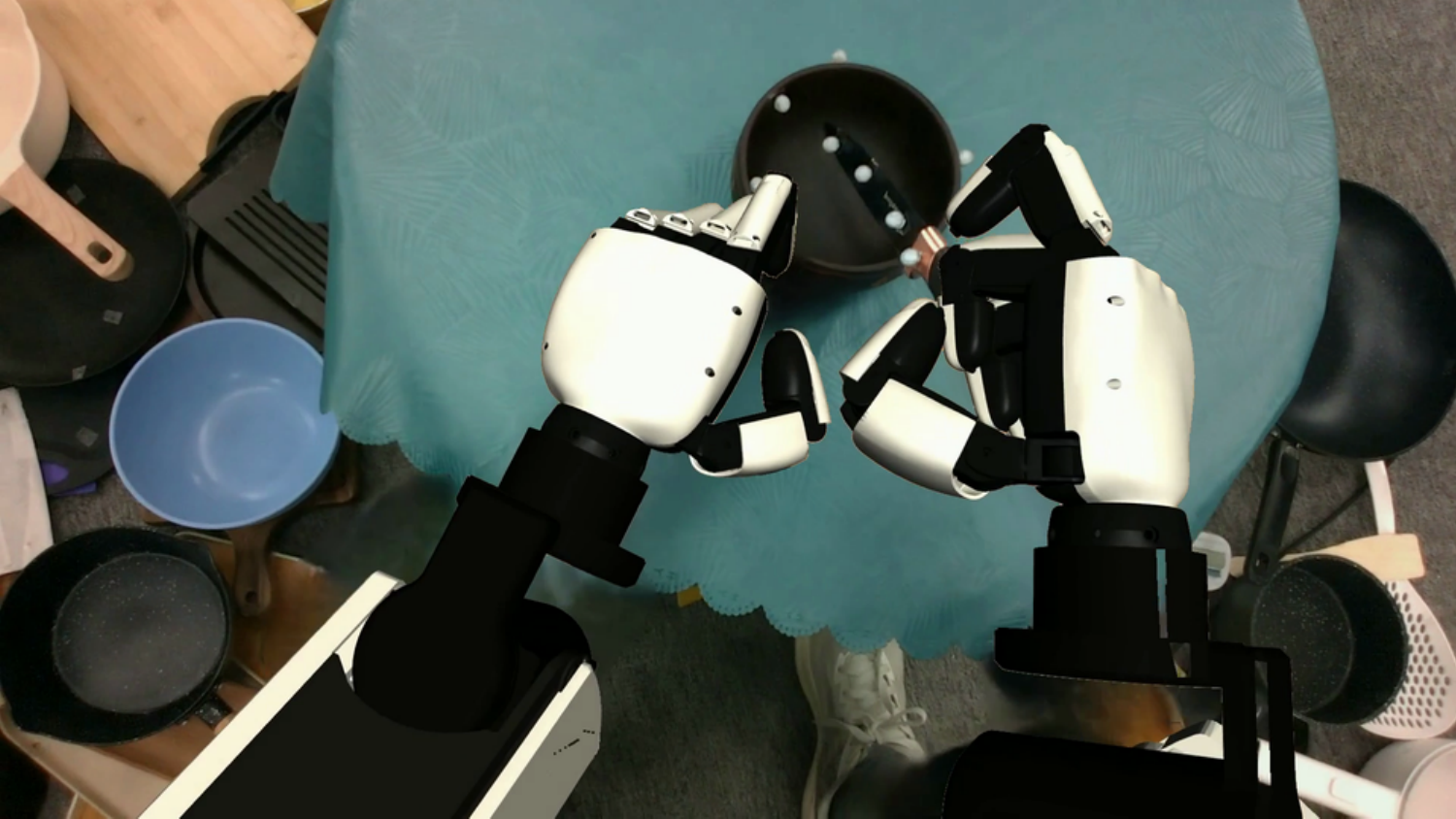}
& \cellimgtop{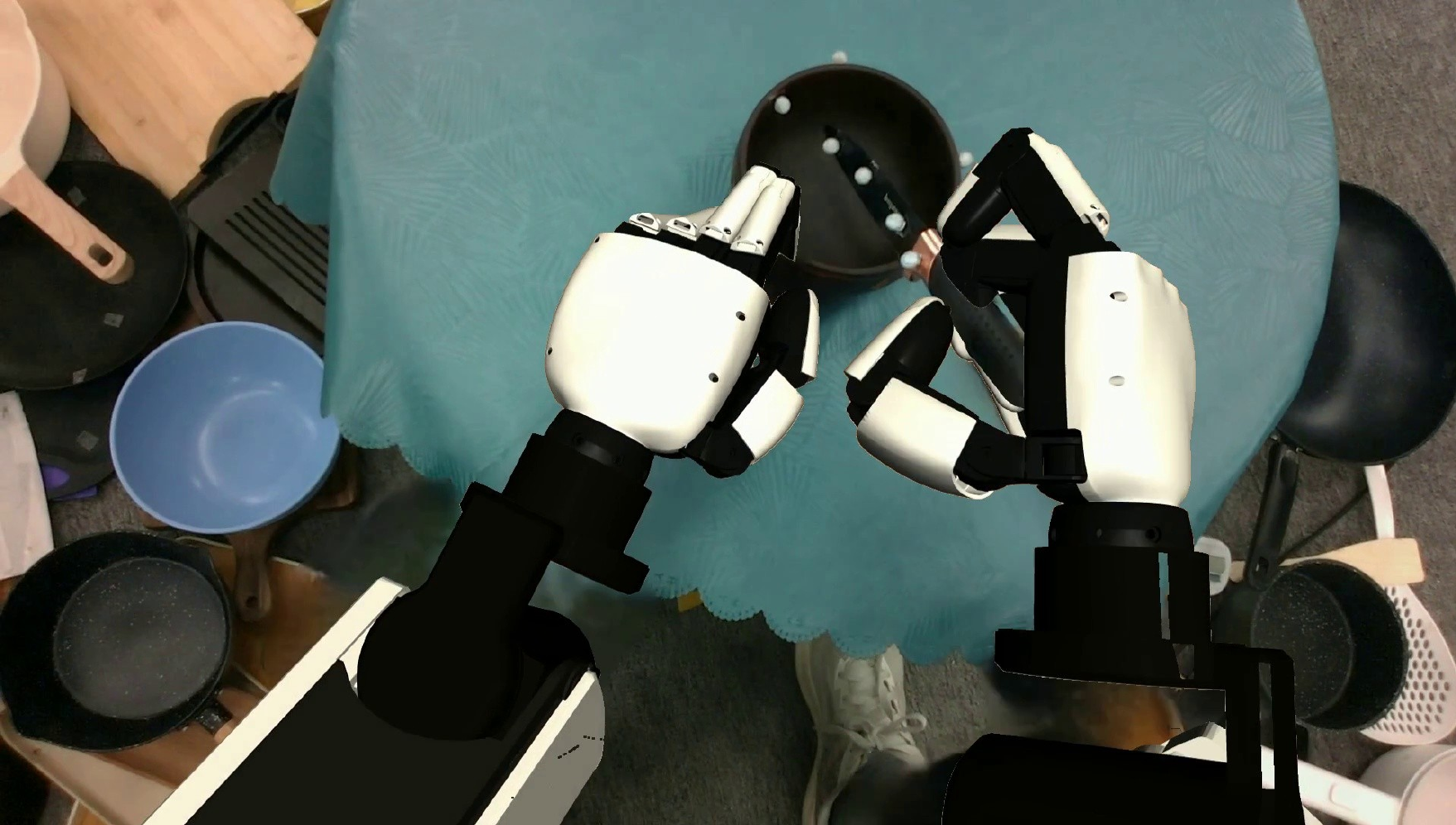} \\[0.4ex]
&
& \cellimg{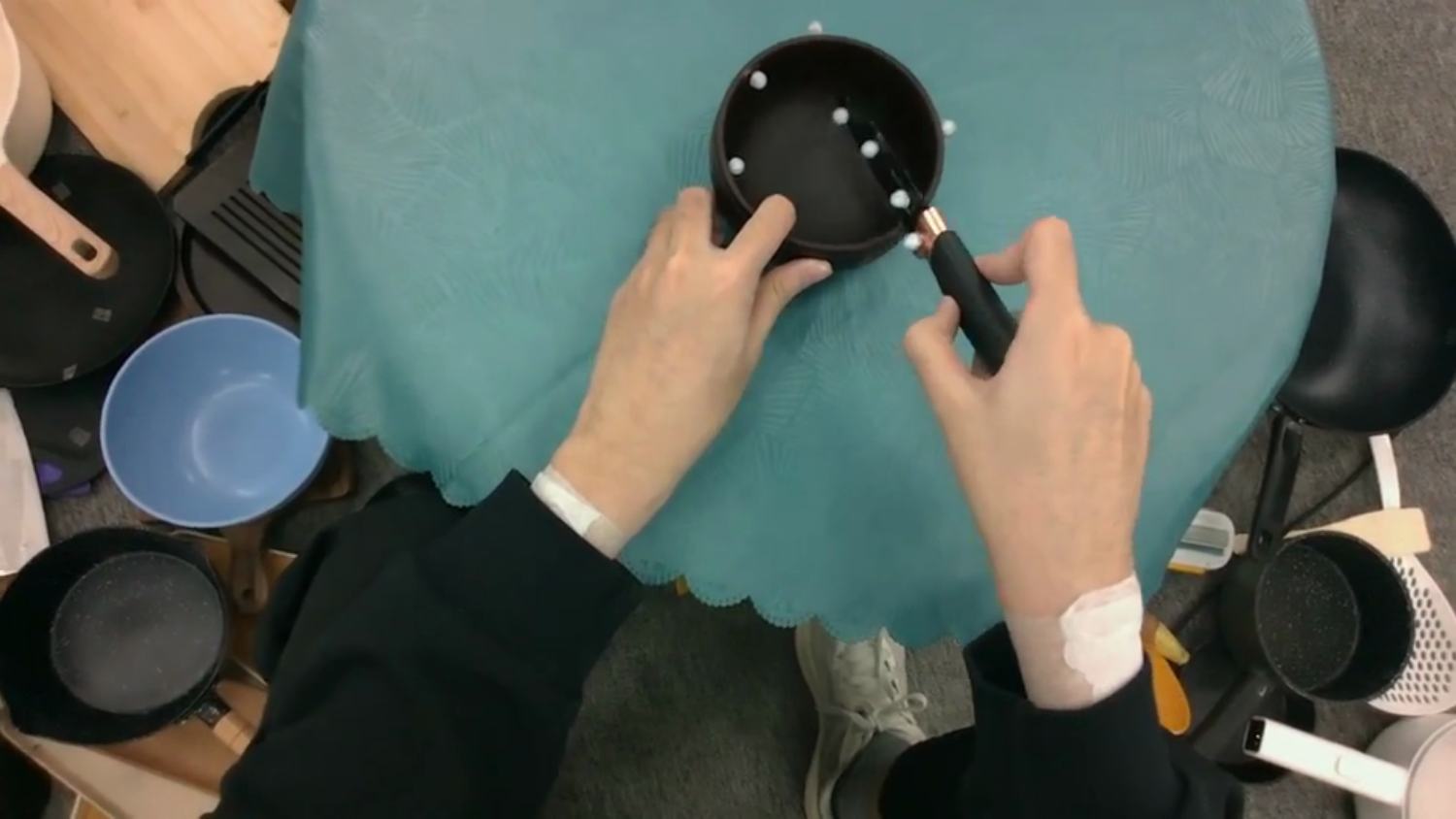}
& \cellimg{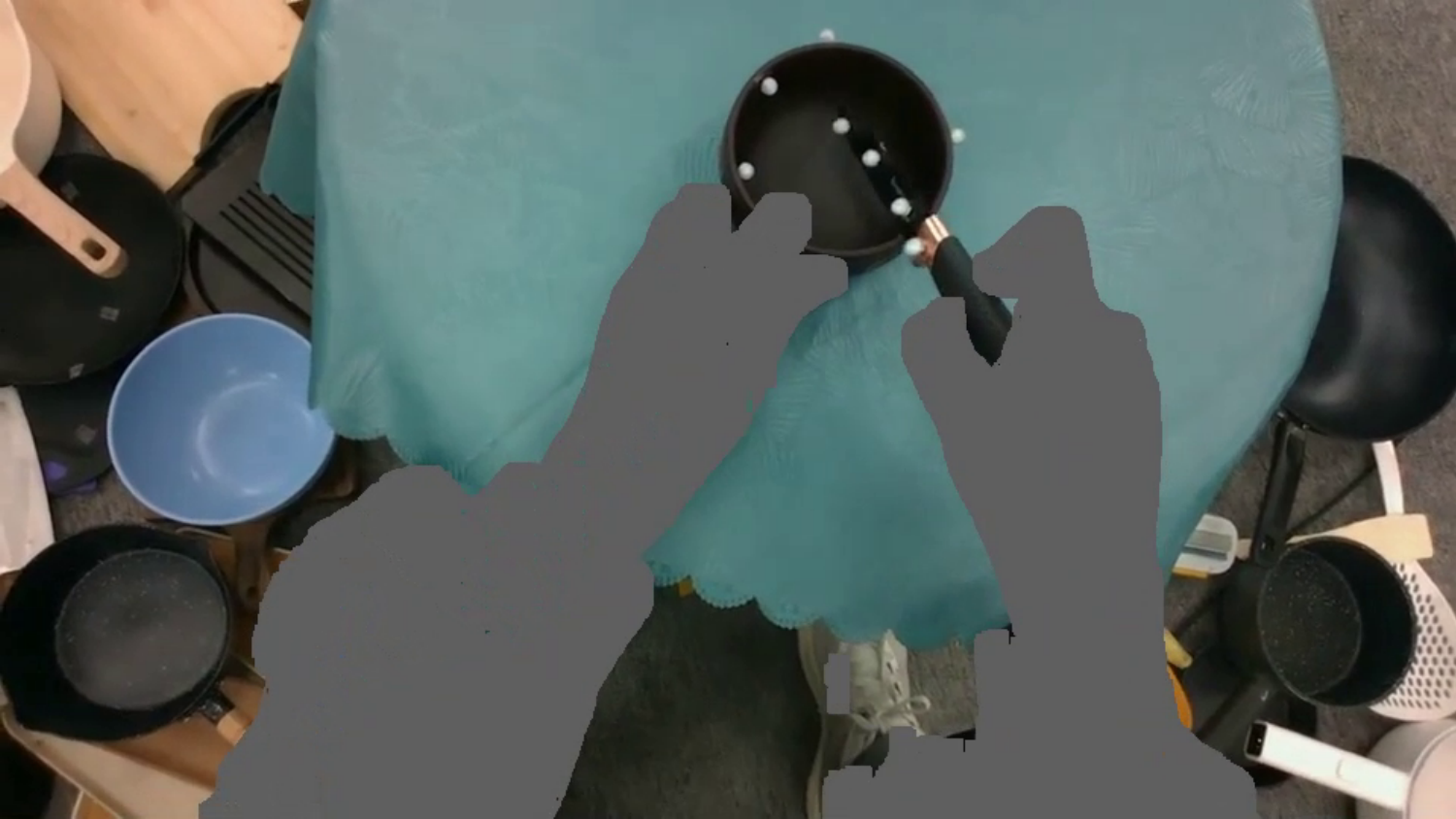}
& \cellimg{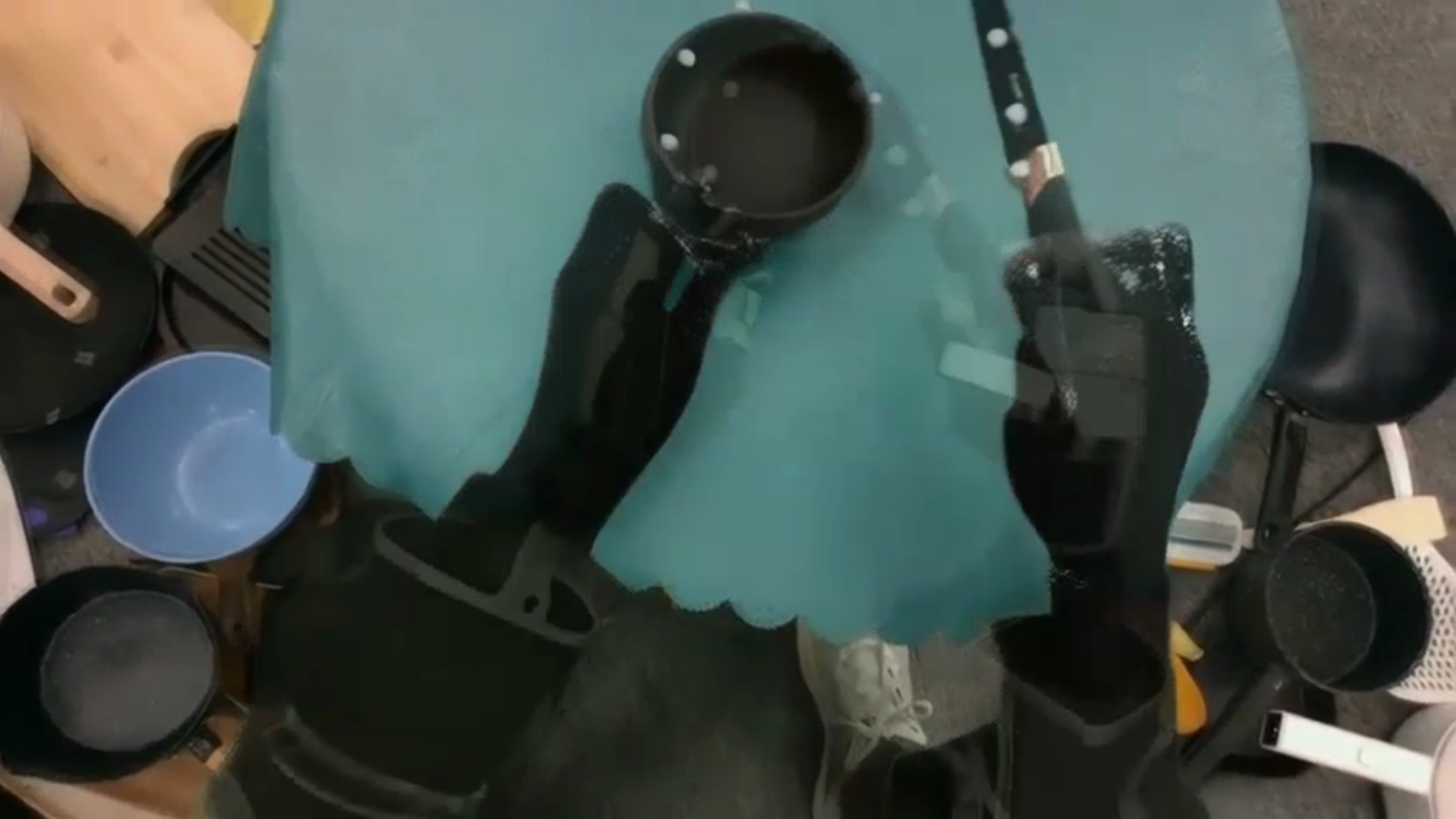}
& \cellimg{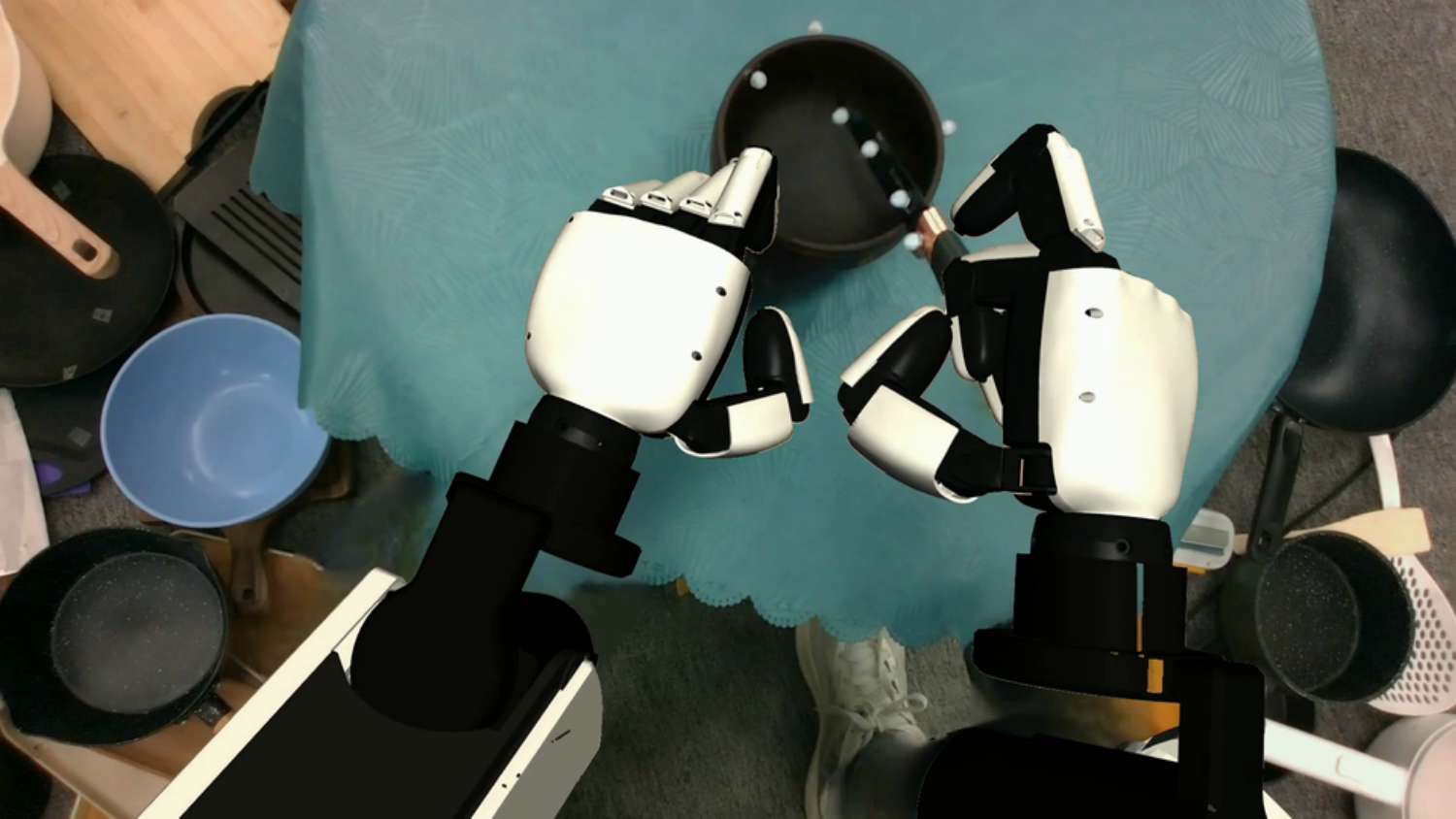}
& \cellimg{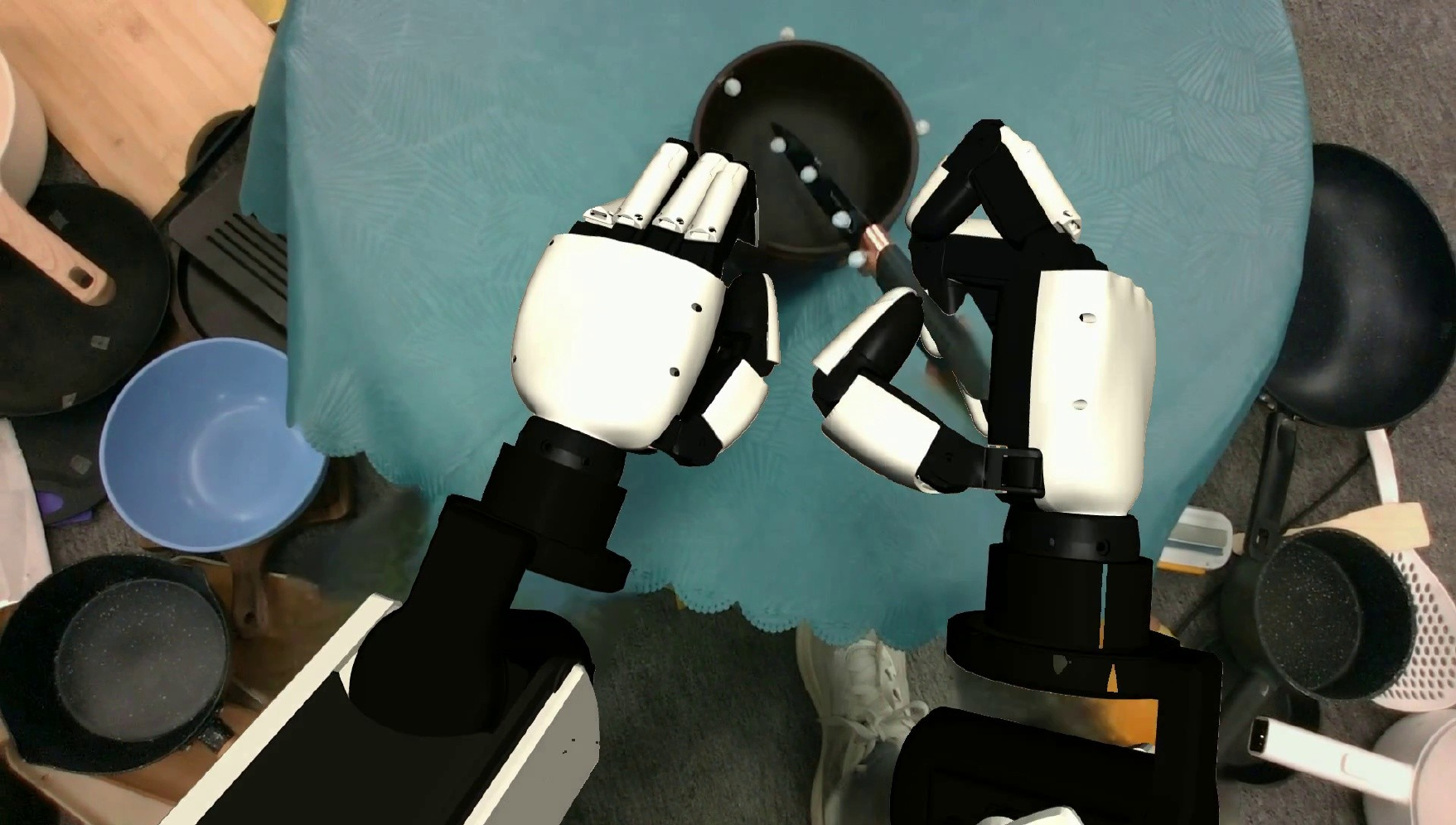} \\[-0.3ex]
\cline{2-7}

& \multirow[c]{2}{*}{\tasklabel{(Brush, Roller, Box)}}
& \cellimgtop{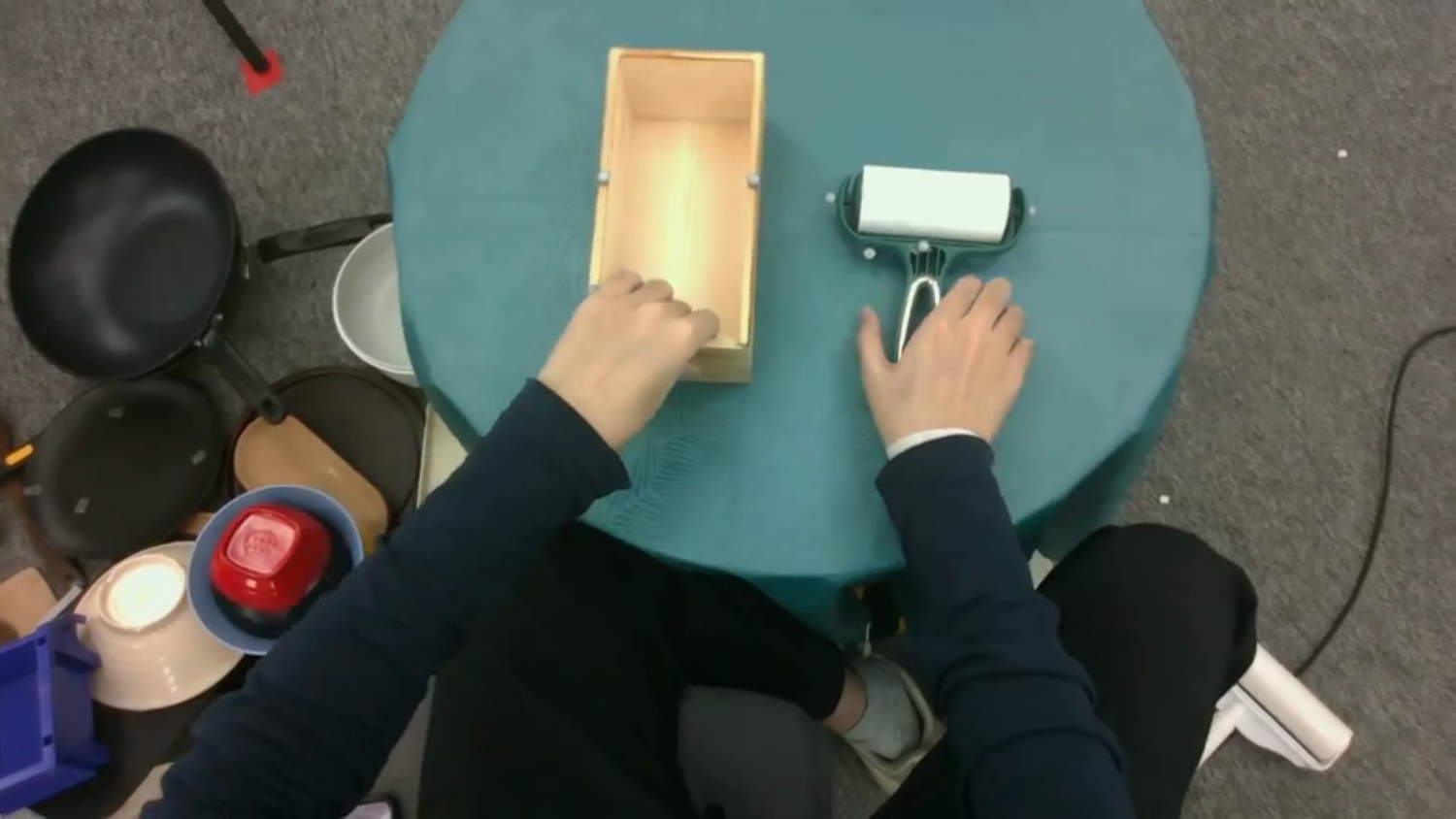}
& \cellimgtop{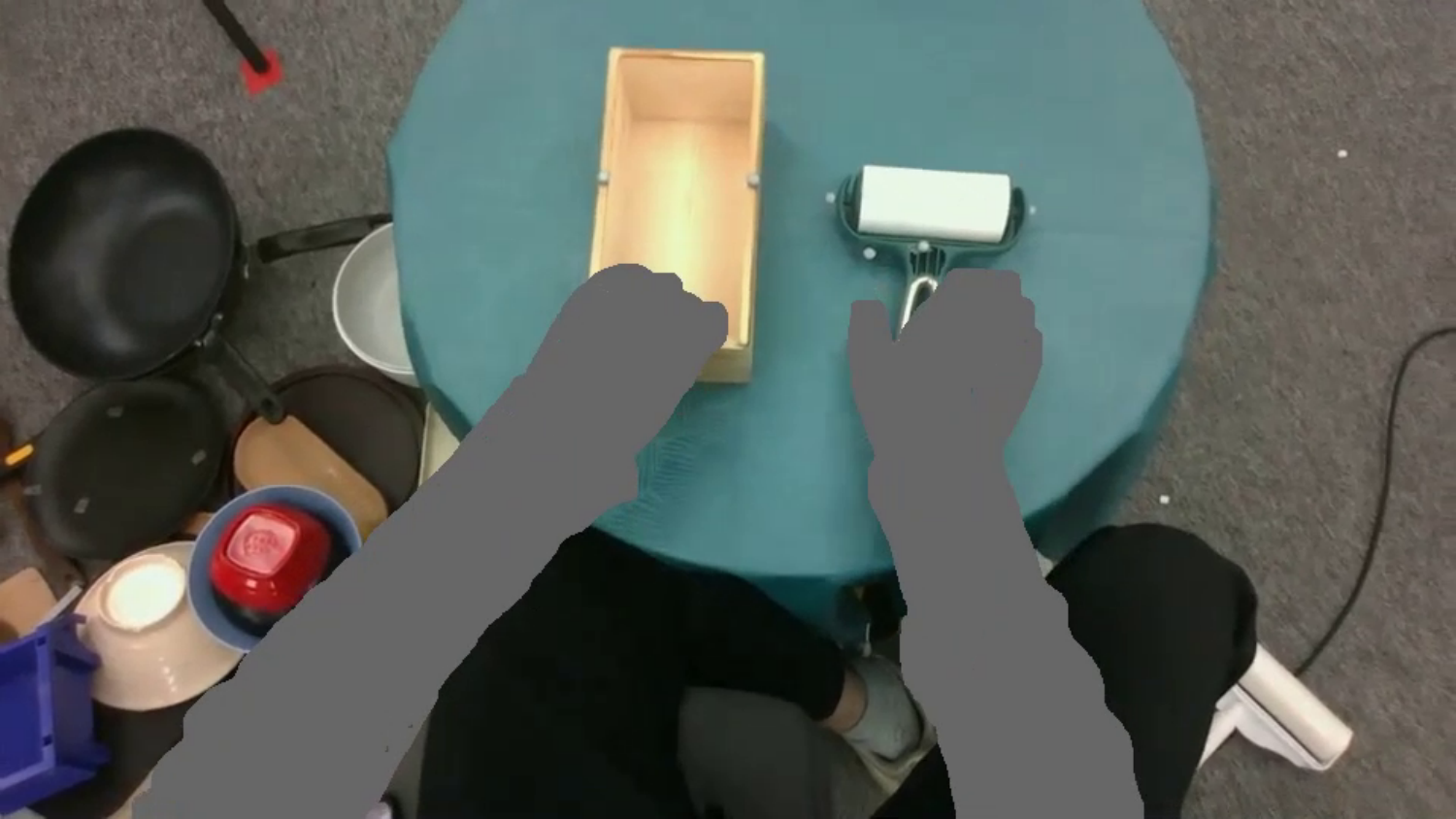}
& \cellimgtop{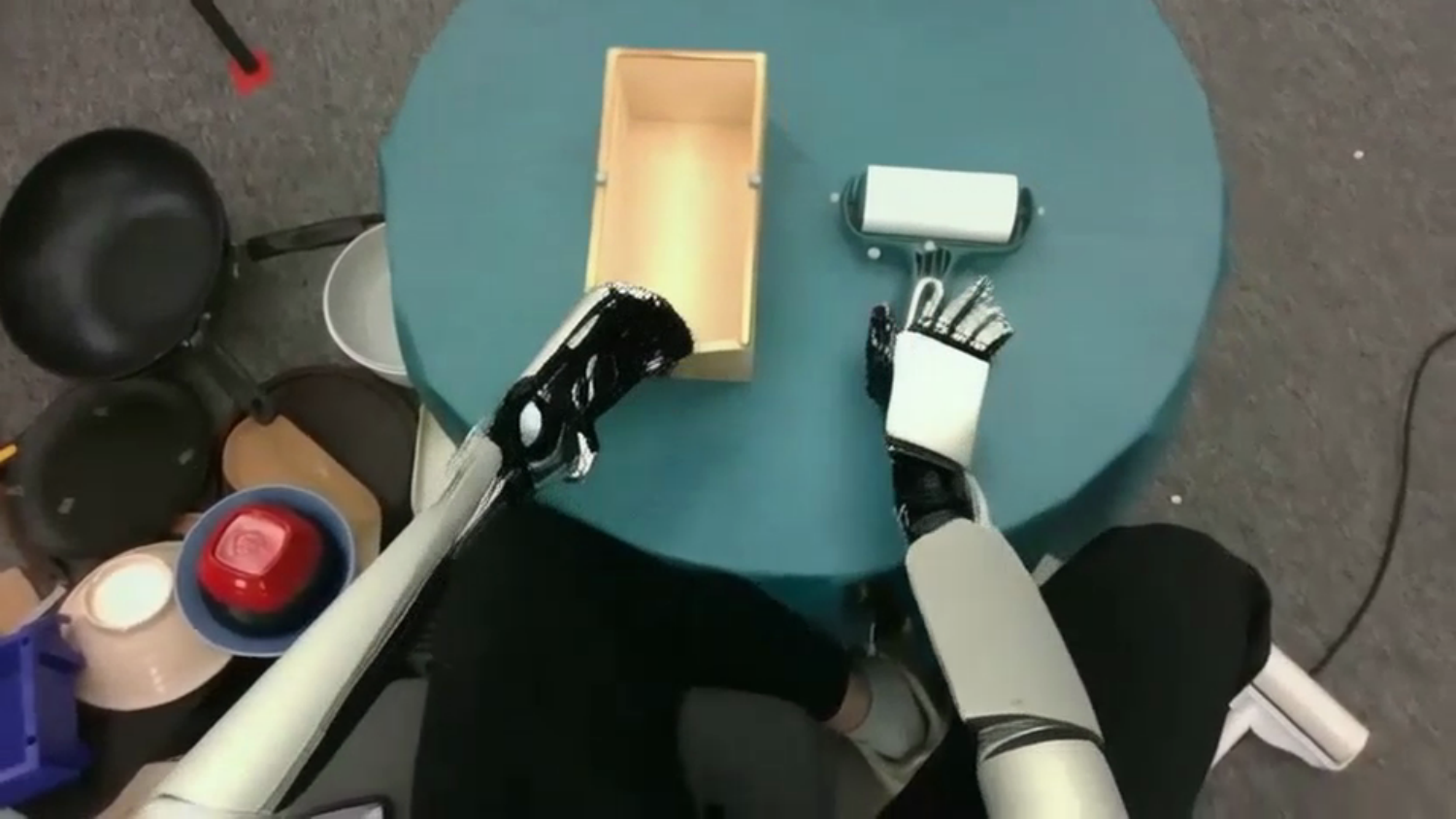}
& \cellimgtop{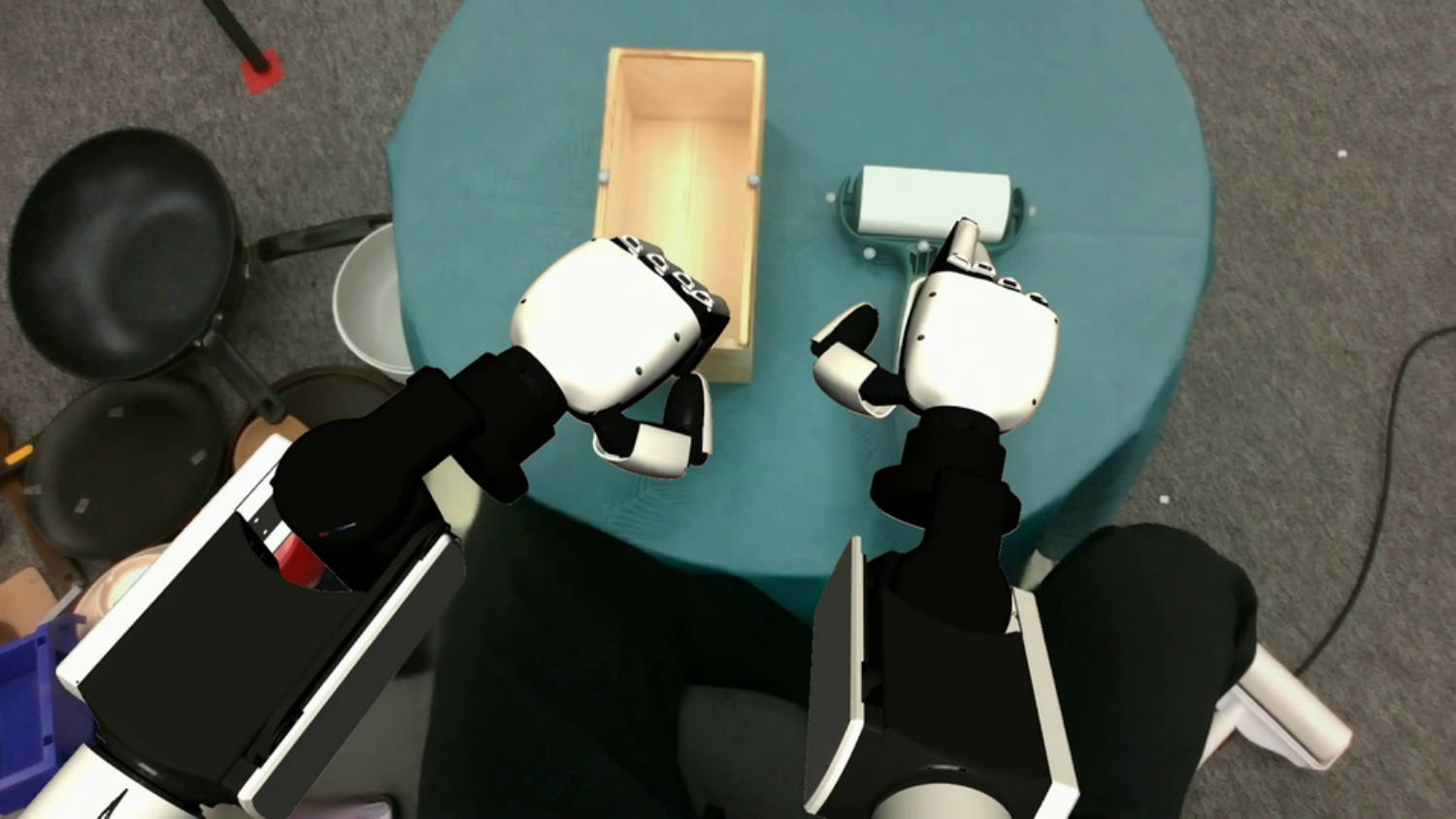}
& \cellimgtop{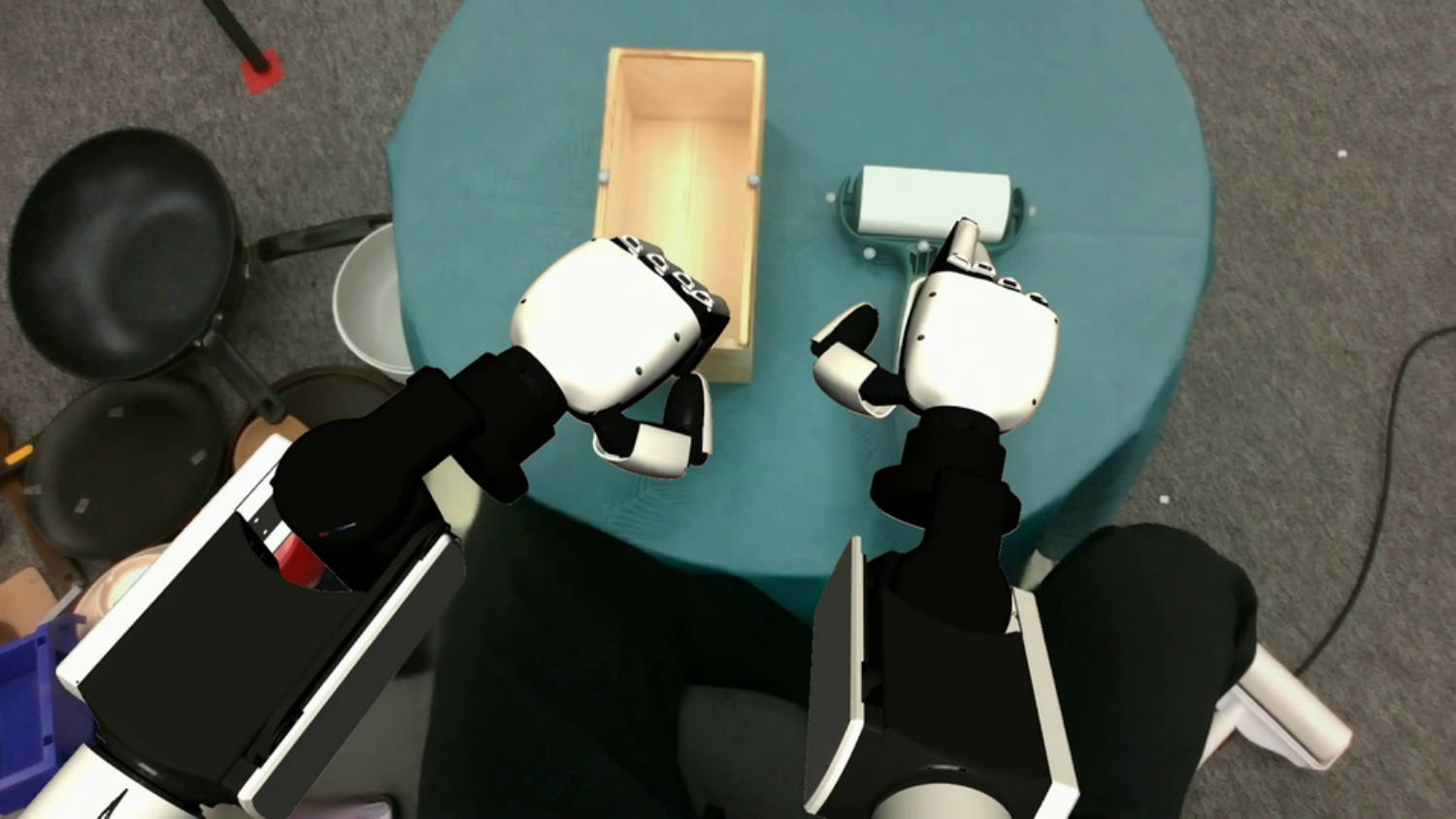} \\[0.4ex]
&
& \cellimg{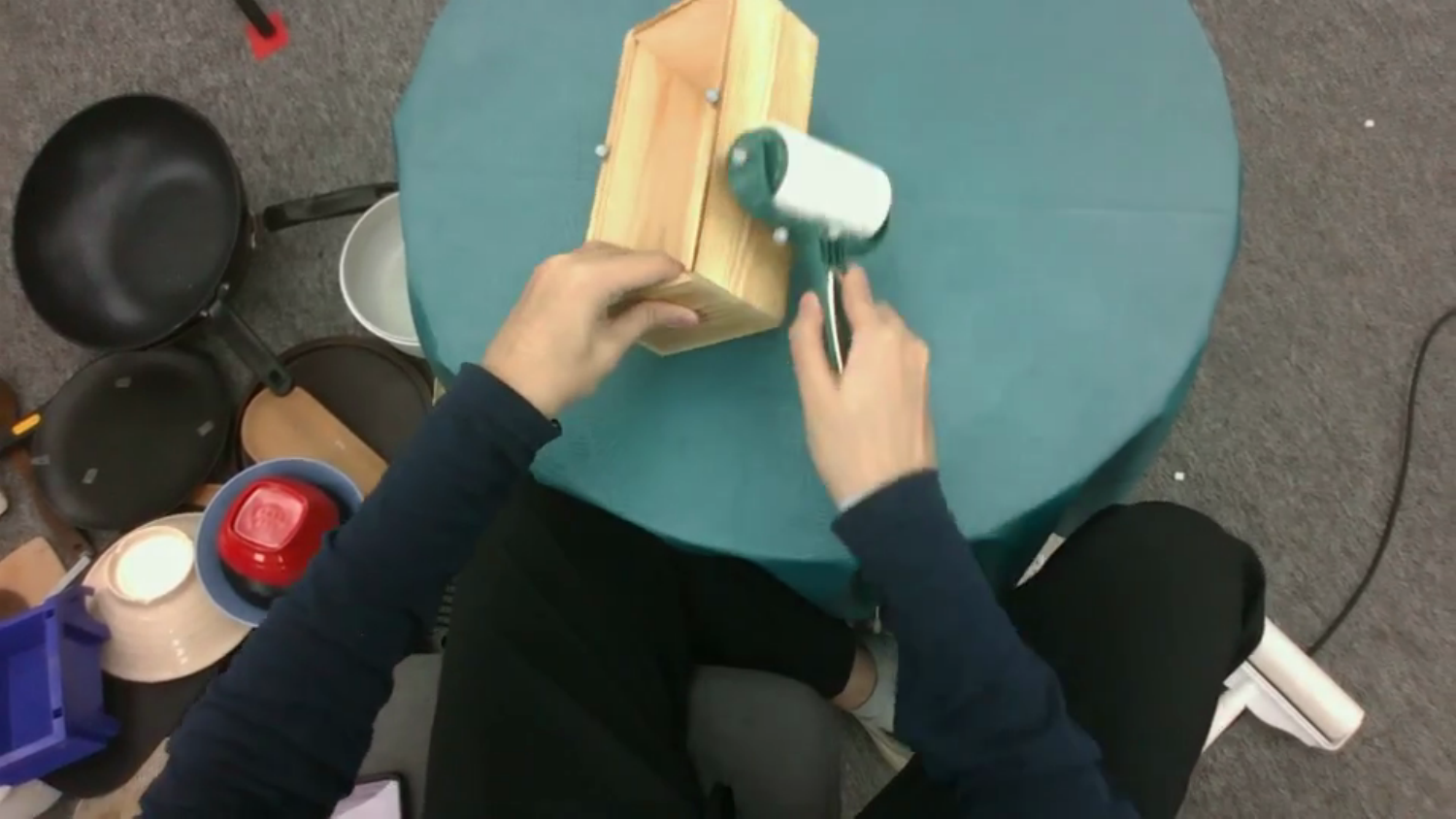}
& \cellimg{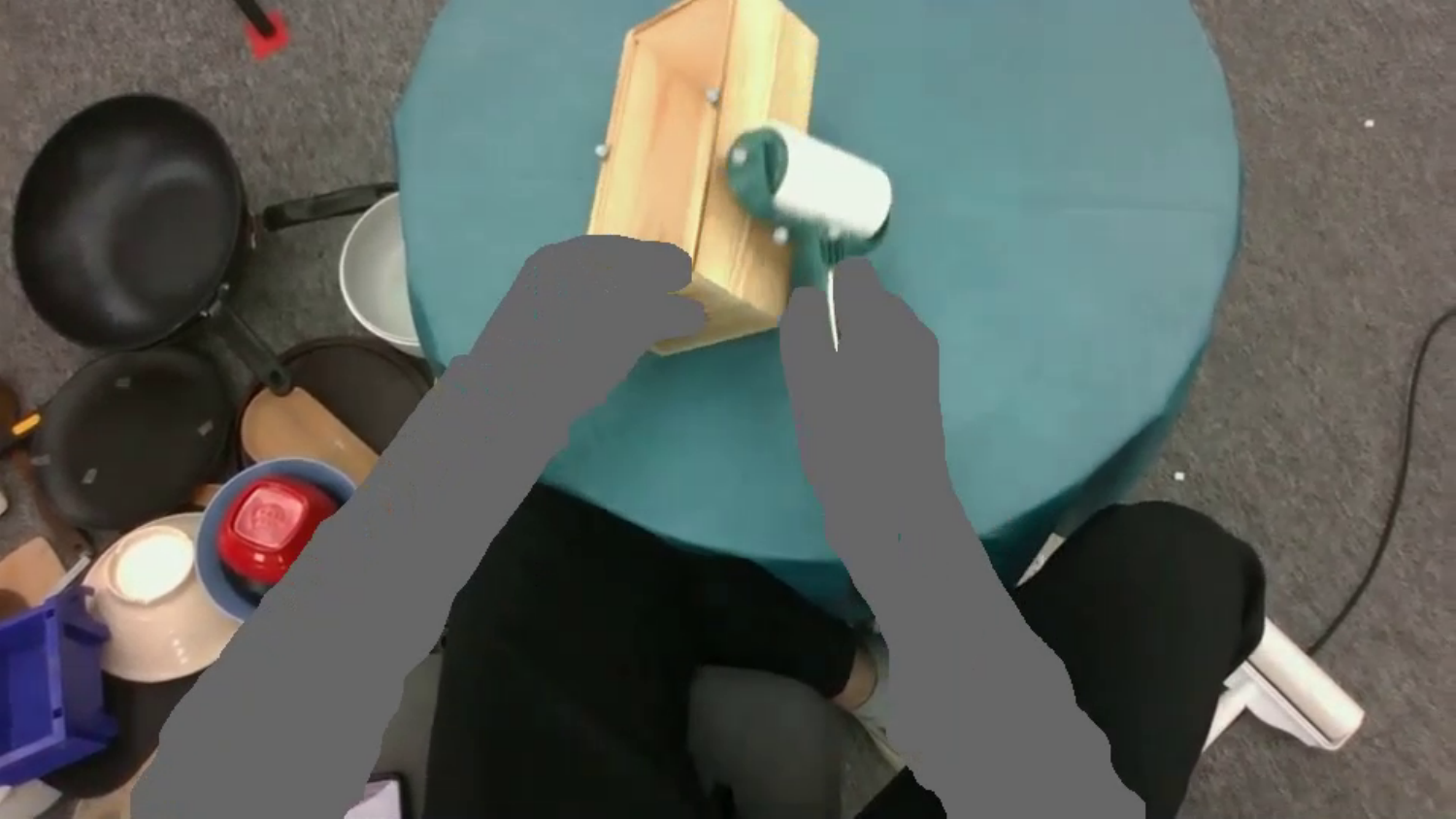}
& \cellimg{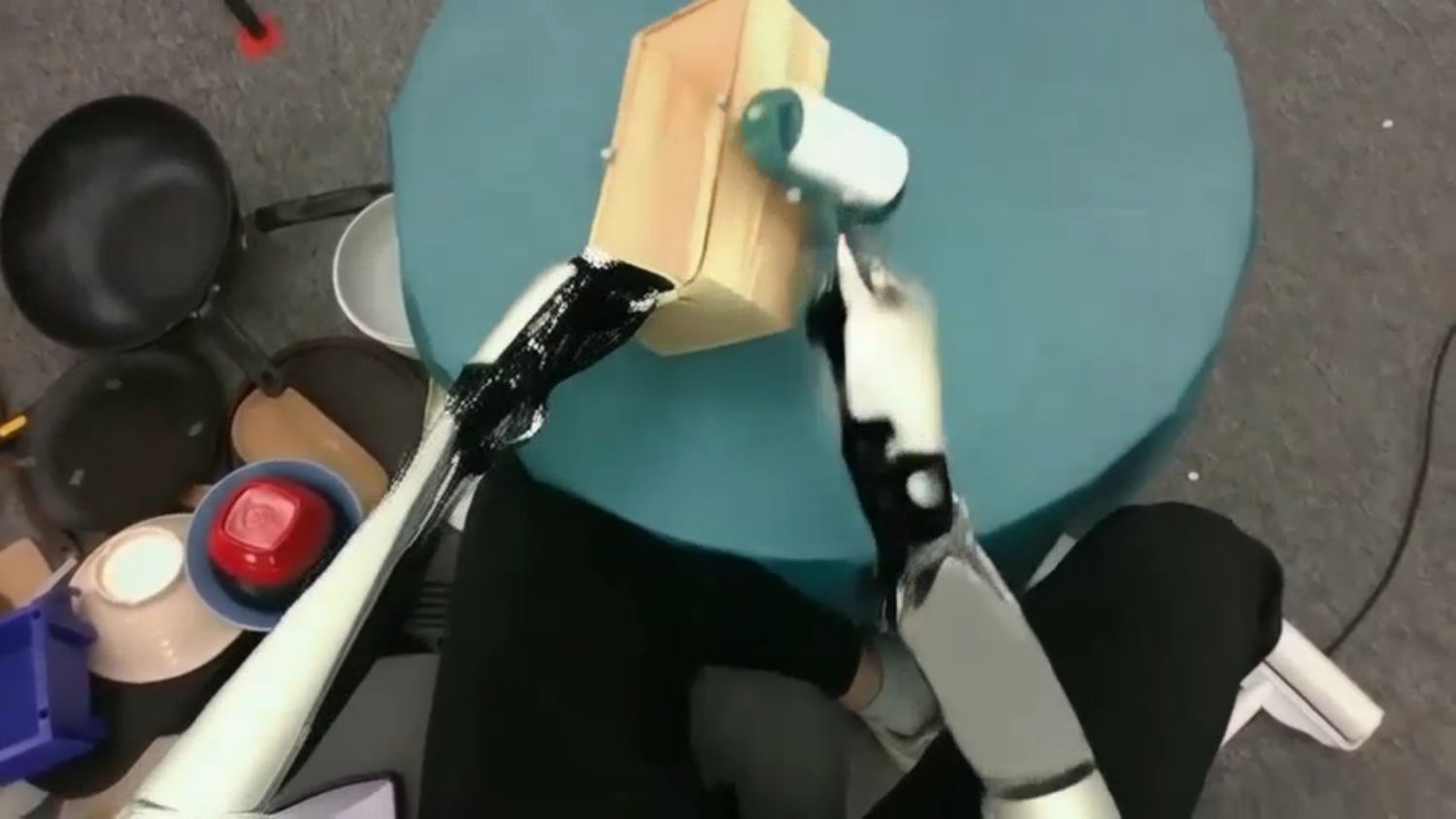}
& \cellimg{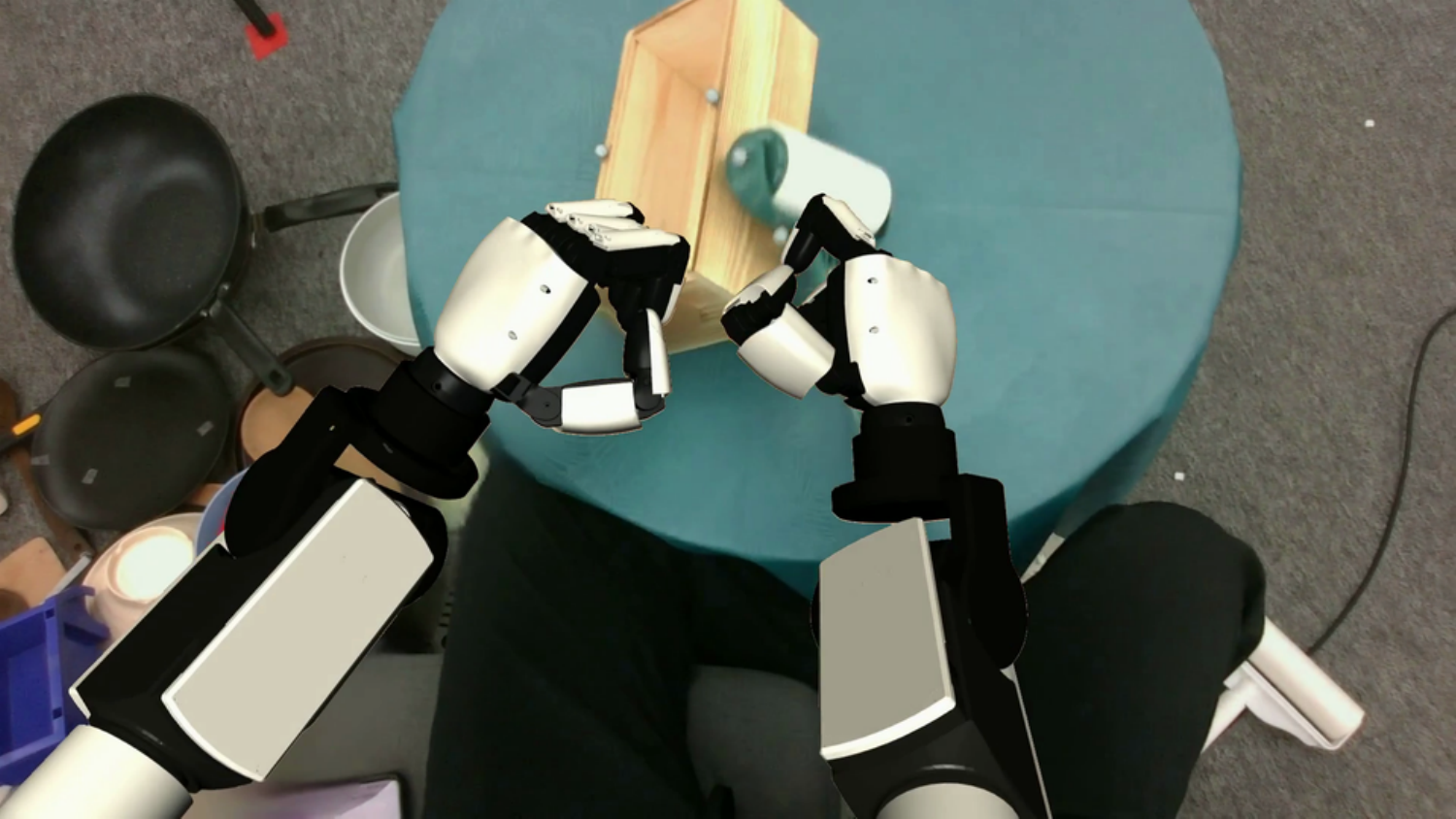}
& \cellimg{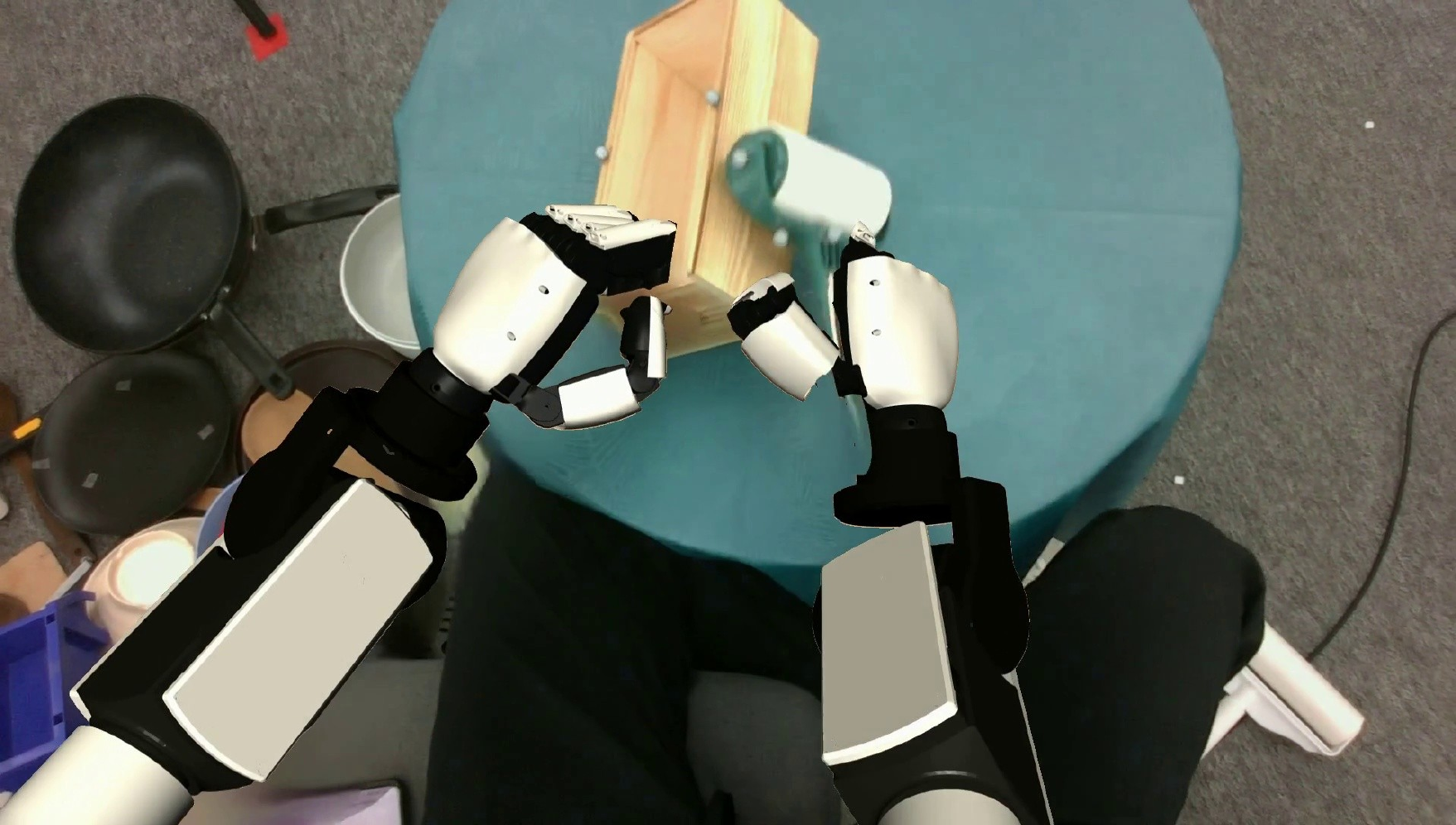} \\[-0.3ex]
\hline

\multirow[c]{6}{*}{\datasetlabelaria}
& \multirow[c]{2}{*}{\tasklabel{Mustard}}
& \cellimg{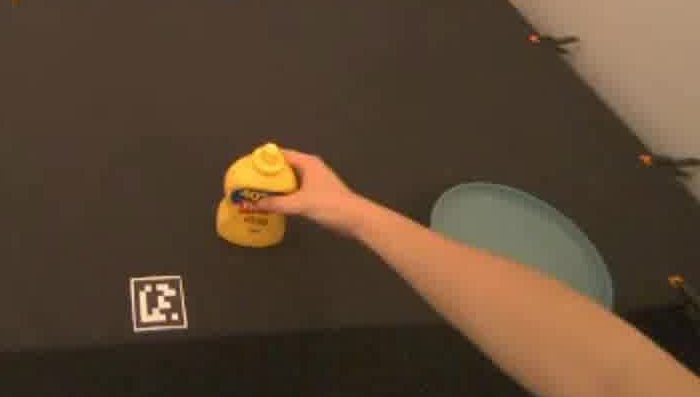}
& \cellimg{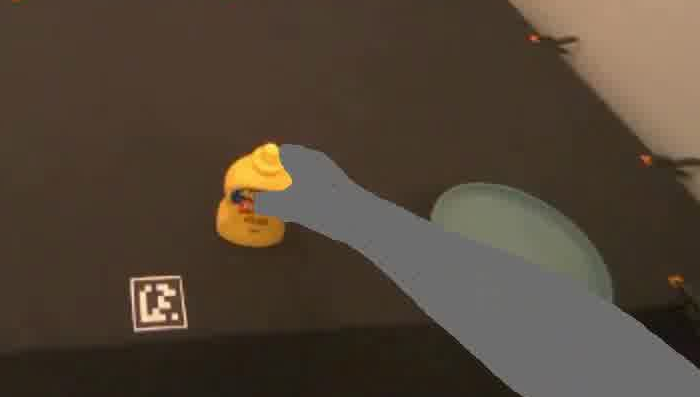}
& \cellimg{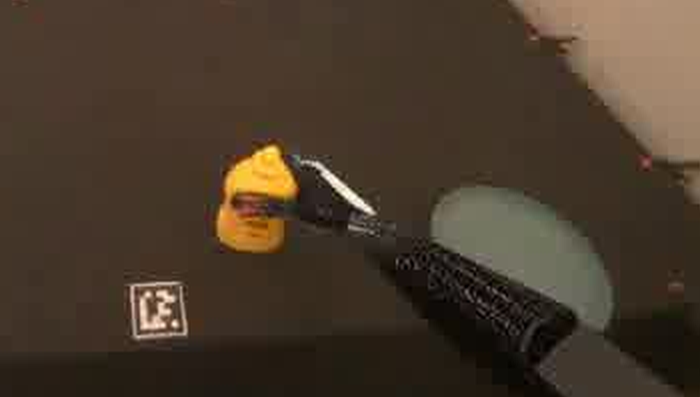}
& \cellimg{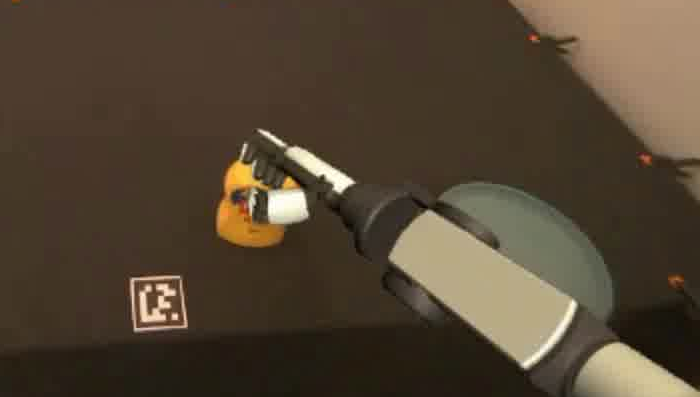}
& \cellimg{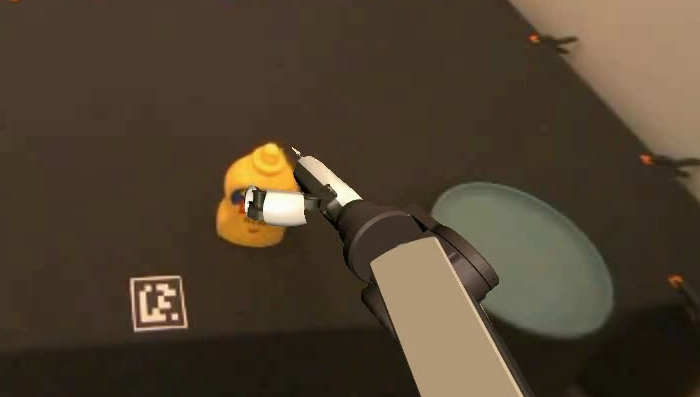} \\[-0.30ex]
&
& \cellimg{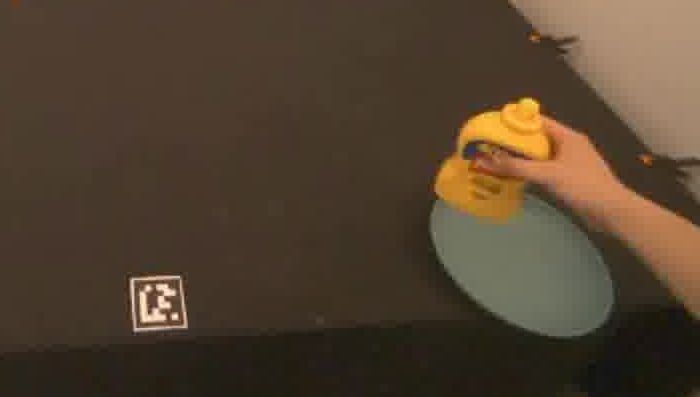}
& \cellimg{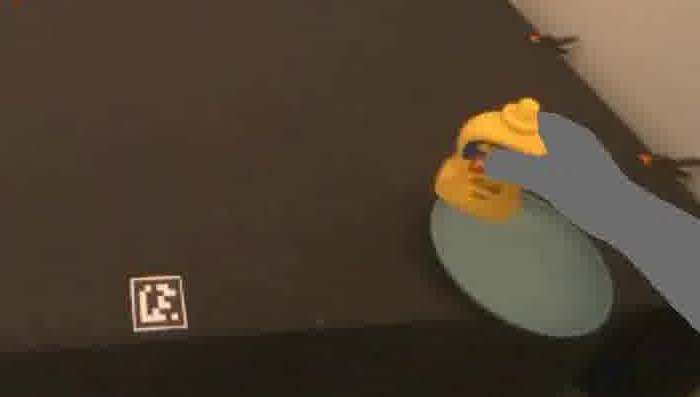}
& \cellimg{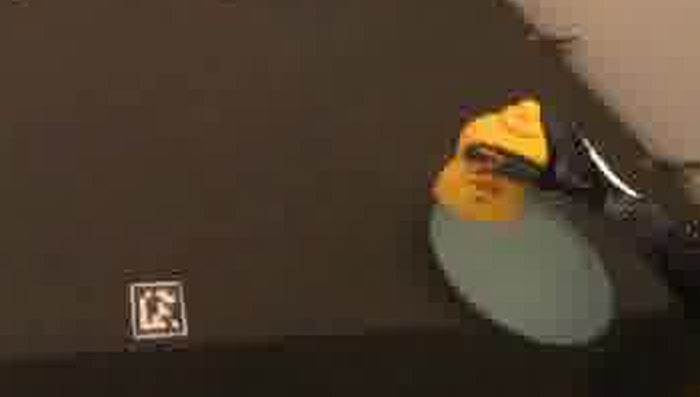}
& \cellimg{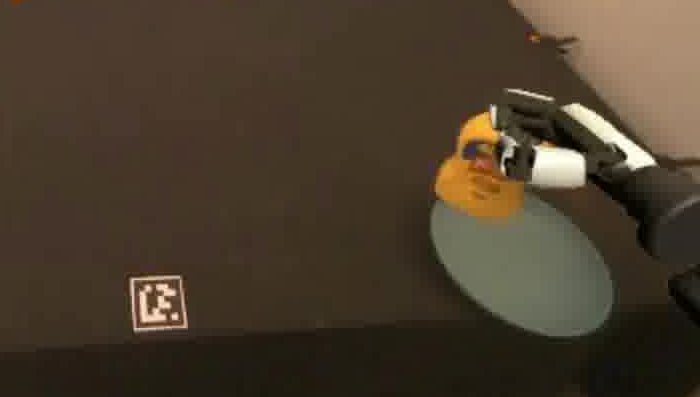}
& \cellimg{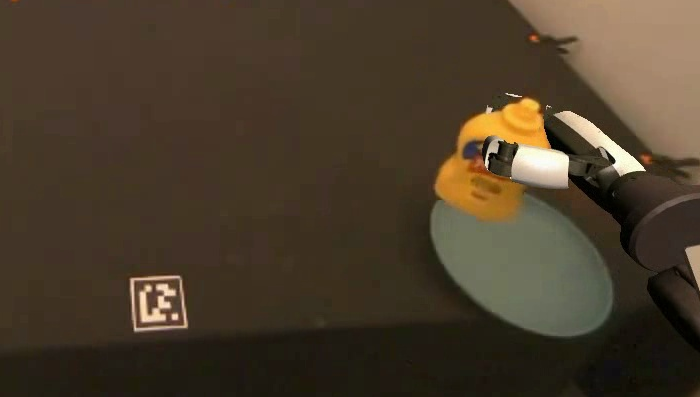} \\[0.12ex]
\cline{2-7}

& \multirow[c]{2}{*}{\tasklabel{Drawer}}
& \cellimg{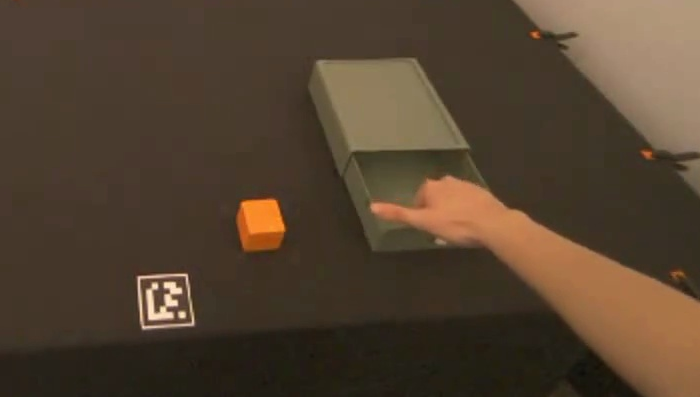}
& \cellimg{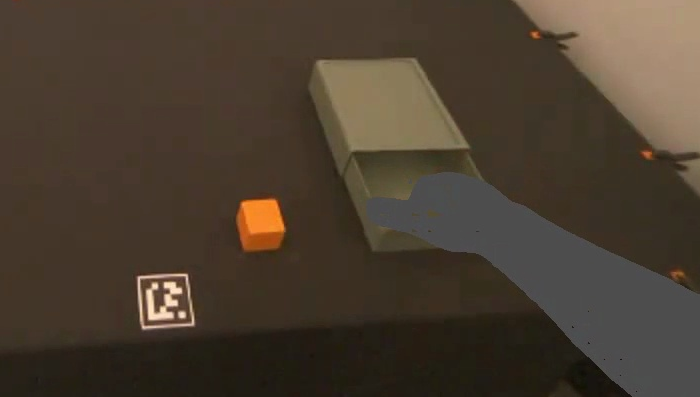}
& \cellimg{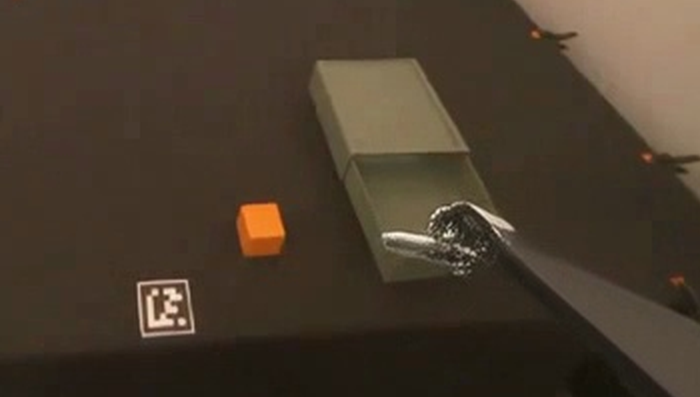}
& \cellimg{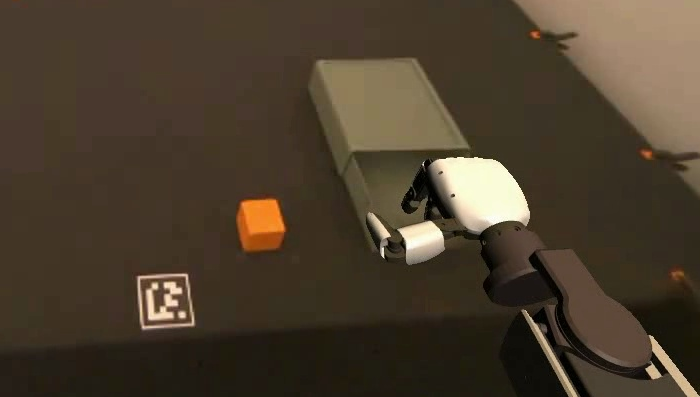}
& \cellimg{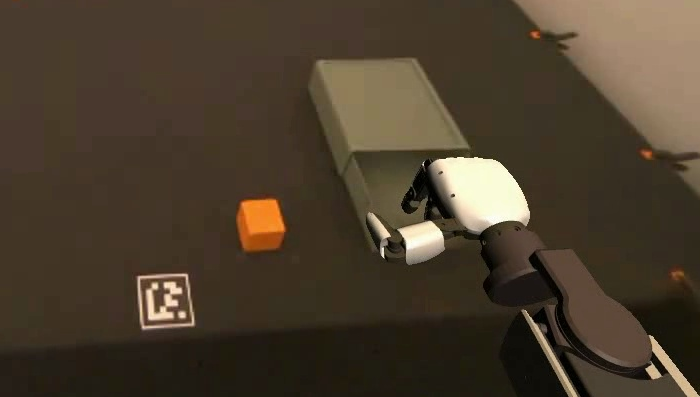} \\[0.30ex]
&
& \cellimg{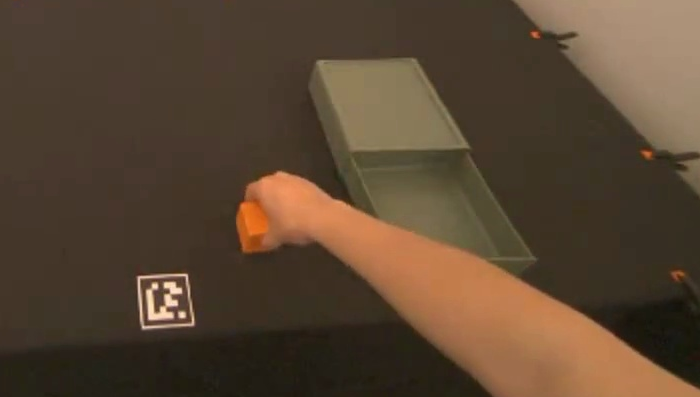}
& \cellimg{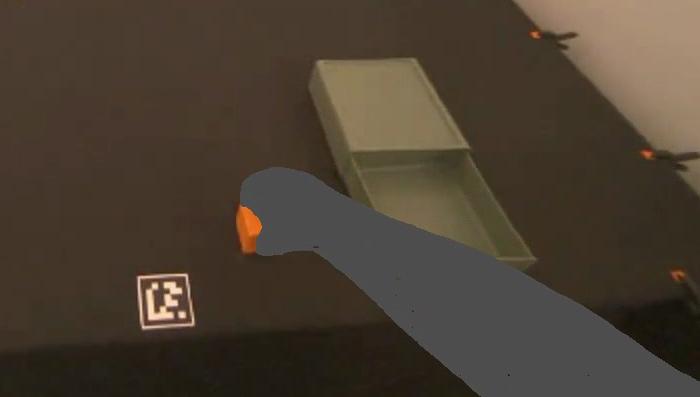}
& \cellimg{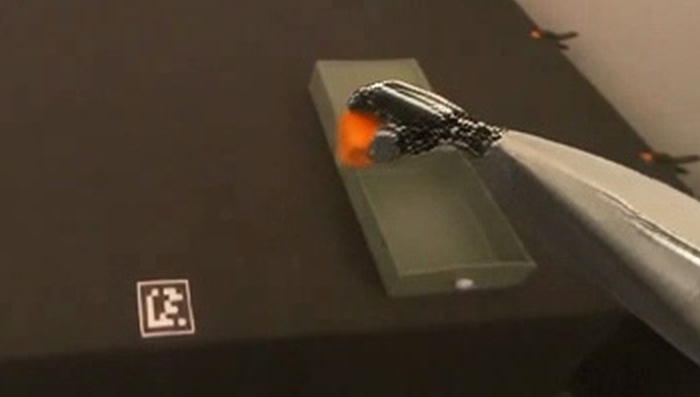}
& \cellimg{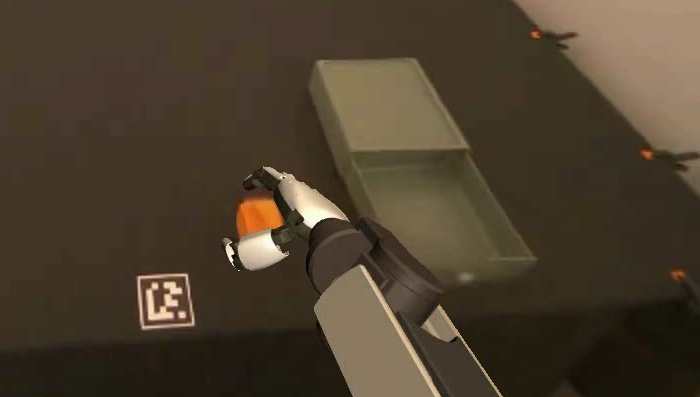}
& \cellimg{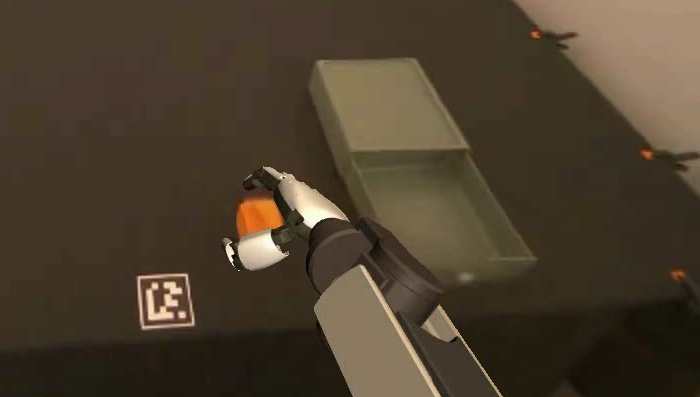} \\[0.12ex]
\cline{2-7}

& \multirow[c]{2}{*}{\tasklabel{Hammer}}
& \cellimg{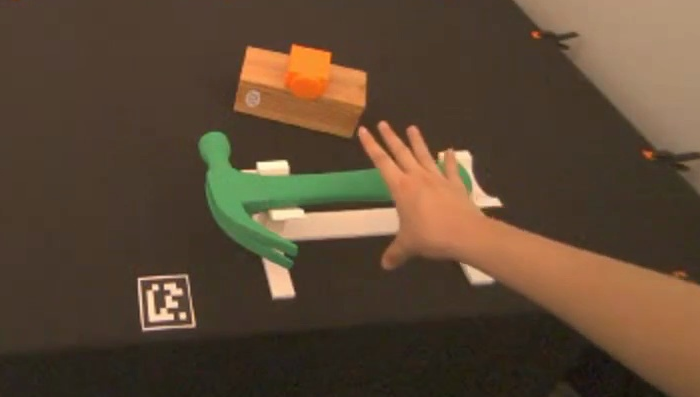}
& \cellimg{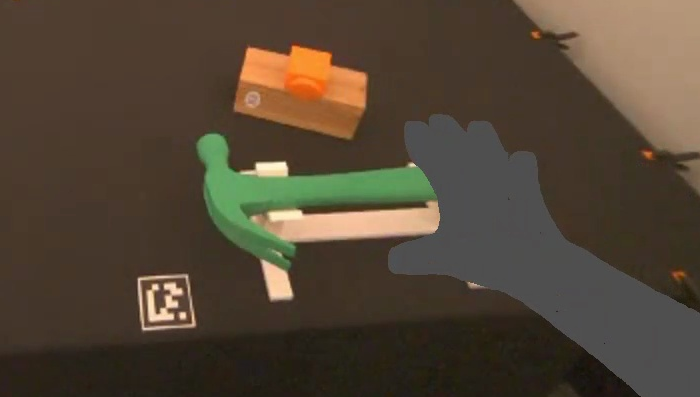}
& \cellimg{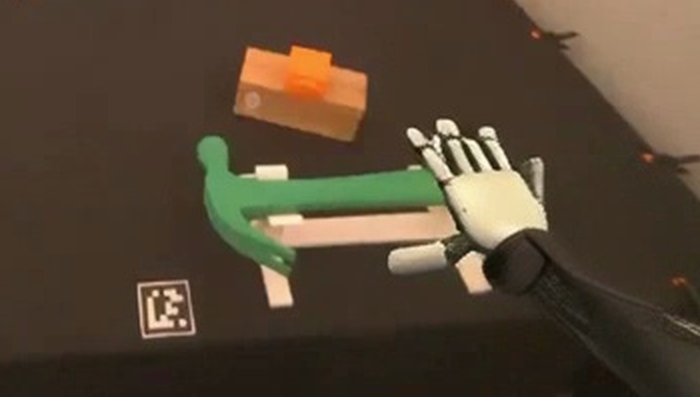}
& \cellimg{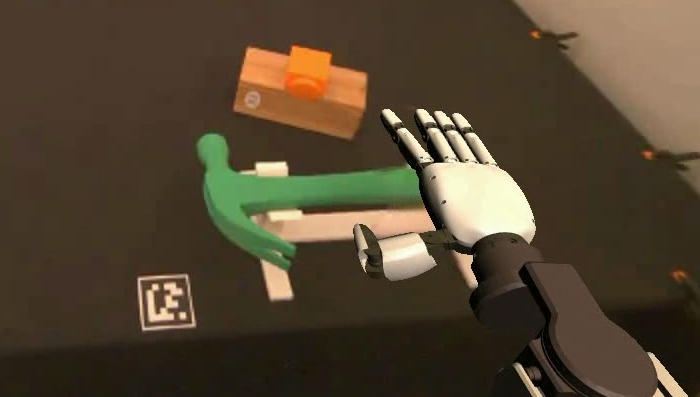}
& \cellimg{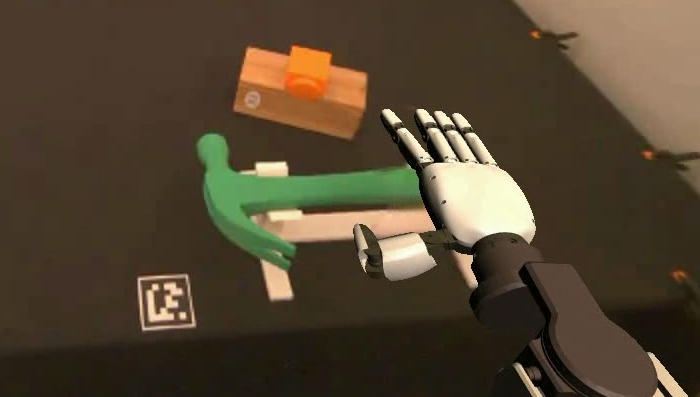} \\[0.30ex]
&
& \cellimg{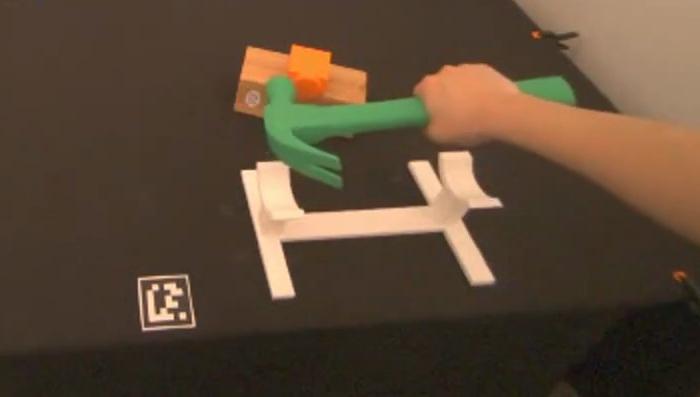}
& \cellimg{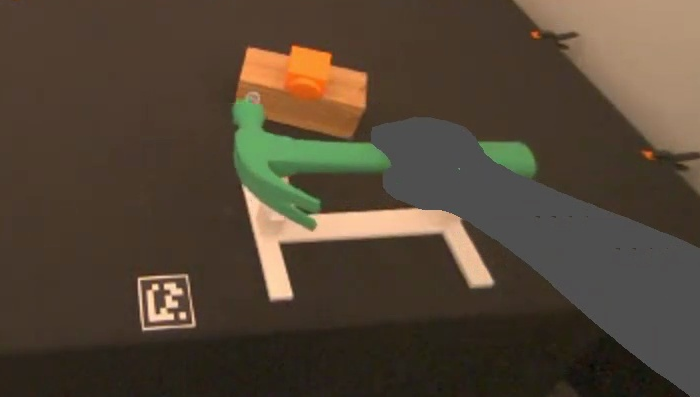}
& \cellimg{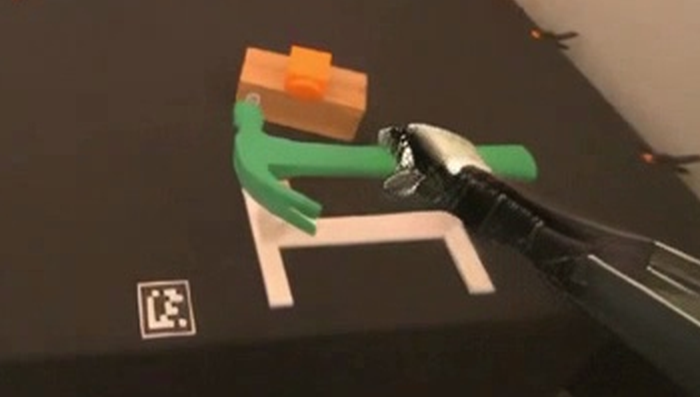}
& \cellimg{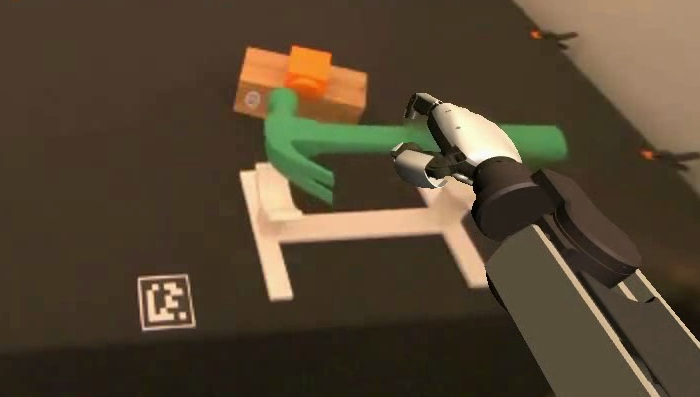}
& \cellimg{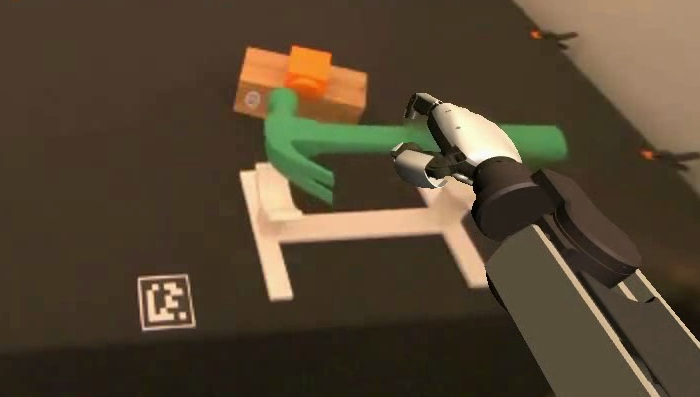} \\
\end{tabular}%
}
\caption{Visualization for different human-to-robot video generation methods. The figure shows six tasks from the TACO and Aria datasets for EgoMimic, VACE, Phantom, and EgoEngine.}
\label{fig:qualitative_tasks}
\end{figure}

We provide additional qualitative results for the visual branch of EgoEngine in Fig.~\ref{fig:qualitative_tasks}. The examples cover six manipulation tasks from both the TACO and Aria datasets, and compare EgoEngine with representative human-to-robot video generation baselines, including EgoMimic, VACE, and Phantom.

Across different tasks and viewpoints, EgoEngine produces more coherent robot-centric videos while better preserving the original scene layout, object appearance, and task-relevant interactions. In particular, compared with methods that primarily rely on direct video editing or appearance transfer, EgoEngine explicitly leverages robot rendering and scene-aware composition, which leads to more stable robot morphology, fewer visual artifacts around the manipulated objects, and better consistency between the generated robot motion and the underlying human demonstration.

These qualitative results further demonstrate that the visual branch can effectively bridge the embodiment gap between egocentric human videos and robot demonstrations. By replacing human hands with physically grounded robot embodiments while maintaining the original task context, EgoEngine generates visual demonstrations that are more suitable for downstream robot policy learning.

\section{Trajectory Optimization}

\subsection{Mode Switching}
The adaptive switching in the action branch operates sequentially over fixed temporal chunks of $H=20$ control steps, escalating complexity from Replay to MPC and RL only when lower-cost solvers fail to meet the object-centric tracking feasibility criteria.
The search space can be viewed as a MCTS-style implicit tree over chunk-level mode choices, where each node corresponds to the simulator state at a chunk boundary and each branch selects one mode from Replay, MPC, and RL.
Instead of using learned value estimates or full tree backup, EgoEngine applies a heuristic greedy selection from low-cost to high-cost modes.

\begin{figure}[t]
    \centering
    \vspace{-1em}
    \includegraphics[width=1.0\linewidth]{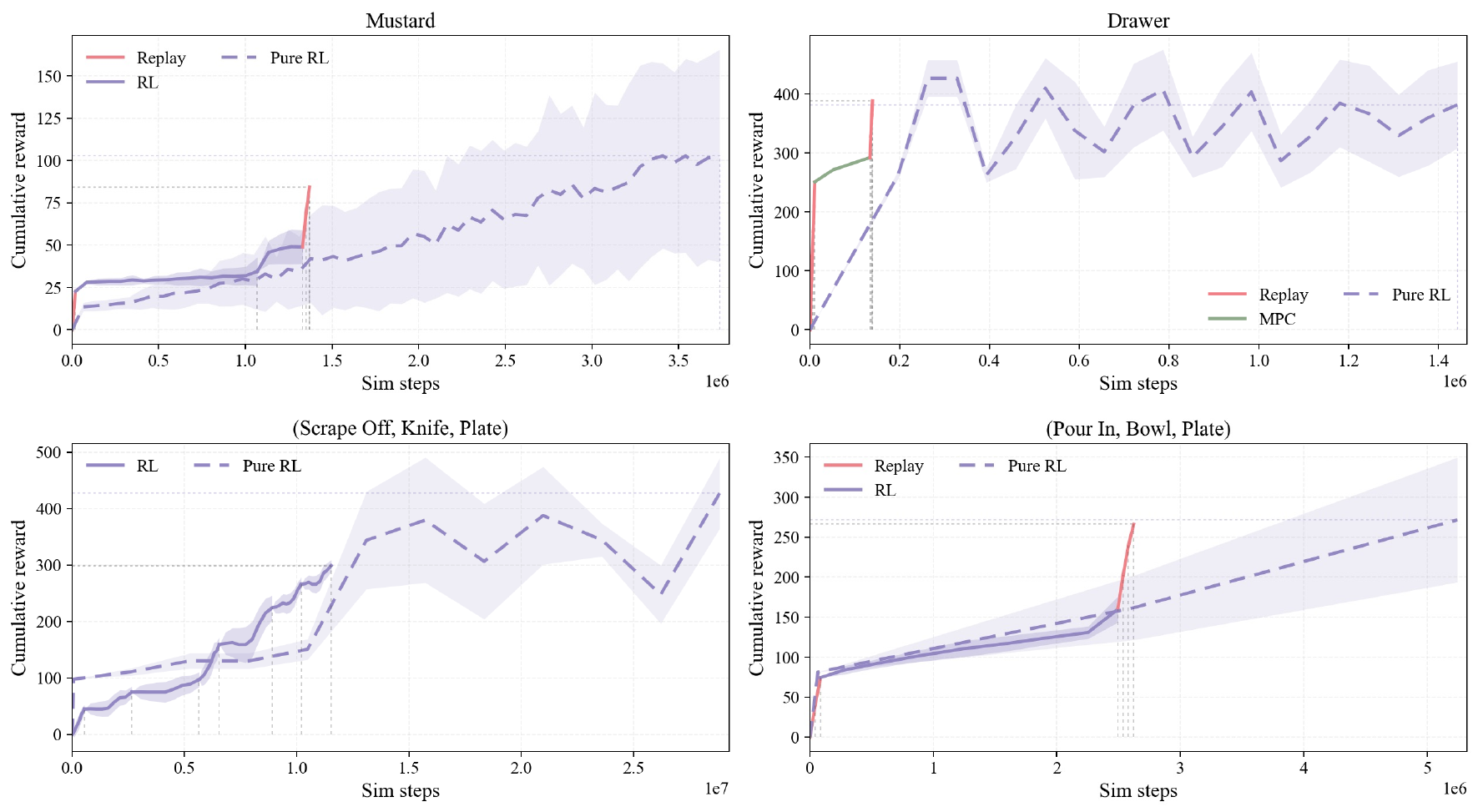}
    \vspace{-2em}
    \caption{
    Reward curves comparing EgoEngine with Pure RL on four long-horizon tasks: the Aria tasks Mustard and Drawer, and the TACO tasks (Scrape Off, Knife, Plate) and (Pour, Bowl, Plate).
    The cumulative reward is accumulated until task success, and the vertical dashed lines indicate chunk boundaries used by EgoEngine.
    EgoEngine combines replay, MPC, and RL-based refinement to solve each chunk, reaching task completion with substantially fewer simulation steps than Pure RL.
    Mustard is a challenging right-hand task, Drawer is a relatively simple left-hand task, and the two TACO tasks involve longer-horizon contact-rich manipulation.
    }
    \label{fig:reward_curve}
\end{figure}

At each chunk boundary, EgoEngine re-plans from the current simulator state and checks whether each candidate mode can satisfy the feasibility threshold over a two-chunk horizon, covering both the current chunk and the next chunk.
If the rollout is feasible, EgoEngine executes only the current chunk and re-evaluates the solver mode at the next chunk boundary.
Otherwise, it escalates from Replay to MPC and then to RL when stronger correction is needed.
This chunk-wise strategy avoids using expensive solvers on easy segments and returns to cheaper modes once the trajectory becomes feasible again.
We visualize this adaptive process in Fig.~\ref{fig:reward_curve}.
By filtering out redundant RL refinement over extended sequences, EgoEngine improves generation efficiency while maintaining trajectory feasibility, especially for long-horizon contact-rich tasks.

\subsection{Reward (Objective) Design}
\textbf{Object tracking reward and early termination.}
Let $T_o^t$ be the object pose estimated from the human demonstration, and let $\hat{T}_o^t$ be the corresponding object pose in simulation. 
We define the translation and rotation errors as
\begin{equation}
e_p^t =
\left\|
\mathrm{trans}(\hat{T}_o^t) -
\mathrm{trans}(T_o^t)
\right\|_2 ,
\end{equation}
and
\begin{equation}
e_R^t =
d_R\!\left(
\mathrm{rot}(\hat{T}_o^t),
\mathrm{rot}(T_o^t)
\right),
\end{equation}
where $d_R(\cdot,\cdot)$ denotes the geodesic distance on $\mathrm{SO}(3)$, defined as $d_R(R_1, R_2) = \arccos\!\bigl((\mathrm{tr}(R_1^\top R_2) - 1)/2\bigr)$.
The object tracking reward is then written as
\begin{equation}
r_{\mathrm{obj}}^t
=
C -
\sqrt{
\lambda_R (e_R^t)^2
+
\lambda_p (e_p^t)^2
},
\end{equation}
where $\lambda_R$ and $\lambda_p$ balance rotation and translation tracking. 
The constant $C$ is selected according to the early termination boundary. 
Specifically, if
\begin{equation}
\sqrt{
\lambda_R (e_R^t)^2
+
\lambda_p (e_p^t)^2
}
>
C ,
\end{equation}
the rollout is terminated early. 
Thus, the same object tracking score defines both the dense reward and the feasibility boundary.

\textbf{Human-mimic reward.}
Since action optimization uses a floating Cartesian wrist base abstraction with the same XHand joints, we regularize both the floating base pose and the finger joints toward the retargeted reference:
\begin{equation}
r_{\mathrm{human}}^t
=
-
\left(
\beta_x
\|x_t - x_t^{\mathrm{retar}}\|_2^2
+
\beta_R
d_R(R_t, R_t^{\mathrm{retar}})^2
+
\beta_q
\|q_t - q_t^{\mathrm{retar}}\|_2^2
\right),
\end{equation}
where $(x_t, R_t)$ is the floating hand base pose, $q_t$ is the finger joint vector, and $(x_t^{\mathrm{retar}}, R_t^{\mathrm{retar}}, q_t^{\mathrm{retar}})$ is the retargeted reference, and $\beta_x$, $\beta_R$, and $\beta_q$ are weighting coefficients.
This reward acts as a soft prior that keeps the optimized motion close to the human-retargeted trajectory while still allowing contact-driven corrections.

\textbf{Action smoothness reward.}
We penalize abrupt action changes between consecutive control steps:
\begin{equation}
r_{\mathrm{smooth}}^t
=
-
\left\|
a_t - a_{t-1}
\right\|_2^2 .
\end{equation}
This encourages temporally smooth control and reduces unstable contacts caused by high-frequency action oscillations.

\textbf{Contact reward.}
For dexterous manipulation, we use a sparse binary contact bonus. 
Let $\mathcal{C}_{\mathrm{thumb}}^t$ indicate whether the thumb is in contact with the manipulated object, and let $\mathcal{C}_{\mathrm{other}}^t$ indicate whether at least one of the other four fingers is in contact with the object. 
The contact reward is
\begin{equation}
r_{\mathrm{contact}}^t
=
c_{\mathrm{contact}}
\cdot
\mathbf{1}
\left[
\mathcal{C}_{\mathrm{thumb}}^t
\land
\mathcal{C}_{\mathrm{other}}^t
\right],
\end{equation}
where $c_{\mathrm{contact}}$ is a constant bonus. 
This reward encourages stable opposition-style grasping: the thumb and at least one non-thumb finger must contact the object simultaneously to receive the bonus.

\textbf{Lifting reward.}
For tasks that require lifting the manipulated object, we additionally encourage upward object motion along the vertical axis. 
Let $z_o^t$ denote the $z$-coordinate of the object in simulation at timestep $t$, and let $z_o^0$ denote its initial height. 
The lifting reward is defined as
\begin{equation}
r_{\mathrm{lift}}^t
=
\lambda_z
\left(
z_o^t - z_o^0
\right),
\end{equation}
where $\lambda_z$ controls the strength of the lifting objective. 
This term provides a simple dense signal for raising the object and is only applied when vertical object displacement is relevant to the task.

We use different optimization configurations for simulation-only evaluation and real-robot transfer. For TACO, no domain randomization is applied, and the human-mimic constraint reward and action-smoothness reward are disabled. The feasibility thresholds and reward scales are set for direct simulation evaluation, where object trajectories are directly available. For the Aria real-robot tasks, the object trajectories are extracted from egocentric human videos rather than ground-truth simulator states, and therefore contain reconstruction, tracking, and calibration errors. We consequently use looser feasibility thresholds, the human-mimic constraint reward on the floating base pose, action-smoothness regularization, and stronger perturbations to make the optimized trajectories robust to real-world transfer.

\textbf{Representative reward parameters.}
Across all tasks, we share the tracking weights, contact bonus, residual noise schedule, chunk length, and episode length.
Since different tasks exhibit different object-motion ranges and contact geometries, we only adjust the object position and rotation thresholds that define the early termination boundary, while keeping all reward coefficients fixed.
These thresholds are set from the reference object motion rather than searched as reward parameters.

For example, a static blade pressing on a plate (Cut, Knife, Plate) uses a tight rotation threshold ($0.9$ rad) to prevent the optimizer from satisfying the objective by tilting the blade away from the surface.
Tasks with larger demonstrated rotations use looser rotation thresholds: Scrape Off (Knife, Plate) and Brush (Brush, Bowl) use $1.2$ rad, while Pour (Bowl, Plate), which lifts and tilts the bowl by roughly $1.13$ rad, uses $1.5$ rad together with a tighter position threshold ($0.12$ m).
This limited threshold adjustment preserves a shared optimization objective while allowing EgoEngine to handle tasks with different contact physics and object-motion profiles.

\textbf{Domain randomization.} For the four Aria real-robot tasks, we apply domain randomization during trajectory optimization and policy-based refinement to account for real-world modeling errors, calibration errors, and noisy object trajectories extracted from egocentric human videos. Compared with the simulation-only setting, we use looser feasibility thresholds, including an object position threshold of $0.08\,\mathrm{m}$ and an object rotation threshold of $2.5\,\mathrm{rad}$, so that the optimizer can tolerate imperfect object-trajectory estimates instead of overfitting to noisy references. We set the contact reward scale to $2.0$ to encourage stable object interaction. For the human-mimic constraint, we use base-position and base-rotation reward scales of $0.2$ and $1.0$, respectively, while setting the joint reward scale to $0.0$ to avoid over-constraining noisy retargeted finger motions. We also use an action-smoothness reward scale of $0.8$ to reduce high-frequency action oscillations during real-robot transfer. For robustness, we use $8$ randomized rollouts with position, rotation, and joint noise scales of $0.045$, $1.0$, and $0.8$, respectively. We additionally randomize pair margins within $[-0.005, 0.005]$, workspace $xy$ offsets within $[-0.015, 0.015]$, and object mass scales within $[0.8, 1.2]$. These perturbations are specific to the Aria real-robot setting and are used to keep the generated actions robust while preserving alignment with the demonstrated task.

\subsection{Evaluation}
\label{app:evaluation}

We use \textbf{SR} (Success Rate), \textbf{Step}, \textbf{Reward}, and \textbf{Cost} for action branch evaluation in the main paper.
For evaluation, we only use the object-tracking reward, which is the same score used for the object-centric objective and early termination:
\begin{equation}
r_{\mathrm{eval}}^t=C - \sqrt{\lambda_R (e_R^t)^2 + \lambda_p (e_p^t)^2}.
\end{equation}
A rollout is terminated if the corresponding tracking error exceeds the boundary $C$.
No auxiliary terms, including human-mimic regularization, action smoothness, or contact bonus, are used during evaluation.
These auxiliary rewards are only used during optimization to improve stability and contact quality.

For a reference trajectory of length $T_i$, let $\tau_i$ denote the number of valid rollout steps before success or early termination.
To calculate \textbf{SR}, a trajectory is counted as successful only if it completes the full reference horizon without violating the object-tracking boundary.
Given $N$ evaluated trajectories, we define:
\begin{equation}
\mathrm{SR}=\frac{1}{N}\sum_{i=1}^{N}\mathbbm{1}[\tau_i=T_i],
\end{equation}
and the normalized rollout completion \textbf{Step} ratio as:
\begin{equation}
\mathrm{Step}=\frac{1}{N}\sum_{i=1}^{N}\frac{\tau_i}{T_i}.
\end{equation}
Thus, it measures the fraction of the reference trajectory completed before early termination, rather than the raw number of simulator steps.

We report \textbf{Reward} as the object-tracking reward ratio compared with the perfect reference trajectory.
For trajectory $i$, the accumulated evaluation reward is
$R_i=\sum_{t=1}^{\tau_i} r_{\mathrm{eval}}^t.$
Since a perfect trajectory obtains reward $C$ at every timestep, its maximum accumulated reward is $CT_i$.
We define
\begin{equation}
\mathrm{Reward}=\frac{1}{N}\sum_{i=1}^{N}\frac{R_i}{CT_i}.
\end{equation}
This metric jointly reflects tracking accuracy and rollout progress, because early termination reduces the accumulated reward.

\textbf{Cost} measures the average simulation steps required to generate one successful trajectory timestep.
For successful trajectories, we define
\begin{equation}
\mathrm{Cost}=\frac{\sum_i M_i}{\sum_i T_i},
\end{equation}
where $M_i$ is the number of simulator steps used to generate trajectory $i$, and the summation is taken only over successful trajectories with $\tau_i=T_i$.
Replay has the lowest cost because it directly executes the retargeted reference trajectory, while MPC and RL require additional simulation rollouts for local search or policy refinement.

\section{Policy Architecture and Training Details}
\label{app:policy}

We train the policy solely on synthetic robot demonstration data generated by EgoEngine. No real-robot teleoperation data or co-training scheme is used in this stage.

\subsection{Observation Encoder}

Our policy is based on HPT~\cite{hpt}, taking RGB observations and proprioceptive states as input, and directly outputs actions with a flow-matching head.

\textbf{Visual stem.}
Given an RGB image $\mathbf{I}\in\mathbb{R}^{H\times W\times 3}$, we first apply ImageNet normalization and encode it with a ResNet-18 backbone truncated before global average pooling. 
This produces a spatial feature map, which is flattened and linearly projected to the transformer hidden dimension. 
A set of learnable query tokens attends to the visual features through a single cross-attention block, producing a fixed number of visual tokens.

\textbf{Proprioceptive stem.}
The proprioceptive observation vector $\mathbf{q}\in\mathbb{R}^{d_q}$, including robot state information such as joint positions and end-effector pose, is first normalized and then projected with a linear layer. 
A set of learnable query tokens extracts a fixed number of proprioceptive tokens through a cross-attention block.

\textbf{Token fusion encoder.}
Tokens from all observation stems are concatenated along the sequence dimension. 
We prepend a small set of learnable context tokens and process the resulting sequence with a transformer encoder. 
The output context tokens serve as the compact observation representation for action prediction.

\subsection{Flow-Matching Action Decoder}

We predict an action horizon of length $T$ with a flow-matching transformer decoder conditioned on the encoder context tokens.

During training, given a ground-truth action sequence $a_1=\mathbf{a}$ and Gaussian noise $a_0\sim\mathcal{N}(0,I)$, we sample a continuous interpolation time $\tau\in(0,1]$ and construct
\begin{equation}
x_\tau = \tau a_0 + (1-\tau)a_1.
\end{equation}
The decoder takes $x_\tau$ together with a time embedding of $\tau$ as input and predicts the velocity field $v_\theta(x_\tau, \tau)$, which is regressed against the target velocity $\frac{dx_\tau}{d\tau} = a_0 - a_1$ that points from the clean action sequence toward the noise.

The decoder is implemented as a transformer with alternating self-attention and cross-attention blocks, where the action tokens attend to the encoder context tokens. A final linear layer maps decoder outputs to the action space.

At inference time, we initialize the action sequence from Gaussian noise at $\tau=1$ and integrate the learned velocity field in the decreasing-$\tau$ direction from $\tau=1$ to $\tau=0$ using fixed-step Euler updates. We use 10 inference steps in all experiments.

\subsection{Real Robot Experiments Setup}
\label{sec:supp_real_robot_setup}

\begin{figure}[t]
    \centering
    \includegraphics[width=\linewidth]{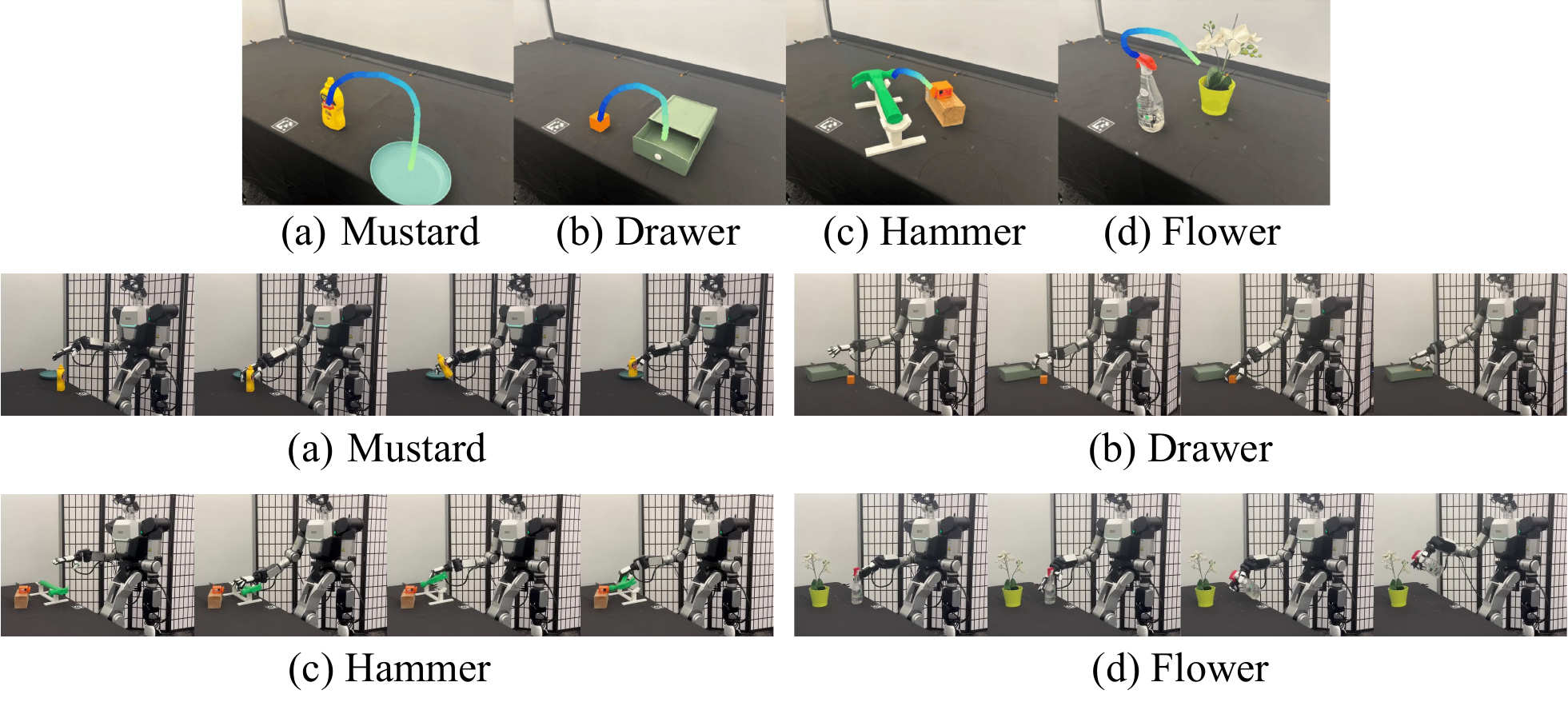}
    \caption{Real-robot task flow visualizations. (a) \textbf{Mustard}: pick and place the mustard bottle onto the target plate. (b) \textbf{Drawer}: open the drawer and place the cube inside. (c) \textbf{Hammer}: lift the hammer and strike the target nail. (d) \textbf{Flower}: grasp the water bottle and aim it toward the flower pot. Each task is evaluated with randomized object or target poses. }
    \label{fig:real_robot_task_flows}
\end{figure}

We evaluate EgoEngine on four real-robot manipulation tasks: Mustard, Drawer, Hammer, and Flower, as shown in Fig.~\ref{fig:real_robot_task_flows}. Each task introduces randomized initial object poses to evaluate robustness under realistic execution variations.

\textbf{Mustard.}
The Mustard task is a pick-and-place task. The robot needs to grasp a mustard bottle from its initial pose and place it onto the target plate. The initial position of the mustard bottle is randomized within a $10\,\mathrm{cm} \times 10\,\mathrm{cm}$ square region, with an orientation perturbation of $\pm 30^\circ$ around the vertical axis. 
The object is initially placed on the table surface. 
A trial is considered successful only if the mustard bottle is fully placed on the target plate and remains stable after release.

\textbf{Drawer.}
The Drawer task requires the robot to first open the drawer and then place a cube inside it. The cube initialization follows the same randomization setting as the Mustard task, with a $10\,\mathrm{cm} \times 10\,\mathrm{cm}$ positional offset range and a $\pm 30^\circ$ orientation perturbation. A trial is considered successful only if the drawer is opened and the cube is placed inside the drawer.

\textbf{Hammer.}
The Hammer task requires the robot to pick up a hammer from its holder and use it to strike a target nail. The nail position is randomized within a $5\,\mathrm{cm}$ offset range, and the initial hammer position is randomized with a lateral offset of up to $5\,\mathrm{cm}$. A trial is considered successful only if the robot successfully lifts the hammer from the holder and strikes the target nail.

\textbf{Flower.}
The Flower task requires the robot to pick up a water bottle and aim it toward a flower pot. The water bottle is randomized within a $10\,\mathrm{cm} \times 10\,\mathrm{cm}$ square region, with an orientation perturbation of $\pm 30^\circ$ around the vertical axis. The flower pot position is randomized within a $5\,\mathrm{cm}$ offset range, with arbitrary in-plane rotation. A trial is considered successful only if the robot picks up the water bottle and finally aligns it toward the flower pot.

\section{Dataset Quality Analysis}
\label{data_quality}

\subsection{Scalability and Sample Efficiency}
We further study whether EgoEngine-generated demonstrations can improve sample efficiency compared with real robot demonstrations under limited data budgets.
Specifically, we train downstream policies with increasing numbers of demonstrations from each data source and evaluate how quickly the success rate improves as more demonstrations are added.
This experiment tests whether EgoEngine can provide effective supervision without requiring the same amount of real-robot teleoperation data.

\begin{wrapfigure}{r}{0.78\textwidth}
    \centering
    \vspace{-1em}
\includegraphics[width=1.0\linewidth]{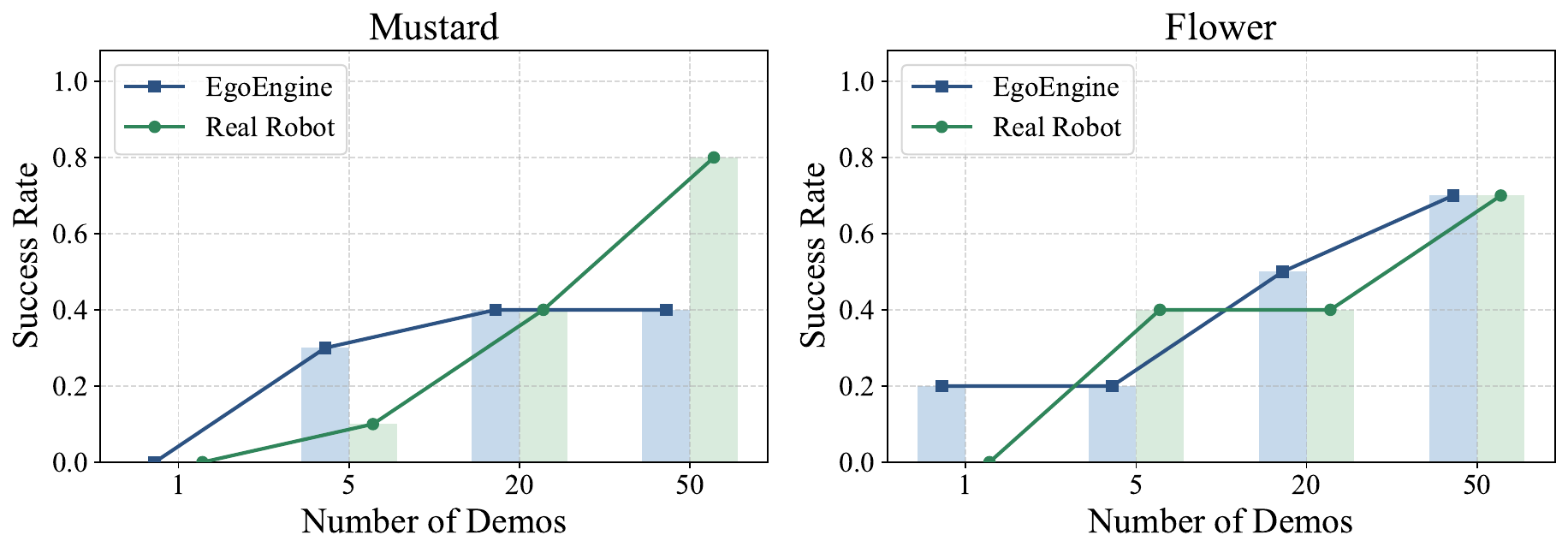}
    \vspace{-2em}
    \caption{Sample-efficiency comparison on Mustard and Flower under different demonstration budgets.}
    \label{fig:scaling}
    \vspace{-1em}
\end{wrapfigure}

We compare two training sources.
\textbf{Real Robot} uses real teleoperation demonstrations only, and serves as the reference for how performance scales with additional robot data.
\textbf{EgoEngine} uses robot demonstrations generated from egocentric human videos, evaluating whether the visual and action branches can produce training data that is useful for downstream policy learning.
As shown in Fig.~\ref{fig:scaling}, EgoEngine is not uniformly better than real robot demonstrations, especially at larger data budgets where real teleoperation can provide stronger task-specific supervision.
However, EgoEngine still shows a clear scaling trend as the number of generated demonstrations increases, and reaches non-trivial performance with only a small number of demonstrations.
On Flower, EgoEngine approaches the real-robot trend at larger data budgets and achieves positive success even with one demonstration.
On Mustard, EgoEngine improves steadily in the low-data regime, although real-robot teleoperation remains stronger at the largest budget.
These results suggest that EgoEngine-generated demonstrations are not a replacement for high-quality real robot data in all cases, but provide a sample-efficient source of supervision that can reduce the dependence on collecting large numbers of real-robot demonstrations for each task.

\subsection{Motion Pattern Analysis}

\textbf{Smoothness.}
We quantify action smoothness using Spectral Arc Length (SPARC)~\cite{sparc}, a frequency-domain metric defined on the trajectory speed profile. Given a speed sequence $s_{1:T}$, SPARC maps the trajectory to a scalar smoothness score:
\begin{equation}
    \mathrm{SPARC}(s_{1:T})
    \triangleq
    -\int_{0}^{\omega_c}
    \sqrt{
        \left(\frac{1}{\omega_c}\right)^2
        +
        \left(
            \frac{d}{d\omega}
            \frac{S(\omega)}{S(0)}
        \right)^2
    }\, d\omega ,
\end{equation}
where $S(\omega)$ is the magnitude spectrum of $s_{1:T}$. The cutoff frequency $\omega_c$ is chosen adaptively using an amplitude threshold $\overline{S}$ and an upper bound $\omega_c^{\max}$. We use zero-padding with factor $K=4$, $\omega_c^{\max}=15$, and $\overline{S}=0.05$. 
Higher values indicate smoother trajectories. 

The results are reported in Tab.~\ref{tab:action_quality}. EgoEngine datasets exhibit smoother action trajectories than real-robot demonstrations on average.
This is expected because EgoEngine starts from human motions, which are naturally smoother than visual teleoperation trajectories, and further applies action-smoothness regularization during real-robot action generation.
In contrast, visual teleoperation is affected by hand detection errors, sensing latency, and intermittent corrective motions, which can introduce high-frequency action changes.
Thus, beyond task success, EgoEngine also improves the motion quality of generated demonstrations by combining smooth human action priors with explicit smoothness constraints.

\begin{wraptable}{r}{0.36\textwidth}
    \vspace{-0.8em}
    \centering
    \resizebox{0.36\textwidth}{!}{
    \begin{tabular}{lcc}
        \toprule
        \multirow{2}{*}{Task} & \multicolumn{2}{c}{SPARC $\uparrow$} \\
        \cmidrule(lr){2-3}
        & Real & EgoEngine \\
        \midrule
        Mustard & -8.68  & \textbf{-4.88} \\
        Drawer  & -10.40 & \textbf{-7.49} \\
        Hammer  & \textbf{-3.21}  & -3.25 \\
        Flower  & -4.66  & \textbf{-3.88} \\
        \midrule
        All     & -6.60  & \textbf{-4.81} \\
        \bottomrule
    \end{tabular}
    }
    \caption{Action quality comparison between real-robot demonstrations and EgoEngine datasets across four manipulation tasks, measured by trajectory smoothness.}
    \label{tab:action_quality}
    \vspace{-1.0em}
\end{wraptable}

\textbf{Motion Pattern Robustness.}
We qualitatively compare the grasping behaviors induced by teleoperation and EgoEngine across four tasks by visualizing the grasp stage of representative trajectories.
The comparison reveals that EgoEngine is particularly effective when the task can benefit from smooth human-like motion priors and contact-aware correction, while real teleoperation remains stronger for some precise pinch-grasp cases.
For tasks such as Mustard and Drawer, which require stable pinch grasps during object pickup, teleoperated demonstrations can still provide more reliable finger placement.
In these cases, EgoEngine occasionally exhibits a small wrist-orientation offset at the pickup stage, which leads to unstable finger contact and less reliable grasps, even though the generated trajectories are robust in simulation under substantial domain randomization.

\begin{figure}[t]
    \centering
    \includegraphics[width=\linewidth]{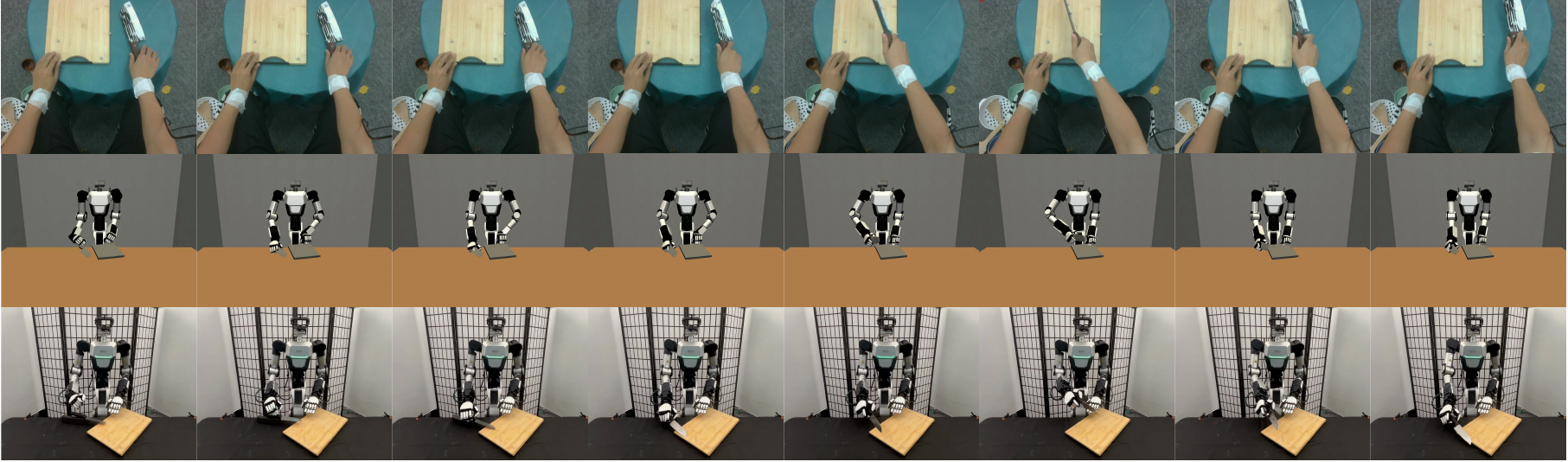}
    \caption{Visualization of (Cut, Knife, Plate) task from TACO dataset. We show the original video, simulation, and real-robot rollout after action refinement. For safety, we add an offset for the real robot rollout.}
    \label{fig:knife_action}
\end{figure}

In contrast, for Flower and Hammer, which rely more on power grasping, EgoEngine performs comparably to or better than teleoperation.
For Flower, EgoEngine produces grasps on the water bottle that are nearly as stable as those from teleoperated demonstrations.
For Hammer, EgoEngine performs noticeably better, consistently adopting a more effective pre-grasp strategy by rotating the wrist toward the thumb side and opening the thumb and index fingers widely before contact, which leads to more stable hammer acquisition.
These results suggest that EgoEngine is currently more reliable for manipulation patterns that benefit from smooth human motion and power-grasp structure, while precision pinch grasps remain a key failure mode.

\textbf{Complex Tasks Beyond Visual Teleoperation.}
We further demonstrate that EgoEngine can use human videos to generate robot actions for complex manipulation tasks that are difficult to collect through visual teleoperation, as shown in Fig.~\ref{fig:knife_action}. We take a long-horizon TACO video demonstration of cutting on a plate with a knife, reconstruct the task in simulation, and transfer the generated action trajectory back to the real robot with a similar knife and cutting board. This task is substantially more difficult than simple pick-and-place: the robot must grasp the knife handle, establish a stable tool grasp, perform in-hand reorientation, and maintain a useful cutting pose while interacting with the plate. Such behaviors are extremely difficult to collect through visual teleoperation because the operator does not directly observe the full robot hand state and must control a high-dimensional dexterous hand through noisy visual hand tracking, sensing latency, and limited real-time dexterity. As a result, precise tool acquisition and in-hand adjustment, such as grasping the knife handle and reorienting it for cutting, are nearly infeasible to perform reliably through teleoperation. EgoEngine addresses this difficulty through a Real-to-Sim-to-Real pipeline: the human video provides a task-level motion prior, the action branch converts it into an executable robot trajectory in simulation, and the resulting trajectory is replayed on the real robot. The successful real-robot rollout shows that EgoEngine can transfer complex human tool-use behaviors into physically plausible and task-directed robot motions, highlighting an advantage over visual teleoperation in settings where direct robot data collection is prohibitively difficult.

\end{document}